\documentclass[acmsmall]{acmart}

\usepackage{footnote}
\usepackage{multirow}
\usepackage{makecell}
\usepackage{algorithm}
\usepackage{algorithmic}
\usepackage{amsmath}

\usepackage{float}
\usepackage{subfigure}
\usepackage{tikz}
\usetikzlibrary{shapes,arrows}
\usepackage{hyperref}
\usepackage{appendix}
\usepackage[skins]{tcolorbox}
\tcbuselibrary{breakable}
\usepackage{bm}
\usepackage{amsbsy}
\usepackage{amsfonts}
\usepackage{etoolbox}
\usepackage{tabularx}

\DeclareMathOperator*{\argmax}{arg\,max}

\makeatletter
\newcommand\textlist[4]{
  \let\last@item\relax
  \let\last@sep\relax
  \@for\@ii:=#4\do{
    \ifx\last@item\relax\else
      \ifx\last@sep\relax
        \def\last@sep{#2}
      \else#1\fi
      #3{\last@item}
    \fi
    \let\last@item\@ii
  }%
  \ifx\last@item\relax\else
    \last@sep#3{\last@item}
  \fi
}
\makeatother

\newcommand{\citett}{\textlist{, }{ and }{\citet}}

\AtBeginDocument{
  }

\setcopyright{acmlicensed}
\copyrightyear{2024}
\acmYear{2024}
\acmDOI{XXXXXXX.XXXXXXX}

\acmJournal{JACM}
\acmVolume{37}
\acmNumber{4}
\acmArticle{111}
\acmMonth{9}

\begin{document}

\title{Graph Retrieval-Augmented Generation: A Survey}

\author{Boci Peng}
\authornote{Both authors contributed equally to this research.}
\affiliation{
  \institution{School of Intelligence Science and
Technology, Peking University}
  \city{Beijing}
  \country{China}
}
\email{bcpeng@stu.pku.edu.cn}

\author{Yun Zhu}
\authornotemark[1]
\affiliation{
  \institution{College of Computer Science and Technology, Zhejiang University}
  \city{Hangzhou}
  \country{China}
}
\email{zhuyun_dcd@zju.edu.cn}

\author{Yongchao Liu}
\affiliation{
  \institution{Ant Group}
  \city{Hangzhou}
  \country{China}
}
\email{yongchao.ly@antgroup.com}

\author{Xiaohe Bo}
\affiliation{
  \institution{Gaoling School of Artificial Intelligence, Renmin University of China}
  \city{Beijing}
  \country{China}}
\email{bellebxh@gmail.com}

\author{Haizhou Shi}
\affiliation{
 \institution{Rutgers University}
 \city{New Brunswick}
 \state{New Jersey}
 \country{US}
 }
 \email{haizhou.shi@rutgers.edu}

\author{Chuntao Hong}
\affiliation{
  \institution{Ant Group}
  \city{Hangzhou}
  \country{China}
}
\email{chuntao.hct@antgroup.com}

\author{Yan Zhang}
\authornote{Corresponding Author.}
\affiliation{
  \institution{School of Intelligence Science and
Technology, Peking University}
  \city{Beijing}
  \country{China}
}
\email{zhyzhy001@pku.edu.cn}

\author{Siliang Tang}
\affiliation{
  \institution{College of Computer Science and Technology, Zhejiang University}
  \city{Hangzhou}
  \country{China}
}
\email{siliang@zju.edu.cn}

\renewcommand{\shortauthors}{Peng et al.}

\begin{abstract}
Recently, Retrieval-Augmented Generation (RAG) has achieved remarkable success in addressing the challenges of Large Language Models (LLMs) without necessitating retraining. By referencing an external knowledge base, RAG refines LLM outputs, effectively mitigating issues such as ``hallucination'', lack of domain-specific knowledge, and outdated information. However, the complex structure of relationships among different entities in databases presents challenges for RAG systems. In response, GraphRAG leverages structural information across entities to enable more precise and comprehensive retrieval, capturing relational knowledge and facilitating more accurate, context-aware responses.
Given the novelty and potential of GraphRAG, a systematic review of current technologies is imperative. This paper provides the first comprehensive overview of GraphRAG methodologies. We formalize the GraphRAG workflow, encompassing Graph-Based Indexing, Graph-Guided Retrieval, and Graph-Enhanced Generation. We then outline the core technologies and training methods at each stage. Additionally, we examine downstream tasks, application domains, evaluation methodologies, and industrial use cases of GraphRAG. Finally, we explore future research directions to inspire further inquiries and advance progress in the field. In order to track recent progress in this field, we set up a repository at \url{https://github.com/pengboci/GraphRAG-Survey}.
\end{abstract}

\ccsdesc[500]{Computing methodologies~Knowledge representation and reasoning}
\ccsdesc[500]{Information systems~Information retrieval}
\ccsdesc[300]{Information systems~Data mining}

\keywords{Large Language Models, Graph Retrieval-Augmented Generation, Knowledge Graphs, Graph Neural Networks}

\maketitle

\section{Introduction}\label{sec:intro}
The development of Large Language Models like GPT-4~\cite{ref:gpt4}, Qwen2~\cite{ref:qwen2}, and LLaMA~\cite{ref:llama3} has sparked a revolution in the field of artificial intelligence, fundamentally altering the landscape of natural language processing. These models, built on Transformer~\cite{ref:transformer} architectures and trained on diverse and extensive datasets, have demonstrated unprecedented capabilities in understanding, interpreting, and generating human language. The impact of these advancements is profound, stretching across various sectors including healthcare~\cite{ref:med1,ref:med2,ref:med3}, finance~\cite{ref:financialgpt,ref:finalcialgpt2}, and education~\cite{ref:edu1,ref:edu2}, where they facilitate more nuanced and efficient interactions between humans and machines.

Despite their remarkable language comprehension and text generation capabilities, LLMs may exhibit limitations due to a lack of domain-specific knowledge, real-time updated information, and proprietary knowledge, which are outside LLMs' pre-training corpus. These gaps can lead to a phenomenon known as ``hallucination''~\cite{ref:hallucination} where the model generates inaccurate or even fabricated information. Consequently, it is imperative to supplement LLMs with external knowledge to mitigate this problem. Retrieval-Augmented Generation (RAG)~\cite{ref:ragsurvey1,ref:ragsurvey2,ref:ragsurvey3,ref:ragsurvey4,ref:ragsurvey5,ref:ragsurvey6,ref:ragsurvey7} emerged as a significant evolution, which aims to enhance the quality and relevance of generated content by integrating a retrieval component within the generation process. The essence of RAG lies in its ability to dynamically query a large text corpus to incorporate relevant factual knowledge into the responses generated by the underlying language models. This integration not only enriches the contextual depth of the responses but also ensures a higher degree of factual accuracy and specificity. RAG has gained widespread attention due to its exceptional performance and broad applications, becoming a key focus within the field.

\begin{figure*}
    \centering
    \setlength{\abovecaptionskip}{0.15cm}
    \includegraphics[scale=0.37]{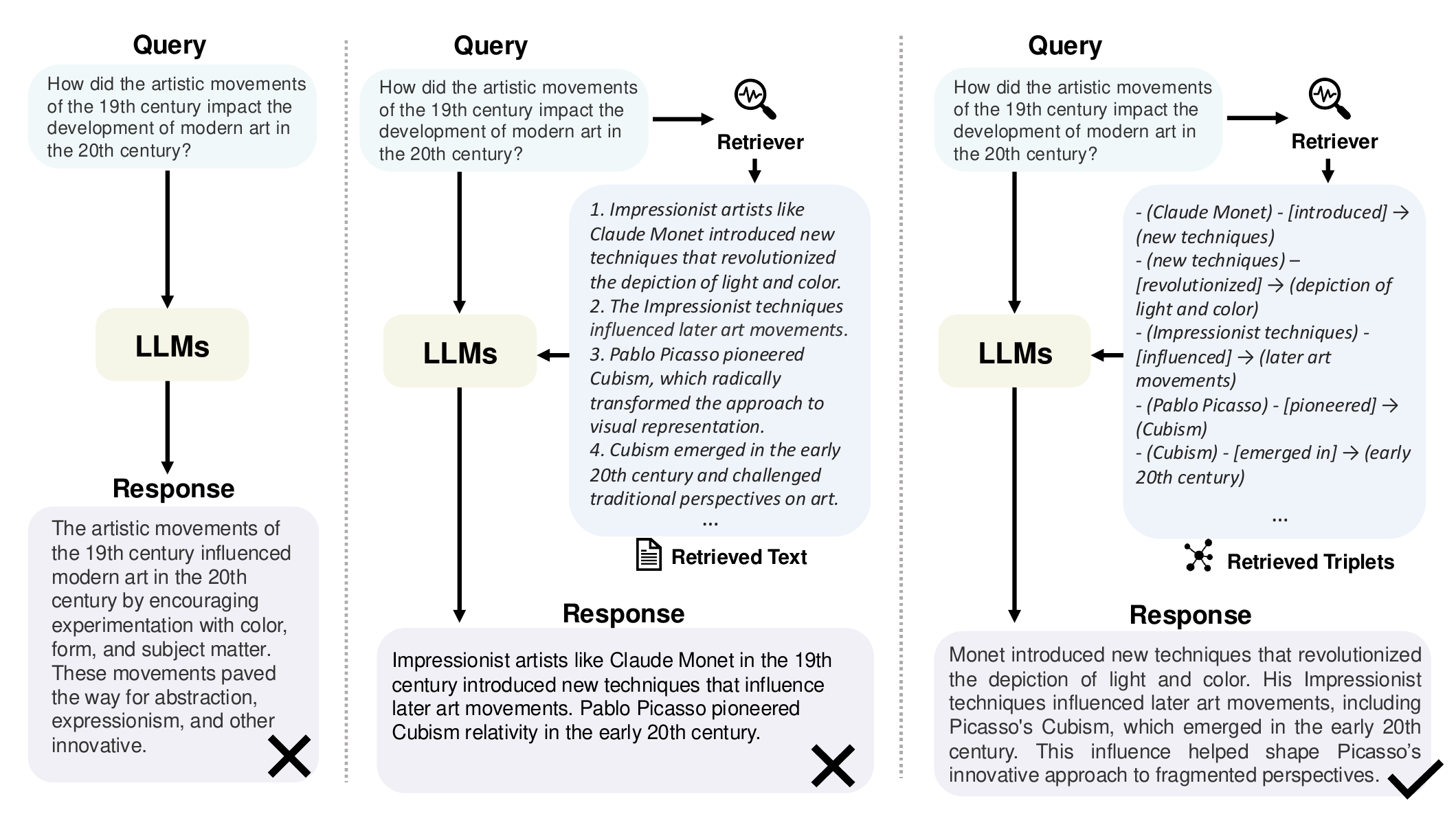}
    \caption{Comparision between Direct LLM, RAG, and GraphRAG. Given a user query, direct answering by LLMs may suffer from shallow responses or lack of specificity. RAG addresses this by retrieving relevant textual information, somewhat alleviating the issue. However, due to the text's length and flexible natural language expressions of entity relationships, RAG struggles to emphasize ``influence'' relations, which is the core of the question. While, GraphRAG methods leverage explicit entity and relationship representations in graph data, enabling precise answers by retrieving relevant structured information.}
    \label{fig:1}
\end{figure*}

Although RAG has achieved impressive results and has been widely applied across various domains, it faces limitations in real-world scenarios: (1) \emph{Neglecting Relationships}: In practice, textual content is not isolated but interconnected. Traditional RAG fails to capture significant structured relational knowledge that cannot be represented through semantic similarity alone. For instance, in a citation network where papers are linked by citation relationships, traditional RAG methods focus on finding the relevant papers based on the query but overlook important citation relationships between papers. (2) \emph{Redundant Information}: RAG often recounts content in the form of textual snippets when concatenated as prompts. This makes context become excessively lengthy, leading to the ``lost in the middle'' dilemma~\cite{ref:lostin}. (3) \emph{Lacking Global Information}: RAG can only retrieve a subset of documents and fails to grasp global information comprehensively, and hence struggles with tasks such as Query-Focused Summarization~(QFS).

Graph Retrieval-Augmented Generation (GraphRAG)~\cite{ref:graphrag,ref:grag,ref:gnn-rag} emerges as an innovative solution to address these challenges. Unlike traditional RAG, GraphRAG retrieves graph elements containing relational knowledge pertinent to a given query from a pre-constructed graph database, as depicted in Figure~\ref{fig:1}. These elements may include nodes, triples, paths, or subgraphs, which are utilized to generate responses. GraphRAG considers the interconnections between texts, enabling a more accurate and comprehensive retrieval of relational information. Additionally, graph data, such as knowledge graphs, offer abstraction and summarization of textual data, thereby significantly shortening the length of the input text and mitigating concerns of verbosity. By retrieving subgraphs or graph communities, we can access comprehensive information to effectively address the QFS challenge by capturing the broader context and interconnections within the graph structure.

In this paper, we are the first to provide a systematic survey of GraphRAG. Specifically, we begin by introducing the GraphRAG workflow, along with the foundational background knowledge that underpins the field. Then, we categorize the literature according to the primary stages of the GraphRAG process: Graph-Based Indexing (G-Indexing), Graph-Guided Retrieval (G-Retrieval), and Graph-Enhanced Generation (G-Generation) in Section~\ref{sec:indexing}, Section~\ref{sec:retrieval} and Section~\ref{sec:generation} respectively, detailing the core technologies and training methods within each phase. 
Furthermore, we investigate downstream tasks, application domains, evaluation methodologies, and industrial use cases of GraphRAG. This exploration elucidates how GraphRAG is being utilized in practical settings and reflects its versatility and adaptability across various sectors. Finally, acknowledging that research in GraphRAG is still in its early stages, we delve into potential future research directions. This prognostic discussion aims to pave the way for forthcoming studies, inspire new lines of inquiry, and catalyze progress within the field, ultimately propelling GraphRAG toward more mature and innovative horizons.

Our contributions can be summarized as follows:

\begin{itemize}
\item We provide a comprehensive and systematic review of existing state-of-the-art GraphRAG methodologies. We offer a formal definition of GraphRAG, outlining its universal workflow which includes G-Indexing, G-Retrieval, and G-Generation. 

\item We discuss the core technologies underpinning existing GraphRAG systems, including G-Indexing, G-Retrieval, and G-Generation. For each component, we analyze the spectrum of model selection, methodological design, and enhancement strategies currently being explored. Additionally, we contrast the diverse training methodologies employed across these modules.

\item We delineate the downstream tasks, benchmarks, application domains, evaluation metrics, current challenges, and future research directions pertinent to GraphRAG, discussing both the progress and prospects of this field. Furthermore, we compile an inventory of existing industry GraphRAG systems, providing insights into the translation of academic research into real-world industry solutions.

\end{itemize}

\noindent\textbf{Organization.}\quad The rest of the survey is organized as follows: Section~\ref{sec:surveys} compares related techniques, while Section~\ref{sec:formalization} outlines the general process of GraphRAG. Sections~\ref{sec:indexing} to~\ref{sec:generation} categorize the techniques associated with GraphRAG's three stages: G-Indexing, G-Retrieval, and G-Generation. Section~\ref{sec:training} introduces the training strategies of retrievers and generators. Section~\ref{sec:app} summarizes GraphRAG's downstream tasks, corresponding benchmarks, application domains, evaluation metrics, and industrial GraphRAG systems. Section~\ref{sec:future} provides an outlook on future directions. Finally, Section~\ref{sec:conclusion} concludes the content of this survey.

\begin{figure*}
    \centering
    \includegraphics[width=1\linewidth]{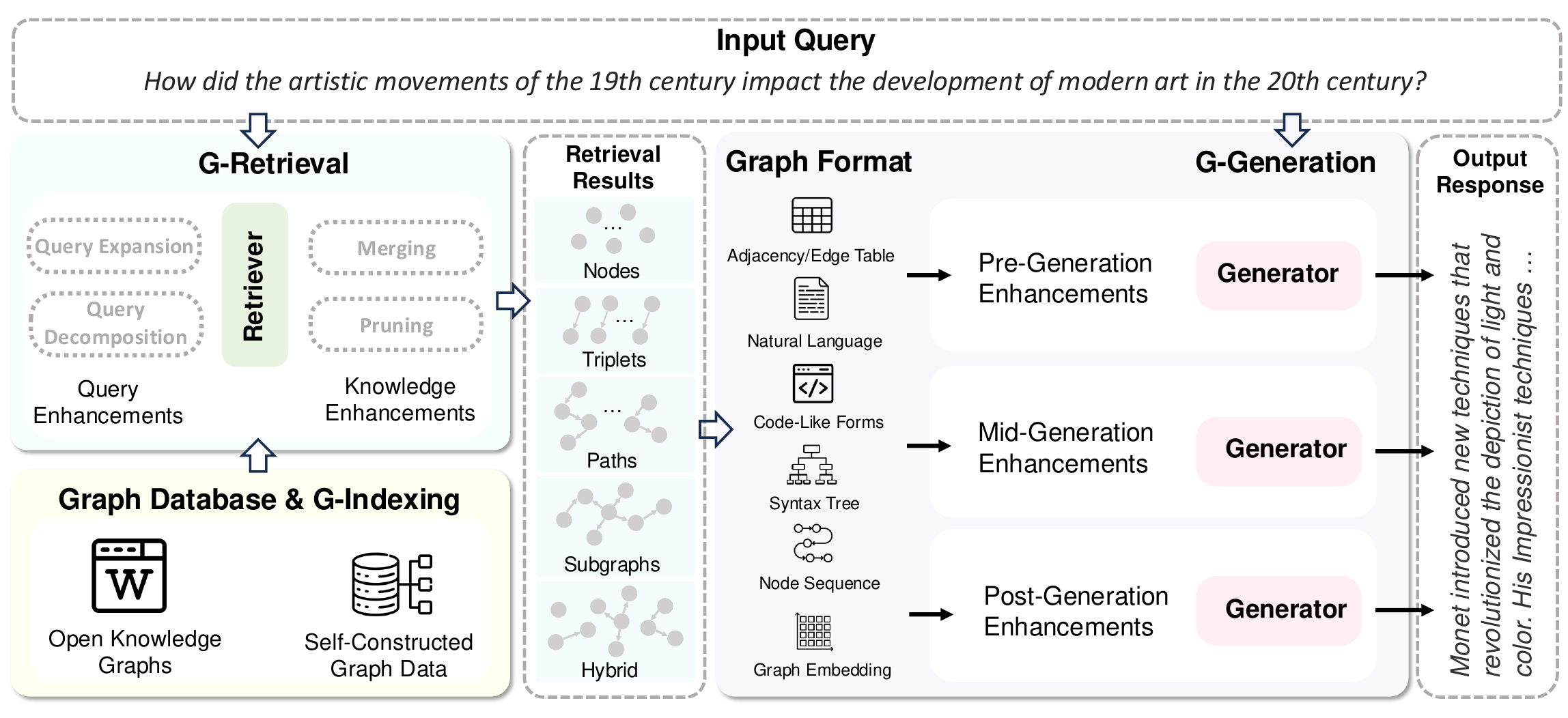}
    \caption{The overview of the GraphRAG framework for question answering task. In this survey, we divide GraphRAG into three stages: G-Indexing, G-Retrieval, and G-Generation. We categorize the retrieval sources into open-source knowledge graphs and self-constructed graph data. Various enhancing techniques like query enhancement and knowledge enhancement may be adopted to boost the relevance of the results. Unlike RAG, which uses retrieved text directly for generation, GraphRAG requires converting the retrieved graph information into patterns acceptable to generators to enhance the task performance.}
    \label{fig:overview}
\end{figure*}

\section{Comparison with Related Techniques and Surveys}
\label{sec:surveys}
In this section, we compare Graph Retrieval-Augmented Generation (GraphRAG) with related techniques and corresponding surveys, including RAG, LLMs on graphs, and Knowledge Base Question Answering (KBQA).

\subsection{RAG}
RAG combines external knowledge with LLMs for improved task performance, integrating domain-specific information to ensure factuality and credibility. In the past two years, researchers have written many comprehensive surveys about RAG~\cite{ref:ragsurvey1,ref:ragsurvey2,ref:ragsurvey3,ref:ragsurvey4,ref:ragsurvey5,ref:ragsurvey6,ref:ragsurvey7}. For example,~\citett{ref:ragsurvey1,ref:ragsurvey5} categorize RAG methods from the perspectives of retrieval, generation, and augmentation.~\citett{ref:ragsurvey4} review RAG methods for databases with different modalities.~\citett{ref:ragsurvey6} systematically summarize the evaluation of RAG methods. These works provide a structured synthesis of current RAG methodologies, fostering a deeper understanding and suggesting future directions of the area.

From a broad perspective, GraphRAG can be seen as a branch of RAG, which retrieves relevant relational knowledge from graph databases instead of text corpus. However, compared to text-based RAG, GraphRAG takes into account the relationships between texts and incorporates the structural information as additional knowledge beyond text. Furthermore, during the construction of graph data, raw text data may undergo filtering and summarization processes, enhancing the refinement of information within the graph data. Although previous surveys on RAG have touched upon GraphRAG, they predominantly center on textual data integration. This paper diverges by placing a primary emphasis on the indexing, retrieval, and utilization of structured graph data, which represents a substantial departure from handling purely textual information and spurs the emergence of many new techniques. 

\subsection{LLMs on Graphs}
LLMs are revolutionizing natural language processing due to their excellent text understanding, reasoning, and generation capabilities, along with their generalization and zero-shot transfer abilities. 
Although LLMs are primarily designed to process pure text and struggle with non-Euclidean data containing complex structural information, such as graphs~\cite{ref:cansolve,ref:gpt4graph}, numerous studies~\cite{ref:gfmsurvey1,ref:gfmsurvey2,ref:gfmsurvey3,ref:gfmsurvey4,ref:gfmsurvey5,ref:gfmsurvey6,ref:gfmsurvey7,ref:gfmsurvey8,ref:engine,ref:graphbridge} have been conducted in these fields. These papers primarily integrate LLMs with GNNs to enhance modeling capabilities for graph data, thereby improving performance on downstream tasks such as node classification, edge prediction, graph classification, and others. For example,~\citett{ref:engine} propose an efficient fine-tuning method named ENGINE, which combines LLMs and GNNs through a side structure for enhancing graph representation.

Different from these methods, GraphRAG focuses on retrieving relevant graph elements using queries from an external graph-structured database. In this paper, we provide a detailed introduction to the relevant technologies and applications of GraphRAG, which are not included in previous surveys of LLMs on Graphs.

\subsection{KBQA}
KBQA is a significant task in natural language processing, aiming to respond to user queries based on external knowledge bases~\cite{ref:kbqasurvey1,ref:kbqasurvey2,ref:kbqasurvey3,ref:kbqasurvey4}, thereby achieving goals such as fact verification, passage retrieval enhancement, and text understanding. Previous surveys typically categorize existing KBQA approaches into two main types: Information Retrieval (IR)-based methods and Semantic Parsing (SP)-based methods. Specifically, IR-based methods~\cite{ref:rog,ref:tog,ref:tog2,ref:unikgqa,ref:sr,ref:rra,ref:kd-cot,ref:kg-agent} retrieve information related to the query from the knowledge graph (KG) and use it to enhance the generation process. While SP-based methods~\cite{ref:noname3,ref:dara,ref:retrack,ref:rng,ref:arcaneqa,ref:hsge} generate a logical form (LF) for each query and execute it against knowledge bases to obtain the answer.

GraphRAG and KBQA are closely related, with IR-based KBQA methods representing a subset of GraphRAG approaches focused on downstream applications. In this work, we extend the discussion beyond KBQA to include GraphRAG’s applications across various downstream tasks. Our survey provides a thorough and detailed exploration of GraphRAG technology, offering a comprehensive understanding of existing methods and potential improvements.

\section{Preliminaries}
\label{sec:formalization}
In this section, we introduce background knowledge of GraphRAG for easier comprehension of our survey. First, we introduce Text-Attributed Graphs which is a universal and general format of graph data used in GraphRAG. Then, we provide formal definitions for two types of models that can be used in the retrieval and generation stages: Graph Neural Networks and Language Models.

\subsection{Text-Attributed Graphs}
The graph data used in Graph RAG can be represented uniformly as Text-Attributed Graphs (TAGs), where nodes and edges possess textual attributes. Formally, a text-attributed graph can be denoted as $\mathcal{G} = (\mathcal{V}, \mathcal{E}, \mathcal{A}, \{\mathbf{x}_v\}_{v \in \mathcal{V}},\{\mathbf{e}_{i,j}\}_{i,j \in \mathcal{E}})$, where $\mathcal{V}$ is the set of nodes, $\mathcal{E} \subseteq \mathcal{V} \times \mathcal{V}$ is the set of edges, $\mathcal{A} \in \{0,1\}^{\lvert \mathcal{V} \rvert \times \lvert \mathcal{V} \rvert}$ is the adjacent matrix. Additionally, $\{\mathbf{x}_v\}_{v \in \mathcal{V}}$ and $\{\mathbf{e}_{i,j}\}_{i,j \in \mathcal{E}}$ are textual attributes of nodes and edges, respectively. One typical kind of TAGs is Knowledge Graphs (KGs), where nodes are entities, edges are relations among entities, and text attributes are the names of entities and relations.

\subsection{Graph Neural Networks}
Graph Neural Networks (GNNs) are a kind of deep learning framework to model the graph data. Classical GNNs, e.g., GCN~\cite{ref:gcn}, GAT~\cite{ref:gat}, GraphSAGE~\cite{ref:graphsage}, adopt a message-passing manner to obtain node representations. Formally, each node representation $\mathbf{h}^{(l-1)}_i$ in the $l$-th layer is updated by aggregating the information from neighboring nodes and edges:
\begin{equation}
    \mathbf{h}^{(l)}_i = \textbf{UPD}(\mathbf{h}^{(l-1)}_i, \textbf{AGG}_{j \in \mathcal{N}(i)}\textbf{MSG}(\mathbf{h}^{(l-1)}_i,\mathbf{h}^{(l-1)}_j, \mathbf{e}^{(l-1)}_{i,j})),
\end{equation}
where $\mathcal{N}_{(i)}$ represents the neighbors of node $i$. \textbf{MSG} denotes the message function, which computes the message based on the node, its neighbor, and the edge between them. \textbf{AGG} refers to the aggregation function that combines the received messages using a permutation-invariant method, such as mean, sum, or max. \textbf{UPD} represents the update function, which updates each node's attributes with the aggregated messages.

Subsequently, a readout function, e.g., mean, sum, or max pooling, can be applied to obtain the global-level representation:
\begin{equation}
    \mathbf{h}_G = \textbf{READOUT}_{i \in \mathcal{V}_G}(\mathbf{h}_i^{(L)}).
\end{equation}

In GraphRAG, GNNs can be utilized to obtain representations of graph data for the retrieval phase, as well as to model the retrieved graph structures.

\subsection{Language Models}
Language models (LMs) excel in language understanding and are mainly classified into two types: discriminative and generative. Discriminative models, like BERT~\cite{ref:bert}, RoBERTa~\cite{ref:roberta} and SentenceBERT~\cite{ref:sentence-bert}, focus on estimating the conditional probability \( P(\mathbf{y}|\mathbf{x}) \) and are effective in tasks such as text classification and sentiment analysis. In contrast, generative models, including GPT-3~\cite{ref:gpt3} and GPT-4~\cite{ref:gpt4}, aim to model the joint probability \( P(\mathbf{x},\mathbf{y}) \) for tasks like machine translation and text generation. These generative pre-trained models have significantly advanced the field of natural language processing (NLP) by leveraging massive datasets and billions of parameters, contributing to the rise of Large Language Models (LLMs) with outstanding performance across various tasks.

In the early stages, RAG and GraphRAG focused on improving pre-training techniques for discriminative language models~\cite{ref:bert,ref:roberta,ref:sentence-bert}. Recently, LLMs such as ChatGPT~\cite{ref:chatgpt}, LLaMA~\cite{ref:llama3}, and Qwen2~\cite{ref:qwen2} have shown great potential in language understanding, demonstrating powerful in-context learning capabilities. Subsequently, research on RAG and GraphRAG shifted towards enhancing information retrieval for language models, addressing increasingly complex tasks and mitigating hallucinations, thereby driving rapid advancements in the field. 

\section{Overview of GraphRAG}
GraphRAG is a framework that leverages external structured knowledge graphs to improve contextual understanding of LMs and generate more informed responses, as depicted in Figure~\ref{fig:overview}. The goal of GraphRAG is to retrieve the most relevant knowledge from databases, thereby enhancing the answers of downstream tasks. The process can be defined as
\begin{equation}
    \begin{aligned}
        a^* = \arg\max_{a \in A} p(a|q,\mathcal{G}),
    \end{aligned}
\end{equation}
where $a^*$ is the optimal answer of the query $q$ given the TAG $\mathcal{G}$, and $A$ is the set of possible responses. After that, we jointly model the target distribution $p(a|q,\mathcal{G})$ with a graph retriever $p_\theta(G|q,\mathcal{G})$ and an answer generator $p_\phi (a|q,G)$ where $\theta, \phi$ are learnable parameters, and utilize the total probability formula to decompose $p(a|q,\mathcal{G})$, which can be formulated as
\begin{equation}
    \begin{aligned}
    p(a|q,\mathcal{G}) &= \sum\limits_{G \subseteq \mathcal{G}} p_\phi (a|q,G) p_\theta(G|q,\mathcal{G}) \\
        &\approx p_\phi (a|q,G^*) p_\theta(G^*|q,\mathcal{G}),
    \end{aligned}
    \label{eq:total}
\end{equation}
where $G^*$ is the optimal subgraph. Because the number of candidate subgraphs can grow exponentially with the size of the graph, efficient approximation methods are necessary. The first line of Equation~\ref{eq:total} is thus approximated by the second line. Specifically, a graph retriever is employed to extract the optimal subgraph $G^*$, after which the generator produces the answer based on the retrieved subgraph.

Therefore, in this survey, we decompose the entire process of GraphRAG into three main stages: Graph-Based Indexing, Graph-Guided Retrieval, and Graph-Enhanced Generation. The overall workflow of GraphRAG is illustrated in Figure 2 and detailed introductions of each stage are as follows.

\paragraph{Graph-Based Indexing (G-Indexing)} 
\sloppy
Graph-Based Indexing constitutes the initial phase of GraphRAG, aimed at identifying or constructing a graph database $\mathcal{G}$ that aligns with downstream tasks and establishing indices on it. The graph database can originate from public knowledge graphs~\cite{ref:wikidata,ref:freebase,ref:dbpedia,ref:yago,ref:conceptnet,ref:atomic}, graph data~\cite{ref:tudataset}, or be constructed based on proprietary data sources such as textual~\cite{ref:graphrag,ref:dalk,ref:kgp,ref:hipporag} or other forms of data~\cite{ref:noname1}. The indexing process typically includes mapping node and edge properties, establishing pointers between connected nodes, and organizing data to support fast traversal and retrieval operations. Indexing determines the granularity of the subsequent retrieval stage, playing a crucial role in enhancing query efficiency.

\paragraph{Graph-Guided Retrieval (G-Retrieval)} 
Following graph-based indexing, the graph-guided retrieval phase focuses on extracting pertinent information from the graph database in response to user queries or input. Specifically, given a user query $q$ which is expressed in natural language, the retrieval stage aims to extract the most relevant elements (e.g., entities, triplets, paths, subgraphs) from knowledge graphs, which can be formulated as
\begin{equation}
    \begin{aligned}
        G^* &= \textbf{G-Retriever}(q, \mathcal{G}) \\
        &= \argmax_{G \subseteq \mathcal{R}(\mathcal{G})} \, p_\theta(G|q,\mathcal{G}) \\
        &= \argmax_{G \subseteq \mathcal{R}(\mathcal{G})} \, \textbf{Sim}(q, G),
    \end{aligned}
\end{equation}
where $G^*$ is the optimal retrieved graph elements and $\textbf{Sim}(\cdot,\cdot)$ is a function that measures the semantic similarity between user queries and the graph data. $\mathcal{R(\cdot)}$ represents a function to narrow down the search range of subgraphs, considering the efficiency.

\paragraph{Graph-Enhanced Generation (G-Generation)} 
The graph-enhanced generation phase involves synthesizing meaningful outputs or responses based on the retrieved graph data. This could encompass answering user queries, generating reports, etc. In this stage, a generator takes the query, retrieved graph elements, and an optional prompt as input to generate a response, which can be denoted as
\begin{equation}
    \begin{aligned}
    a^* &= \textbf{G-Generator}(q, G^*) \\
    &= \argmax_{a \in A} p_\phi (a|q,G^*) \\
    &= \argmax_{a \in A} p_\phi (a|\mathcal{F}(q,G^*)),
    \end{aligned}
\end{equation}
where $\mathcal{F}(\cdot,\cdot)$ is a function that converts graph data into a form the generator can process.

\section{Graph-Based Indexing}
\label{sec:indexing}
The construction and indexing of graph databases form the foundation of GraphRAG, where the quality of the graph database directly impacts GraphRAG's performance. In this section, we categorize and summarize the selection or construction of graph data and various indexing methods that have been employed.

\subsection{Graph Data}
Various types of graph data are utilized in GraphRAG for retrieval and generation. Here, we categorize these data into two categories based on their sources, including Open Knowledge Graphs and Self-Constructed Graph Data.

\subsubsection{Open Knowledge Graphs}
Open knowledge graphs refer to graph data sourced from publicly available repositories or databases~\cite{ref:wikidata,ref:freebase,ref:dbpedia,ref:yago}. Using these knowledge graphs could dramatically reduce the time and resources required to develop and maintain. In this survey, we further classify them into two categories according to their scopes, i.e., General Knowledge Graphs and Domain Knowledge Graphs.

\paragraph{(1) General Knowledge Graphs} General knowledge graphs primarily store general, structured knowledge, and typically rely on collective input and updates from a global community, ensuring a comprehensive and continually refreshed repository of information.

Encyclopedic knowledge graphs are a typical type of general knowledge graph, which contains large-scale real-world knowledge collected from human experts and encyclopedias. For example, Wikidata\footnote{https://www.wikidata.org/}~\cite{ref:wikidata} is a free and open knowledge base that stores structured data of its Wikimedia sister projects like Wikipedia, Wikivoyage, Wiktionary, and others. Freebase\footnote{http://www.freebase.be/}~\cite{ref:freebase} is an extensive, collaboratively edited knowledge base that compiles data from various sources, including individual contributions and structured data from databases like Wikipedia. DBpedia\footnote{https://www.dbpedia.org/}~\cite{ref:dbpedia} represents information about millions of entities, including people, places, and things, by leveraging the infoboxes and categories present in Wikipedia articles.
YAGO\footnote{https://yago-knowledge.org/}~\cite{ref:yago} collects knowledge from Wikipedia, WordNet, and GeoNames.

Commonsense knowledge graphs are another type of general knowledge graph. They include abstract commonsense knowledge, such as semantic associations between concepts and causal relationships between events. Typical Commonsense Knowledge Graphs include: ConceptNet\footnote{https://conceptnet.io/}~\cite{ref:conceptnet} is a semantic network built from nodes representing words or phrases connected by edges denoting semantic relationships. ATOMIC~\cite{ref:atomic,ref:atomic2} models the causal relationships between events.

\paragraph{(2) Domain Knowledge Graphs} As discussed in Section~\ref{sec:intro}, domain-specific knowledge graphs are crucial for enhancing LLMs in addressing domain-specific questions. These KGs offer specialized knowledge in particular fields, aiding models in gaining deeper insights and a more comprehensive understanding of complex professional relationships. 
In the biomedical field, CMeKG\footnote{https://cmekg.pcl.ac.cn/} encompasses a wide range of data, including diseases, symptoms, treatments, medications, and relationships between medical concepts. CPubMed-KG\footnote{https://cpubmed.openi.org.cn/graph/wiki} is a medical knowledge database in Chinese, building on the extensive repository of biomedical literature in PubMed. 
In the movie domain, Wiki-Movies~\cite{ref:wikimovie} extracts structured information from Wikipedia articles related to films, compiling data about movies, actors, directors, genres, and other relevant details into a structured format. Additionally,~\citett{ref:graphcot} construct a dataset named GR-Bench, which includes five domain knowledge graphs spanning academic, E-commerce, literature, healthcare, and legal fields. 
Furthermore,~\citett{ref:g-retriever} convert triplet-format and JSON files from ExplaGraphs and SceneGraphs into a standard graph format and selects questions requiring 2-hop reasoning from WebQSP to create the universal graph-format dataset GraphQA for evaluating GraphRAG systems.

\subsubsection{Self-Constructed Graph Data}
Self-Constructed Graph Data facilitates the customization and integration of proprietary or domain-specific knowledge into the retrieval process. For downstream tasks that do not inherently involve graph data, researchers often propose constructing a graph from multiple sources (e.g., documents, tables, and other databases) and leveraging GraphRAG to enhance task performance. Generally, these self-constructed graphs are closely tied to the specific design of the method, distinguishing them from the open-domain graph data previously mentioned.

To model the structural relationships between the documents,~\citett{ref:atlantic} propose to construct a heterogeneous document graph capturing multiple document-level relations, including co-citation, co-topic, co-venue, etc.~\citett{ref:gnn-net,ref:kgp} establish relationship between passages according to shared keywords. To capture the relations between entities in documents,~\citett{ref:noname5,ref:graphrag,ref:hipporag,ref:dalk} utilize the named entity recognition tools to extract entities from documents and language models to further extract relations between entities, where the retrieved entities and relations then form a knowledge graph. There are also some mapping methods for downstream tasks that need to be designed based on the characteristics of the task itself. For example, to solve the patent phrase similarity inference task,~\citett{ref:ra-sim} convert the patent database into a patent-phrase graph. Connections between patent nodes and phrase nodes are established if the phrases appear in the patents, while connections between patent nodes are based on citation relations. Targeting customer service technical support scenarios,~\citett{ref:noname1} propose to model historical issues into a KG, which transforms the issues into tree representations to maintain the intra-issue relations, and utilize semantic similarities and a threshold to preserve inter-issue relations.

\subsection{Indexing}
Graph-Based Indexing plays a crucial role in enhancing the efficiency and speed of query operations on graph databases, directly influencing subsequent retrieval methods and granularity. Common graph-based indexing methods include graph indexing, text indexing, and vector indexing.

\begin{figure}
    \centering
    \includegraphics[width=1\linewidth]{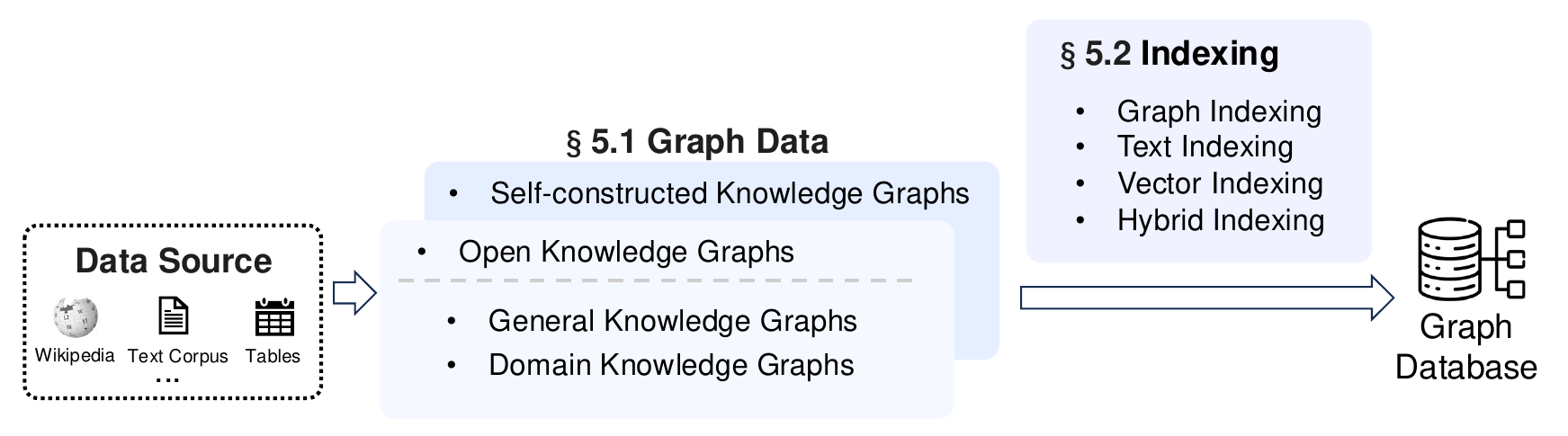}
    \caption{The overview of graph-based indexing.}
    \label{fig:retrieval}
\end{figure}

\subsubsection{Graph Indexing}
Graph indexing represents the most commonly used approach, preserving the entire structure of the graph. This method ensures that for any given node, all its edges and neighboring nodes are easily accessible. During subsequent retrieval stages, classic graph search algorithms such as BFS and Shortest Path Algorithms can be employed to facilitate retrieval tasks~\cite{ref:graphcot,ref:rog,ref:tog,ref:tog2,ref:hykge,ref:qa-gnn,ref:grapeqa}.

\subsubsection{Text Indexing}
Text indexing involves converting graph data into textual descriptions to optimize retrieval processes. These descriptions are stored in a text corpus, where various text-based retrieval techniques, such as sparse retrieval and dense retrieval, can be applied. Some approaches transform knowledge graphs into human-readable text using predefined rules or templates. For instance, \citett{ref:noname2,ref:mvp-tuning,ref:unioqa} use predefined templates to convert each triple in knowledge graphs into natural language, while \citett{ref:decaf} merge triplets with the same head entity into passages. Additionally, some methods convert subgraph-level information into textual descriptions. For example, \citett{ref:graphrag} perform community detection on the graph and generate summaries for each community using LLMs.

\subsubsection{Vector Indexing}
Vector indexing transforms graph data into vector representations to enhance retrieval efficiency, facilitating rapid retrieval and effective query processing. For example, entity linking can be seamlessly applied through query embeddings, and efficient vector search algorithms such as Locality Sensitive Hashing (LSH)~\cite{ref:lsh} can be utilized. G-Retriever~\cite{ref:g-retriever} employs language models to encode textual information associated with each node and edge within the graph, while GRAG~\cite{ref:grag} uses language models to convert $k$-hop ego networks into graph embeddings, thereby better preserving structural information.

\subsubsection{Hybrid Indexing}
Each of the above three indexing methods offers distinct advantages: graph indexing facilitates easy access to structural information, text indexing simplifies retrieval of textual content, and vector indexing enables quick and efficient searches. Therefore, in practical applications, a hybrid approach combining these indexing methods is often preferred over relying solely on one. For instance, HybridRAG~\cite{ref:hybridrag} retrieves both vector and graph data simultaneously to enhance the content retrieved. While EWEK-QA~\cite{ref:ewek-qa} uses both text and knowledge graphs.

\begin{figure}
    \centering
    \includegraphics[width=1\linewidth]{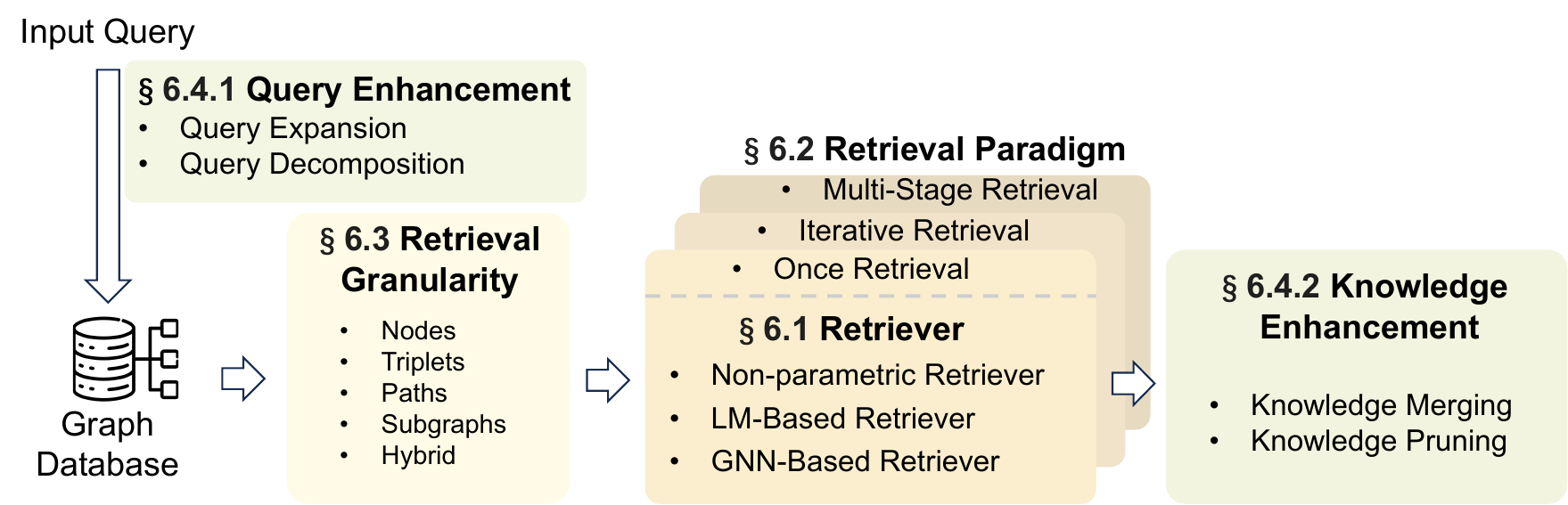}
    \caption{The general architectures of graph-based retrieval.}
    \label{fig:retrieval}
\end{figure}

\section{Graph-Guided Retrieval}\label{sec:retrieval}
In GraphRAG, the retrieval process is crucial for ensuring the quality and relevance of generated outputs by extracting pertinent and high-quality graph data from external graph databases. However, retrieving graph data presents two significant challenges: (1) \emph{Explosive Candidate Subgraphs}: As the graph size increases, the number of candidate subgraphs grows exponentially, requiring heuristic search algorithms to efficiently explore and retrieve relevant subgraphs. (2) \emph{Insufficient Similarity Measurement}: Accurately measuring similarity between textual queries and graph data necessitates the development of algorithms capable of understanding both textual and structural information.

Considerable efforts have previously been dedicated to optimizing the retrieval process to address the above challenges. This survey focuses on examining various aspects of the retrieval process within GraphRAG, including the selection of the retriever, retrieval paradigm, retrieval granularity, and effective enhancement techniques. The general architectures of Graph-Guided Retrieval are depicted in Figure~\ref{fig:retrieval}.

\subsection{Retriever}
In GraphRAG, various retrievers possess unique strengths for addressing different aspects of retrieval tasks. We categorize retrievers into three types based on their underlying models: Non-parametric Retriever, LM-based Retriever, and GNN-based Retriever. It is important to note that models used in pre-processing steps, such as query encoding and entity linking, are not considered here, as these models vary across different methods and are not the primary focus of this paper.

\subsubsection{Non-parametric Retriever}
Non-parametric retrievers, based on heuristic rules or traditional graph search algorithms, do not rely on deep-learning models, thereby achieving high retrieval efficiency. For instance, \citett{ref:qa-gnn,ref:grapeqa} retrieve $k$-hop paths containing the topic entities of each question-choice pair. G-Retriever~\cite{ref:g-retriever} enhances the conventional Prize-Collecting Steiner Tree (PCST) algorithm by incorporating edge prices and optimizing relevant subgraph extraction. \citett{ref:noname5,ref:gnn-rag} first extract entities mentioned in the query and then retrieve the shortest path related to these entities. These methods often involve an entity linking pre-processing step to identify nodes in the graph before retrieval. 

\subsubsection{LM-based Retriever}
LMs serve as effective retrievers in GraphRAG due to their strong natural language understanding capabilities. These models excel in processing and interpreting diverse natural language queries, making them versatile for a wide range of retrieval tasks within graph-based frameworks. We primarily categorized LMs into two types: discriminative and generative language models. Subgraph Retriever~\cite{ref:sr} trains RoBERTa~\cite{ref:roberta} as the retriever, which expands from the topic entity and retrieves the relevant paths in a sequential decision process. KG-GPT~\cite{ref:kg-gpt} adopts LLMs to generate the set of top-$K$ relevant relations of the specific entity.~\citett{ref:noname7} utilize fine-tuned GPT-2 to generate reasoning paths. StructGPT~\cite{ref:structgpt} utilizes LLMs to automatically invoke several pre-defined functions, by which relevant information can be retrieved and combined to assist further reasoning.

\subsubsection{GNN-based Retriever}
GNNs are adept at understanding and leveraging complex graph structures. GNN-based retrievers typically encode graph data and subsequently score different retrieval granularities based on their similarity to the query. For example, GNN-RAG~\cite{ref:gnn-rag} first encodes the graph, assigns a score to each entity, and retrieves entities relevant to the query based on a threshold. EtD~\cite{ref:etd} iterates multiple times to retrieve relevant paths. During each iteration, it first uses LLaMA2~\cite{ref:llama} to select edges connecting the current node, then employs GNNs to obtain embeddings of the new layer of nodes for the next round of LLM selection.

\subsubsection{Discussion} During the retrieval process, non-parametric retrievers exhibit good retrieval efficiency, but they may suffer from inaccurate retrieval due to a lack of training on downstream tasks. Meanwhile, although LM-based retrievers and GNN-based retrievers offer higher retrieval accuracy, they require significant computational overhead. Considering this complementarity, many methods propose hybrid retrieval approaches to improve both retrieval efficiency and accuracy. Many approaches adopt a multi-stage retrieval strategy, employing different models at each stage. For example, RoG~\cite{ref:rog} first utilizes LLMs to generate planning paths and then extracts paths satisfying the planning paths from knowledge graphs. GenTKGQA~\cite{ref:gentkgqa} infers crucial relations and constraints from the query using LLMs and extracts triplets according to these constraints.

\subsection{Retrieval Paradigm}
Within GraphRAG, different retrieval paradigms, including once retrieval, iterative retrieval, and multi-stage retrieval, play crucial roles in improving the relevance and depth of the retrieved information. Once retrieval aims to gather all pertinent information in a single operation. Iterative retrieval conducts further searches based on previously retrieved information, progressively narrowing down to the most relevant results. Here we further divide iterative retrieval into adaptive retrieval and non-adaptive retrieval, with the only difference lying in whether the stopping of the retrieval is determined by the model. Another retrieval paradigm is multi-stage retrieval, where retrieval is divided into multiple stages. Different types of retrievers may be employed at each stage for more precise and diversified search results. Below, we will provide a detailed introduction to these types of retrieval paradigms.

\subsubsection{Once Retrieval}
Once retrieval aims to retrieve all the relevant information in a single query. One category of approaches~\cite{ref:hipporag,ref:noname2,ref:grag} utilize embedding similarities to retrieve the most relevant pieces of information. Another category of methods design pre-defined rules or patterns to directly extract specific structured information such as triplets, paths or subgraphs from graph databases. For example, G-Retriever~\cite{ref:g-retriever} utilizes an extended PCST algorithm to retrieve the most relevant subgraph. KagNet~\cite{ref:kagnet} extracts paths between all pairs of topic entities with lengths not exceeding $k$.~\citett{ref:qa-gnn,ref:grapeqa} extract the subgraph that contains all topic entities along with their $2$-hop neighbors.

Furthermore, in this subsection, we also include some multiple retrieval methods that involve decoupled and independent retrievals, allowing them to be computed in parallel and executed only once. For example, \citett{ref:rog,ref:temple-mqa} first instruct LLMs to generate multiple reasoning paths and then use a BFS retriever to sequentially search for subgraphs in the knowledge graphs that match each path. KG-GPT~\cite{ref:kg-gpt} decomposes the original query into several sub-queries, retrieving relevant information for each sub-query in a single retrieval process.

\subsubsection{Iterative Retrieval} 
In iterative retrieval, multiple retrieval steps are employed, with subsequent searches depending on the results of prior retrievals. These methods aim to deepen the understanding or completeness of the retrieved information over successive iterations. In this survey, we further classify iterative retrieval into two categories: (1) non-adaptive and (2) adaptive retrieval. We provide a detailed summary of these two categories of methods below.

\paragraph{(1) Non-Adaptive Retrieval} 
Non-adaptive methods typically follow a fixed sequence of retrieval, and the termination of retrieval is determined by setting a maximum time or a threshold. For example, PullNet~\cite{ref:pullnet} retrieves problem-relevant subgraphs through $T$ iterations. In each iteration, the paper designs a retrieval rule to select a subset of retrieved entities, and then expands these entities by searching relevant edges in the knowledge graph. In each iteration, KGP~\cite{ref:kgp} first selects seed nodes based on the similarity between the context and the nodes in the graph. It then uses LLMs to summarize and update the context of the neighboring nodes of the seed nodes, which is utilized in the subsequent iteration.

\paragraph{(2) Adaptive Retrieval}
One distinctive characteristic of adaptive retrieval is to let models autonomously determine the optimal moments to finish the retrieval activities. For instance,~\cite{ref:rra,ref:kn} leverage an LM for hop prediction, which serves as an indicator to end the retrieval. There is also a group of researchers who utilize model-generated special tokens or texts as termination signals for the retrieval process. For example, ToG~\cite{ref:tog,ref:tog2} prompts the LLM agent to explore the multiple possible reasoning paths until the LLM determines the question can be answered based on the current reasoning path.~\cite{ref:sr} trains a RoBERTa to expand a path from each topic entity. In the process, a virtual relation named as ``[END]'' is introduced to terminate the retrieval process.

Another common approach involves treating the large model as an agent, enabling it to directly generate answers to questions to signal the end of iteration. For instance,~\cite{ref:graphcot,ref:structgpt,ref:kg-agent,ref:knowledgpt,ref:oda} propose LLM-based agents to reason on graphs. These agents could autonomously determine the information for retrieval, invoke the pre-defined retrieval tools, and cease the retrieval process based on the retrieved information.

\subsubsection{Multi-Stage Retrieval} 
Multi-stage retrieval divides the retrieval process linearly into multiple stages, with additional steps such as retrieval enhancement, and even generation processes occurring between these stages. In multi-stage retrieval, different stages may employ various types of retrievers, which enables the system to incorporate various retrieval techniques tailored to different aspects of the query. For example,~\citett{ref:rok} first utilize a non-parametric retriever to extract $n$-hop paths of entities in the query's reasoning chain, then after a pruning stage, it further retrieves the one-hop neighbors of the entities in the pruned subgraph. OpenCSR~\cite{ref:opencsr} divides the retrieval process into two stages. In the first stage, it retrieves all $1$-hop neighbors of the topic entity. In the second stage, it compares the similarity between these neighbor nodes and other nodes, selecting the top-$k$ nodes with the highest similarity for retrieval. GNN-RAG~\cite{ref:gnn-rag} first employs GNNs to retrieve the top-$k$ nodes most likely to be the answer. Subsequently, it retrieves all shortest paths between query entities and answer entities pairwise.

\subsubsection{Discussion} In GraphRAG, once retrieval typically exhibits lower complexity and shorter response times, making it suitable for scenarios requiring real-time responsiveness. In contrast, iterative retrieval often involves higher time complexity, especially when employing LLMs as retrievers, potentially leading to longer processing times. However, this approach can yield higher retrieval accuracy by iteratively refining retrieved information and generating responses. Therefore, the choice of retrieval paradigm should balance accuracy and time complexity based on specific use cases and requirements. 

\subsection{Retrieval Granularity}
According to different task scenarios and indexing types, researchers design distinct retrieval granularities (i.e., the form of related knowledge retrieved from graph data), which can be divided into nodes, triplets, paths, and subgraphs. Each retrieval granularity has its own advantages, making it suitable for different practical scenarios. We will introduce the details of these granularities in the following sections.

\subsubsection{Nodes}
Nodes allow for precise retrieval focused on individual elements within the graph, which is ideal for targeted queries and specific information extraction. In general, for knowledge graphs, nodes refer to entities. For other types of text attribute graphs, nodes may include textual information that describes the node's attributes. By retrieving nodes within the graph, GraphRAG systems could provide detailed insights into their attributes, relationships, and contextual information. For example,~\citett{ref:atlantic,ref:gnn-net,ref:kgp} construct document graphs and retrieves relevant passage nodes.~\citett{ref:etd,ref:pullnet,ref:hipporag} retrieve entities from constructed knowledge graphs.

\subsubsection{Triplets}
Generally, triplets consist of entities and their relationships in the form of subject-predicate-object tuples, providing a structured representation of relational data within a graph. The structured format of triplets allows for clear and organized data retrieval, making it advantageous in scenarios where understanding relationships and contextual relevance between entities is critical.~\citett{ref:kg-rank} retrieve triplets containing topic entities as relevant information.~\citett{ref:mvp-tuning, ref:noname2,ref:unioqa} first convert each triplet of graph data into textual sentences using predefined templates and subsequently adopt a text retriever to extract relevant triplets. However, directly retrieving triplets from graph data may still lack contextual breadth and depth, thus being unable to capture indirect relationships or reasoning chains. To address this challenge,~\citett{ref:keqing} propose to generate the logical chains based on the original question, and retrieve the relevant triplets of each logical chain.

\subsubsection{Paths}
The retrieval of path-granularity data can be seen as capturing sequences of relationships between entities, enhancing contextual understanding and reasoning capabilities. In GraphRAG, retrieving paths offers distinct advantages due to their ability to capture complex relationships and contextual dependencies within a graph.

However, path retrieval can be challenging due to the exponential growth in possible paths as graph size increases, which escalates computational complexity. To address this, some methods retrieve relevant paths based on pre-defined rules. For example,~\citett{ref:rok,ref:ecpr} first select entity pairs in the query and then traverse to find all the paths between them within $n$-hop. HyKGE~\cite{ref:hykge} first defines three types of paths: path, co-ancestor chain, and co-occurrence chain, and then utilizes corresponding rules to retrieve each of these three types of paths. In addition, some methods utilize models to perform path searching on graphs. ToG~\cite{ref:tog,ref:tog2} proposes to prompt the LLM agent to perform the beam search on KGs and find multiple possible reasoning paths that help answer the question.~\citett{ref:rog,ref:rra,ref:kn} first utilizes the model to generate faithful reasoning plans and then retrieves relevant paths based on these plans. GNN-RAG~\cite{ref:gnn-rag} first identifies the entities in the question. Subsequently, all paths between entities that satisfy a certain length relationship are extracted.

\subsubsection{Subgraphs}
Retrieving subgraphs offers significant advantages due to its ability to capture comprehensive relational contexts within a graph. This granularity enables GraphRAG to extract and analyze complex patterns, sequences, and dependencies embedded within larger structures, facilitating deeper insights and a more nuanced understanding of semantic connections.

To ensure both information completeness and retrieval efficiency, some methods propose an initial rule-based approach to retrieve candidate subgraphs, which are subsequently refined or processed further.~\citett{ref:ra-sim} retrieve the ego graph of the patent phrase from the self-constructed patent-phrase graph.~\citett{ref:qa-gnn,ref:feng,ref:grapeqa} first select the topic entities and their two-hop neighbors as the node set, and then choose the edges with head and tail entities both in the node set to form the subgraph. Besides, there are also some embedding-based subgraph retrieval methods. For example,~\citett{ref:grag} first encode all the $k$-hop ego networks from the graph database, then retrieve subgraphs related to the query based on the similarities between embeddings.~\citett{ref:mindmap,ref:dalk} extract two types of graphs, including Path evidence subgraphs and Neighbor evidence subgraphs, based on pre-defined rules. OpenCSR~\cite{ref:opencsr} starts from a few initial seed nodes and gradually expands to new nodes, eventually forming a subgraph.

In addition to the aforementioned direct subgraph retrieval methods, some works propose first retrieving relevant paths and then constructing related subgraphs from them. For instance, \citett{ref:sr} train a RoBERTa model to identify multiple reasoning paths through a sequential decision process, subsequently merging identical entities from different paths to induce a final subgraph.

\subsubsection{Hybrid Granularties}
Considering the advantages and disadvantages of various retrieval granularities mentioned above, some researchers propose using hybrid granularities, that is, retrieving relevant information of multiple granularities from graph data. This type of granularity enhances the system's ability to capture both detailed relationships and broader contextual understanding, thus reducing noise while improving the relevance of the retrieved data. Various previous works propose to utilize LLM agents to retrieve complex hybrid information.~\citett{ref:graphcot,ref:structgpt,ref:kg-agent,ref:knowledgpt,ref:oda} propose to adopt LLM-based agents for adaptively selecting nodes, triplets, paths, and subgraphs.

\subsubsection{Discussion} (1) In real applications, there are no clear boundaries between these retrieval granularities, as subgraphs can be composed of multiple paths, and paths can be formed by several triplets. (2) Various granularities such as nodes, triplets, paths, and subgraphs offer distinct advantages in the GraphRAG process. Balancing between retrieval content and efficiency is crucial when selecting the granularity, depending on the specific context of the task. For straightforward queries or when efficiency is paramount, finer granularities such as entities or triplets may be preferred to optimize retrieval speed and relevance. In contrast, complex scenarios often benefit from a hybrid approach that combines multiple granularities. This approach ensures a more comprehensive understanding of the graph structure and relationships, enhancing the depth and accuracy of the generated responses. Thus, GraphRAG's flexibility in granularity selection allows it to adapt effectively to diverse information retrieval needs across various domains.

\subsection{Retrieval Enhancement}
To ensure high retrieval quality, researchers propose techniques to enhance both user queries and the knowledge retrieved. In this paper, we categorize query enhancement into query expansion and query decomposition, and knowledge enhancement into merging and pruning. These strategies collectively optimize the retrieval process. Although other techniques such as query rewriting~\cite{ref:rewrite1,ref:rewrite2,ref:rewrite3,ref:rewrite4} are commonly used in RAG, they are less frequently applied in GraphRAG. We do not delve into these methods, despite their potential adaptation for GraphRAG.

\subsubsection{Query Enhancement} 
Strategies applied to queries typically involve pre-processing techniques that enrich the information for better retrieval. This may include query expansion and query decomposition. 

\paragraph{(1) Query Expansion} Due to the generally short length of queries and their limited information content, query expansion aims to improve search results by supplementing or refining the original query with additional relevant terms or concepts.~\citett{ref:rog} generate relation paths grounded by KGs with LLMs to enhance the retrieval query.~\citett{ref:temple-mqa} adopt SPARQL to get all the aliases of the query entities from Wikidata to augment the retrieval queries, which capture lexical variations of the same entity.~\citett{ref:mvp-tuning} propose a consensus-view knowledge retrieval method to improve retrieval accuracy, which first discover semantically relevant queries, and then re-weight the original query terms to enhance the retrieval performance. HyKGE~\cite{ref:hykge} utilizes a large model to generate the hypothesis output of the question, concatenating the hypothesis output with the query as input to the retriever. Golden-Retriever~\cite{ref:golden} first recognizes the jargon in the query and then retrieves explanations of the jargon as a supplement to the query.

\paragraph{(2) Query Decomposition} Query decomposition techniques break down or decompose the original user query into smaller, more specific sub-queries. Each sub-query typically focuses on a particular aspect or component of the original query, which successfully alleviates the complexity and ambiguity of language queries. For instance,~\cite{ref:kg-gpt,ref:lark} breaks down the primary question into sub-sentences, each representing a distinct relation, and sequentially retrieves the pertinent triplets for each sub-sentence.

\subsubsection{Knowledge Enhancement}
After retrieving initial results, knowledge enhancement strategies are employed to refine and improve the retriever's results. This phase often involves knowledge merging and knowledge pruning processes to present the most pertinent information prominently. These techniques aim to ensure that the final set of retrieved results is not only comprehensive but also highly relevant to the user’s information needs. 

\paragraph{(1) Knowledge Merging} Knowledge merging retrieved information enables compression and aggregation of information, which assists in obtaining a more comprehensive view by consolidating relevant details from multiple sources. This approach not only enhances the completeness and coherence of the information but also mitigates issues related to input length constraints in models. KnowledgeNavigator~\cite{ref:kn} merges nodes and condenses the retrieved
sub-graph through triple aggregation to enhance the
reasoning efficiency. In Subgraph Retrieval~\cite{ref:sr}, after retrieving top-$k$ paths from each topic entity to form a single subgraph, researchers propose to merge the same entities from different subgraphs to form the final subgraph.~\citett{ref:mindmap,ref:dalk} merge retrieved subgraphs based on relations, combining head entities and tail entities that satisfy the same relation into two distinct entity sets, ultimately forming a relation paths.

\paragraph{(2) Knowledge Pruning} Knowledge pruning involves filtering out less relevant or redundant retrieved information to refine the results. Previous approaches for pruning encompass two main categories: (re)-ranking-based approaches and LLM-based approaches. (Re)-ranking methods involve the reordering or prioritization of retrieved information using tailored metrics or criteria. 

One line of methods introduces stronger models for reranking. For example,~\citett{ref:noname2} concatenate each retrieved triplet with the question-choice pair, and adopt a pre-trained cross-encoder~\cite{ref:sentence-bert} to re-rank the retrieved triplets.~\citett{ref:hykge} utilize the FlagEmbedding to encode the text to re-rank top-k documents returned by embedding model ``bge\_reranker\_large''.~\citett{ref:kelp} train a PLM to 

Another category utilizes the similarity between queries and retrieved information for ranking. For instance,~\citett{ref:temple-mqa} re-rank the candidate subgraphs based on the similarity for both relation and fine-grained concept between subgraphs and the query.~\citett{ref:grapeqa} first cluster the 2-hop neighbors and then delete the cluster with the lowest similarity score with the input query.~\citett{ref:qa-gnn} prune the retrieved subgraph according to the relevance score between the question context and the KG entity nodes calculated by a pre-trained language model.~\citett{ref:rok,ref:unikgqa,ref:hipporag,ref:rasr} adopt Personalized PageRank algorithm to rank the retrieved candidate information for further filtering.~\citett{ref:kelp} trains a PLM to score the similarity between the retrieved information and the query, and rerank the retrieved paths based on the similarity score. G-G-E~\cite{ref:gge} first divides the retrieved subgraph into several smaller subgraphs, then compares the similarity between each smaller subgraph and the query. Subgraphs with low similarity are removed, and the remaining smaller subgraphs are merged into a larger subgraph. 

Additionally, a third category of methods proposes new metrics for reranking. For example, ~\citett{ref:atlantic} propose a metric that measures both the impact and recency of the retrieved text chunks. KagNet~\cite{ref:kagnet} decomposes the retrieved paths into triplets and reranks the paths  based on the confidence score measured by the knowledge graph embedding (KGE) techniques. LLM-based methods excel in capturing complex linguistic patterns and semantic nuances, which enhances their ability to rank search results or generate responses more accurately. To avoid introducing noisy information, ~\citett{ref:rok,ref:kg-gpt} propose to prune the irrelevant graph data by calling LLMs to check.

\begin{figure}
    \centering
    \includegraphics[width=1\linewidth]{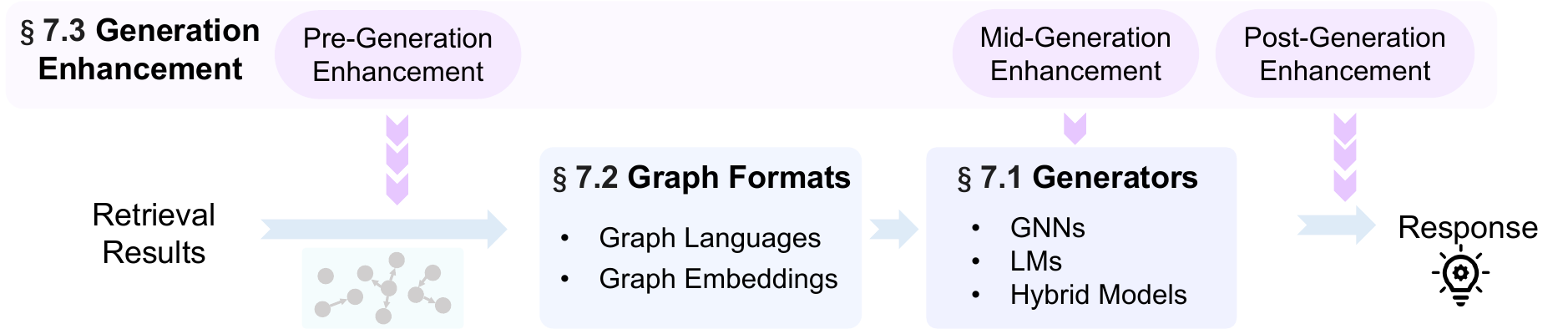}
    \caption{The overview of graph-enhanced generation.}
    \label{fig:generation}
\end{figure}

\section{Graph-Enhanced Generation}
\label{sec:generation}
The generation stage is another crucial step in GraphRAG, aimed at integrating the retrieved graph data with the query to enhance response quality. In this stage, suitable generation models must be selected based on the downstream tasks. The retrieved graph data is then transformed into formats compatible with the generators. The generator takes both the query and the transformed graph data as inputs to produce the final response. Beyond these fundamental processes, generative enhancement techniques can further improve the output by intensifying the interaction between the query and the graph data and enriching the content generation itself. The organization of this section and the overview of graph-enhanced generation are depicted in Figure~\ref{fig:generation}.

\subsection{Generators}
The selection of generators often depends on the type of downstream task at hand. For discriminative tasks (e.g., multi-choice question answering) or generative tasks that can be formulated as discriminative tasks (e.g., KBQA), one can utilize GNNs or discriminative language models to learn representations of the data. These representations can then be mapped to the logits associated with different answer options to provide responses. Alternatively, generative language models can be employed to directly generate answers. For generative tasks, however, the use of GNNs and discriminative language models alone is insufficient. These tasks require the generation of text, which necessitates the deployment of decoders.

\subsubsection{GNNs}
Due to the powerful representational capabilities of GNNs for graph data, they are particularly effective for discriminative tasks. GNNs can directly encode graph data, capturing complex relationships and node features inherent in the graph structure. This encoding is then processed through a Multi-Layer Perceptron (MLP) to generate predictive outcomes. These approaches primarily utilize classical GNN models (e.g., GCN~\cite{ref:gcn}, GAT~\cite{ref:gat}, GraphSAGE~\cite{ref:graphsage}, and Graph Transformers~\cite{ref:graphtransformer}), either in their original form or modified to better align with downstream tasks. For example, HamQA~\cite{ref:hamqa} designs a hyperbolic GNN to learn the representations of retrieved graph data, which learns from the mutual hierarchical information between query and graphs.~\citett{ref:graft} compute PageRank scores for neighboring nodes and aggregates them weighted by these scores, during message-passing. This approach enhances the central node's ability to assimilate information from its most relevant neighboring nodes.~\citett{ref:rearev} decode the query into several vectors (instructions), and enhances instruction decoding and execution for effective reasoning by emulating breadth-first search (BFS) with GNNs to improve instruction execution and using adaptive reasoning to update the instructions with KG-aware information.

\subsubsection{LMs}
LMs possess strong capabilities in text understanding, which also allows them to function as generators. In the context of integrating LMs with graph data, it is necessary to first convert the retrieved graph data into specific graph formats. This conversion process ensures that the structured information is effectively understood and utilized by the LMs. These formats, which will be elaborated on in Section~\ref{sec:graphformats}, are crucial for preserving the relational and hierarchical structure of the graph data, thereby enhancing the model's ability to interpret complex data types. Once the graph data is formatted, it is then combined with a query and fed into an LM.

For encoder-only models, such as BERT~\cite{ref:bert} and RoBERTa~\cite{ref:roberta}, their primary use is in discriminative tasks. Similar to GNNs, these models first encode the input text and then utilize MLPs to map it to the answer space~\cite{ref:unikgqa,ref:mvp-tuning,ref:noname2}. On the other hand, encoder-decoder and decoder-only models, such as T5~\cite{ref:t5}, GPT-4~\cite{ref:gpt4}, and LLaMA~\cite{ref:llama3}, are adept at both discriminative and generative tasks. These models excel in text understanding, generation, and reasoning, allowing them to process textual inputs directly and generate textual responses~\cite{ref:graphrag,ref:rok,ref:rog,ref:graphcot,ref:tog,ref:gnn-rag,ref:keqing,ref:hykge}.

\subsubsection{Hybrid Models}
Considering the strengths of GNNs at representing the structure of graph data, and the robust understanding of text demonstrated by LMs, many studies are exploring the integration of these two technologies to generate coherent responses. This paper categorizes the hybrid generative approaches into two distinct types: cascaded paradigm and parallel paradigm.

\paragraph{(1) Cascaded Paradigm} In the cascaded approaches, the process involves a sequential interaction where the output from one model serves as the input for the next. Specifically, the GNN processes the graph data first, encapsulating its structural and relational information into a form that the LM can understand. Subsequently, this transformed data is fed into the LM, which then generates the final text-based response. These methods leverage the strengths of each model in a step-wise fashion, ensuring detailed attention to both structural and textual data. 

In these methods, prompt tuning~\cite{ref:p-tuning,ref:p-tuning2,ref:prefix-tuning,ref:prompttuning} is a typical approach, where GNNs are commonly employed to encode the retrieved graph data. The encoded graph data is subsequently pre-pended as a prefix to the input text embeddings of an LM. The GNN is then optimized through downstream tasks to produce enhanced encodings of the graph data~\cite{ref:g-retriever,ref:grag,ref:graphtranslator,ref:gentkgqa}.

\paragraph{(2) Parallel Paradigm} On the other hand, the parallel approach operates by concurrently utilizing the capabilities of both the GNN and the LLM. In this setup, both models receive the initial inputs simultaneously and work in tandem to process different facets of the same data. The outputs are then merged, often through another model or a set of rules, to produce a unified response that integrates insights from both the graphical structure and the textual content. 

In the parallel paradigm, a typical approach involves separately encoding inputs using both GNNs and LMs, followed by integrating these two representations, or directly integrating their output responses. For instance,~\citett{ref:safe} aggregate predictions from GNNs and LMs by weighted summation to obtain the final answer.~\citett{ref:kagnet,ref:kg-r3} integrate the graph representations derived from GNNs and the text representations generated by LMs using attention mechanisms. \citett{ref:qa-gnn,ref:atlantic,ref:grapeqa} directly concatenate graph representations with text representations.

Another approach involves designing dedicated modules that integrate GNNs with LMs, enabling the resulting representations to encapsulate both structural and textual information. For instance, \citet{ref:greaselm} introduce a module called the GreaseLM Layer, which incorporates both GNN and LM layers. At each layer, this module integrates textual and graph representations using a two-layer MLP before passing them to the next layer. Similarly, ENGINE~\cite{ref:engine} proposes G-Ladders, which combine LMs and GNNs through a side structure, enhancing node representations for downstream tasks.

\paragraph{Discussion} Hybrid models that harness both the representation capabilities of GNNs for graph data and LMs for text data hold promising applications. However, effectively integrating information from these two modalities remains a significant challenge.

\subsection{Graph Formats}
\label{sec:graphformats}
\begin{figure*}
    \centering
    \setlength{\abovecaptionskip}{0.15cm}
    \includegraphics[scale=0.3]{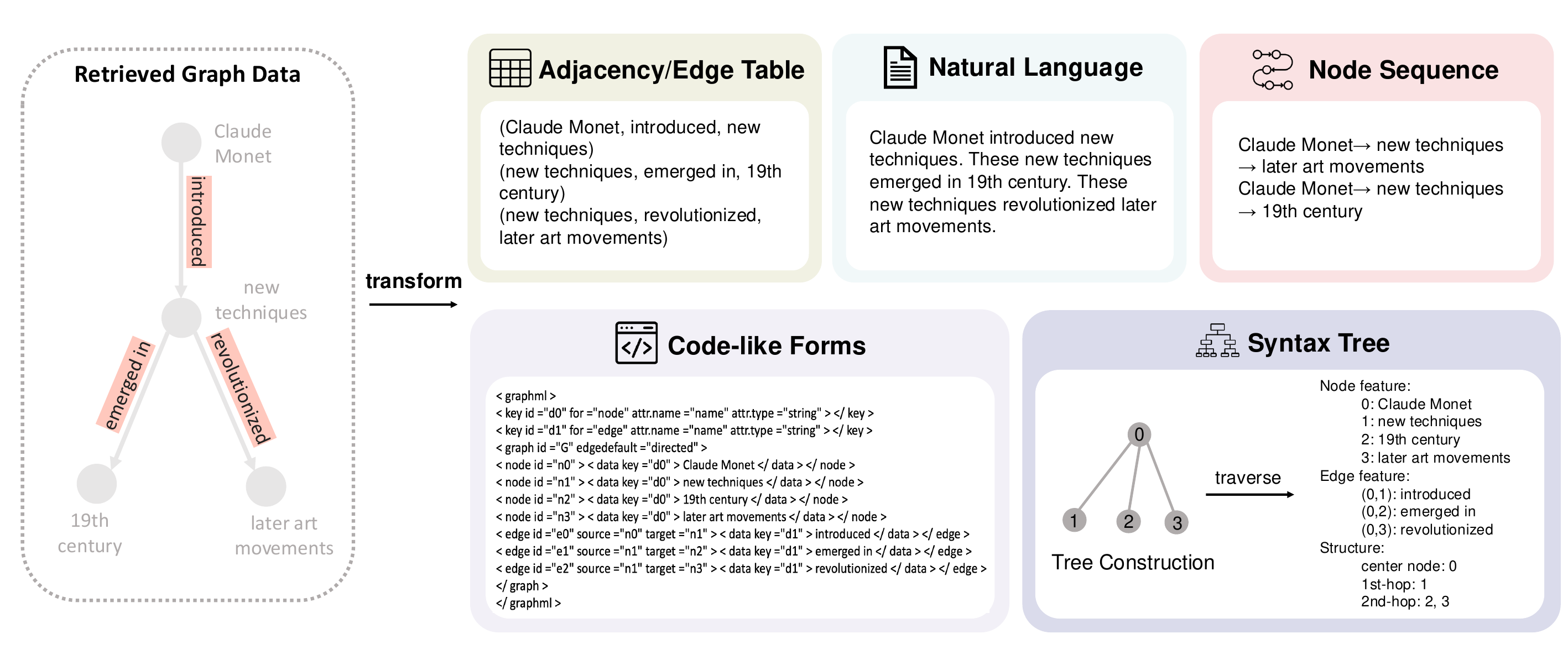}
    \caption{Illustration of the graph languages. Given the retrieved subgraph on the left part, we show how to transform it into adjacency/edge table, natural language, node sequence, code-like forms and syntax trees to adapt the input form requirements of different generators.}
    \label{fig:4}
\end{figure*}

When using GNNs as generators, the graph data can be directly encoded. However, when utilizing LMs as generators, the non-Euclidean nature of graph data poses a challenge, as it cannot be directly combined with textual data for input into the LMs. To address this, graph translators are employed to convert the graph data into a format compatible with LMs. This conversion enhances the generative capabilities of LMs by enabling them to effectively process and utilize structured graph information. In this survey, we summarize two distinct graph formats: graph languages and graph embeddings. We illustrate this process with an example in Figure~\ref{fig:4}, with detailed introductions provided below.

\subsubsection{Graph Languages}
A graph description language is a formalized system of notation that is specifically crafted to characterize and represent graph data. It prescribes a uniform syntax and semantic framework that describes the components and interconnections within a graph. Through these languages, users can consistently generate, manipulate, and interpret graph data in a comprehensible format to machines. They enable the definition of graph architectures, the specification of attributes for nodes and edges, and the implementation of operations and queries on graph structures. Next, we will introduce five types of graph languages separately: Adjacency / Edge Table, Natural Language, Codes, Syntax Tree, and Node Sequence. 

\paragraph{(1) Adjacency / Edge Table} The adjacency table and the edge table are widely used methods for describing graph structures~\cite{ref:talk,ref:gpt4graph,ref:cansolve,ref:noname6}. The adjacency table enumerates the immediate neighbors of each vertex, offering a compact way to represent connections in sparse graphs. For example, KG-GPT~\cite{ref:kg-gpt} linearizes the triples in the retrieved subgraph, which are then concatenated and fed into the LLMs. Conversely, the edge table details all the edges within the graph, providing a straightforward representation that is particularly useful for processing and analyzing graphs in a linear format. Both two methods are brief, easy to understand, and intuitive.

\paragraph{(2) Natural Language} Given that user queries are typically presented in natural language, and considering the outstanding natural language comprehension capabilities of LMs, it becomes a compelling approach to describe the retrieved graph data using natural language. By translating graph data into descriptive, easily comprehensible language, LMs can bridge the gap between raw data representation and user-friendly information, facilitating more effective interactions with data-driven applications. For example, some researchers~\cite{ref:noname2,ref:mvp-tuning} propose defining a natural language template for each type of edge in advance and subsequently filling in the endpoints of each edge into the corresponding template based on its type.~\citett{ref:instructglm} employ natural language to describe the information of $1$-hop and $2$-hop neighboring nodes of the central node.~\citett{ref:graphrag} utilize LLMs to generate report-like summaries for each detected graph community.~\citett{ref:rra,ref:kn} adopt LMs to rewrite the edge table of retrieved subgraphs, generating a natural language description.~\citett{ref:talk} explore different representations of nodes (e.g., Integer encoding, alphabet letters, names, etc.) and edges (e.g., parenthesis, arrows, incident, etc.).~\citett{ref:graphcot,ref:structgpt,ref:kg-agent,ref:knowledgpt,ref:oda} integrate information from different granularities within the graph into prompts through natural language in the form of dialogue.

\paragraph{(3) Code-Like Forms} Considering that natural language descriptions and other $1$-D sequences are inherently inadequate for directly representing the $2$-D structure of graph data, and given the robust code comprehension capabilities of LMs, many researchers~\cite{ref:gpt4graph} explore using code-like formats to represent graph structures. For example, ~\citett{ref:gpt4graph} examine the use of Graph Modeling Language (GML)~\cite{ref:gml} and Graph Markup Language (GraphML)~\cite{ref:graphml} for representing graphs. 
These standardized languages are specifically designed for graph data, providing comprehensive descriptions that encompass nodes, edges, and their interrelationships.

\paragraph{(4) Syntax Tree} Compared to direct flattening of graphs, some research~\cite{ref:graphtext} propose transforming graphs into structures akin to syntax trees. Syntax trees possess a hierarchical structure and, being topological graphs, also maintain a topological order. This method retains more structural information, enhancing the understanding and analysis of the graph's intrinsic properties. Such a transformation not only preserves the relational dynamics between different graph elements but also facilitates more sophisticated algorithms for graph analysis and processing. GRAPHTEXT~\cite{ref:graphtext} proposes transforming the ego network of a central node into a graph-syntax tree format. This format not only encapsulates structural information but also integrates the features of the nodes. By traversing this syntax tree, it is possible to obtain a node sequence that maintains both topological order and hierarchical structure.

\paragraph{(5) Node Sequence} Some studies~\cite{ref:llaga,ref:gnn-rag} propose representing graphs through sequences of nodes, which are often generated using predefined rules. Compared to natural language descriptions, these node sequences are more concise and incorporate prior knowledge, specifically the structural information emphasized by the rules.~\citett{ref:rog,ref:tog} transform the retrieved paths into node sequences and input them into an LLM to enhance the task performance. LLaGA~\cite{ref:llaga} proposes two templates that can transform graphs into node sequences. The first template, known as the Neighborhood Detail Template, offers a detailed examination of the central node along with its immediate surroundings. The second, termed the Hop-Field Overview Template, provides a summarized perspective of a node's neighborhood, which can be expanded to encompass broader areas. GNN-RAG~\cite{ref:gnn-rag} inputs the retrieved reasoning paths into LMs in the form of node sequences as prompts.

\paragraph{Discussion}
Good graph languages should be complete, concise, and comprehensible. Completeness entails capturing all essential information within the graph structure, ensuring no critical details are omitted. Conciseness refers to the necessity of keeping textual descriptions brief to avoid the ``lost in the middle'' phenomenon~\cite{ref:lostin} or exceeding the length limitations of LMs. Lengthy inputs can hinder LMs' processing capabilities, potentially causing loss of context or truncated data interpretation. Comprehensibility ensures that the language used is easily understood by LLMs, facilitating accurate representation of the graph's structure. Due to the characteristics of different graph languages, their choice can significantly impact the performance of downstream tasks~\cite{ref:talk}.

\subsubsection{Graph Embeddings}
The above graph language methods transform graph data into text sequences, which may result in overly lengthy contexts, incurring high computational costs and potentially exceeding the processing limits of LLMs. Additionally, LLMs currently struggle to fully comprehend graph structures even with graph languages~\cite{ref:gpt4graph}. Thus, using GNNs to represent graphs as embeddings presents a promising alternative. The core challenge lies in integrating graph embeddings with textual representations into a unified semantic space. Current research focuses on utilizing prompt tuning methodologies, as discussed earlier. There are also some methods that adopt FiD (Fusion-in-Decoder)~\cite{ref:fid,ref:kgfid}, which first convert the graph data into text, then encode it using an LM-based encoder and input it into the decoders~\cite{ref:reano,ref:decaf,ref:skp}. Notably, feeding graph representations into LMs is feasible primarily with open-source LMs, not closed-source models like GPT-4~\cite{ref:gpt4}. While graph embedding methods avoid handling long text inputs, they face other challenges, such as difficulty in preserving precise information like specific entity names and poor generalization.

\subsection{Generation Enhancement}
In the generation phase, besides converting the retrieved graph data into formats acceptable by the generator and inputting it together with the query to generate the final response, many researchers explore various methods of generation enhancement techniques to improve the quality of output responses. These methods can be classified into three categories based on their application stages: pre-generation enhancement, mid-generation enhancement, and post-generation enhancement.

\subsubsection{Pre-Generation Enhancement}
Pre-generation enhancement techniques focus on improving the quality of input data or representations before feeding them into the generator. In fact, there is no clear boundary between Pre-Generation Enhancement and Retrieval. In this survey, we categorize the retrieval stage as the process of retrieving knowledge from the original graph, and merging and pruning retrieved knowledge. Subsequent operations are considered Pre-Generation Enhancements. 

Commonly used pre-generation enhancement approaches primarily involve semantically enriching the retrieved graph data to achieve tighter integration between the graph data and textual query.~\citett{ref:rra} employ LLMs to rewrite retrieved graph data, enhancing the naturalness and semantic richness of the transformed natural language output. This method not only ensures that graph data is converted into more fluent and natural language but also enriches its semantic content. Conversely, DALK~\cite{ref:dalk} utilizes the retrieved graph data to rewrite the query.~\citett{ref:temple-mqa} first leverage LLMs to generate a reasoning plan and answer queries according to the plan.~\citett{ref:grapeqa,ref:qa-gnn} aim to enhance GNNs by enabling them to learn graph representations relevant to queries. They achieve this by extracting all nouns from the QA pairs (or the QA pairs themselves) and inserting them as nodes into the retrieved subgraph.~\citett{ref:rearev} propose a method where, prior to generation, the representation of the query is decomposed into multiple vectors termed ``instructions'', each representing different features of the query. These instructions are used as conditions during message passing when applying GNNs to learn from retrieved subgraphs. In addition, there are methods that incorporate additional information beyond graph data. For example, PullNet~\cite{ref:pullnet} incorporates documents relevant to entities and MVP-Tuning~\cite{ref:mvp-tuning} retrieves other related questions.

\subsubsection{Mid-Generation Enhancement}
Mid-generation enhancement involves techniques applied during the generation process. These methods typically adjust the generation strategies based on intermediate results or contextual cues. TIARA~\cite{ref:tiara} introduces constrained decoding to control the output space and reduce generation errors. When generating logical forms, if the constrained decoder detects that it is currently generating a pattern item, it restricts the next generated token to options that exist in tries containing KB classes and relations. Compared with the Beam Search, this approach ensures that pattern items generated are guaranteed to exist in the knowledge graph, thereby reducing generation errors. There are other methods adjusting the prompts of LLMs to achieve multi-step reasoning. For example, MindMap~\cite{ref:mindmap} not only produces answers but also generates the reasoning process.

\subsubsection{Post-Generation Enhancement}
Post-generation enhancement occurs after the initial response is generated. Post-generation enhancement methods primarily involve integrating multiple generated responses to obtain the final response. Some methods focus on integrating outputs from the same generator under different conditions or inputs. For example,~\citett{ref:graphrag} generate a summary for each graph community, followed by generating responses to queries based on the summary, and then scoring these responses using an LLM. Ultimately, the responses are sorted in descending order according to their scores and sequentially incorporated into the prompt until the token limit is reached. Subsequently, the LLM generates the final response.~\citett{ref:keqing,ref:kg-gpt} first decompose the query into several sub-questions, then generate answers for each sub-question, and finally merge the answers of all sub-questions to obtain the final answer. Alternatively, other methods combine or select responses generated by different models.~\citett{ref:kagnet,ref:safe} combine the outputs generated by both GNNs and LLMs to reach a synergistic effect. UniOQA~\cite{ref:unioqa} explores two methods for generating answers: one involves generating queries in Cypher Query Language (CQL) to execute and obtain results, while the other method directly generates answers based on retrieved triplets. The final answer is determined through a dynamic selection mechanism. In EmbedKGQA~\cite{ref:embedkgqa}, besides the learned scoring function, researchers additionally design a rule-based score based on the graph structures. These two scores are combined to find the answer entity.~\citett{ref:noname6} combine answers based on retrieved graph data with responses generated according to the LLM's own knowledge. In addition to integrating multiple responses, KALMV~\cite{ref:kalmv} trains a verifier to judge whether the generated answer is correct, and if it is not, to further determine whether the error is due to generation or retrieval.

\section{Training}
\label{sec:training}
In this section, we summarize the individual training of retrievers, generators, and their joint training. We categorize previous works into Training-Free and Training-Based approaches based on whether explicit training is required. Training-Free methods are commonly employed when using closed-source LLMs such as GPT-4~\cite{ref:gpt4} as retrievers or generators. These methods primarily rely on carefully crafted prompts to control the retrieval and generation capabilities of LLMs. Despite LLMs' strong abilities in text comprehension and reasoning, a challenge of Training-Free methods lies in the potential sub-optimality of results due to the lack of specific optimization for downstream tasks. Conversely, Training-Based methods involve training or fine-tuning models using supervised signals. These approaches enhance the model performance by adapting them to specific task objectives, thereby potentially improving the quality and relevance of retrieved or generated content. Joint training of retrievers and generators aims to enhance their synergy, thereby boosting performance on downstream tasks. This collaborative approach leverages the complementary strengths of both components to achieve more robust and effective results in information retrieval and content generation applications.

\subsection{Training Strategies of Retriever}
\subsubsection{Training-Free}
There are two primary types of Training-Free Retrievers currently in use. The first type consists of non-parametric retrievers. These retrievers rely on pre-defined rules or traditional graph search algorithms rather than specific models~\cite{ref:qa-gnn,ref:grapeqa}. The second type utilizes pre-trained LMs as retrievers. Specifically,  one group of works utilizes pre-trained embedding models to encode the queries and perform retrieval directly based on the similarity between the query and graph elements~\cite{ref:noname2}. Another group of works adopts generative language models for training-free retrieval. Candidate graph elements such as entities, triples, paths, or subgraphs are included as part of the prompt input to the LLMs. The LLMs then leverage semantic associations to select appropriate graph elements based on the provided prompt~\cite{ref:graphrag,ref:tog,ref:rok,ref:graphcot,ref:kg-gpt,ref:keqing,ref:gnn-rag}. These methods harness the powerful semantic understanding capabilities of LMs to retrieve relevant graph elements without the need for explicit training. 

\subsubsection{Training-Based}
When the retrieval granularity is nodes or triplets, many methods train retrievers to maximize the similarity between the retrieval ground truth and the query. For instance, MemNNs~\cite{ref:memnns} leverages metric learning to closely align the ground truth with the query in semantic space while differentiating unrelated facts from the query. On the contrary, when the retrieval granularity is paths, training retrievers often adopts an autoregressive approach, where the previous relationship path is concatenated to the end of the query. The model then predicts the next relation based on the concatenated input~\cite{ref:rra,ref:kn}. 

However, the lack of ground truth for retrieval content in the majority of datasets poses a significant challenge. To address this issue, many methods attempt to construct reasoning paths based on distant supervision to guide retriever training. For example,~\citett{ref:sr,ref:ksl,ref:rog} extract all paths (or shortest paths) between entities in the queries and entities in the answers, using them as training data for the retriever. In addition,~\citett{ref:sr} also employ a relationship extraction dataset for distant supervision in unsupervised settings. There is another category of methods that utilize implicit intermediate supervision signals to train Retrievers. For instance, NSM~\cite{ref:nsm} employs a bidirectional search strategy, where two retrievers start searching from the head entity and tail entity, respectively. The supervised objective is to ensure that the paths searched by the two retrievers converge as closely as possible. KnowGPT~\cite{ref:knowgpt} and MINERVA~\cite{ref:minerva} treat the selection of adjacent nodes to build paths or subgraphs as a Markov process. They design the reward function around the inclusion of the answer in the retrieved information and adopt reinforcement learning methods e.g. policy gradient to optimize the retriever.

Some methods argue that distant supervision signals or implicit intermediate supervision signals may contain considerable noise, making it challenging to train effective retrievers. Therefore, they consider employing self-supervised methods for pre-training retrievers. SKP~\cite{ref:skp} pre-trains the DPR (Dense Passage Retrieval) model~\cite{ref:dpr}. Initially, it conducts random sampling on subgraphs and transforms the sampled subgraphs into passages. Subsequently, it randomly masks passages, trains the model using a Masked Language Model (MLM), and employs contrastive learning by treating the masked passages and original passages as positive pairs for comparison.

\subsection{Training of Generator}
\subsubsection{Training-Free}
Training-Free Generators primarily cater to closed-source LLMs or scenarios where avoiding high training costs is essential. In these methods, the retrieved graph data is fed into the LLM alongside the query. The LLMs then generate responses based on the task description provided in the prompt, heavily relying on their inherent ability to understand both the query and the graph data.

\subsubsection{Training-Based}
Training the generator can directly receive supervised signals from downstream tasks. For generative LLMs, fine-tuning can be achieved using supervised fine-tuning (SFT), where task descriptions, queries, and graph data are inputted, and the output is compared against the ground truth for the downstream task~\cite{ref:rog,ref:g-retriever,ref:grag}. On the other hand, for GNNs or discriminative models functioning as generators, specialized loss functions tailored to the downstream tasks are employed to train the models effectively~\cite{ref:grapeqa,ref:qa-gnn,ref:noname2,ref:safe,ref:greaselm}.

\subsection{Joint Training}
Jointly training retrievers and generators simultaneously enhances performance on downstream tasks by leveraging their complementary strengths. Some approaches unify retrievers and generators into a single model, typically LLMs, and train them with both retrieval and generation objectives simultaneously~\cite{ref:rog}. This method capitalizes on the cohesive capabilities of a unified architecture, enabling the model to seamlessly retrieve relevant information and generate coherent responses within a single framework. 

Other methodologies involve initially training retrievers and generators separately, followed by joint training techniques to fine-tune both components. For instance, Subgraph Retriever~\cite{ref:sr} adopts an alternating training paradigm, where the retriever's parameters are fixed to use the graph data for training the generator. Subsequently, the generator's parameters are fixed, and feedback from the generator is used to guide the retriever's training. This iterative process helps both components refine their performance in a coordinated manner.

\section{Applications and Evaluation}
\label{sec:app}
In this section, we will summarize the downstream tasks, application domains, benchmarks and metrics, and industrial applications related to GraphRAG. Table~\ref{tab:app} collects existing GraphRAG techniques, categorizing them by downstream tasks, benchmarks, methods, and evaluation metrics. This table serves as a comprehensive overview, highlighting the various aspects and applications of GraphRAG technologies across different domains.

\begin{table}[htbp]
  \centering
  \caption{The tasks, benchmarks, methods, and metrics of GraphRAG.}
    \resizebox{\linewidth}{!} {
    \begin{tabular}{ccclc}
    \toprule
    \multicolumn{2}{c}{Tasks} & \multicolumn{1}{c}{Benchmarks} & \multicolumn{1}{c}{Methods} & \multicolumn{1}{c}{Metrics} \\
    \midrule
    \multirow{20}[5]{*}{QA} & \multirow{13}[2]{*}{KBQA} & WebQSP~\cite{ref:webqsp} & \multicolumn{1}{p{30.59em}}{\cite{ref:rog}, \cite{ref:tog}, \cite{ref:tog2}, \cite{ref:sr}, \cite{ref:rra}, \cite{ref:kn}, \cite{ref:kd-cot}, \cite{ref:structgpt}, \cite{ref:kg-agent}, \cite{ref:chatkbqa}, \cite{ref:keqing}, \cite{ref:difar}, \cite{ref:tiara}, \cite{ref:etd}, \cite{ref:gnn-rag}, \cite{ref:pullnet}, \cite{ref:decaf},  \cite{ref:skp}, \cite{ref:gge}, \cite{ref:rasr}, \cite{ref:kaping}, \cite{ref:embedkgqa}, \cite{ref:nsm}, \cite{ref:unikgqa}, \cite{ref:nutrea}, \cite{ref:graft}, \cite{ref:unikqa}, \cite{ref:kalmv}, \cite{ref:ewek-qa}} & \multicolumn{1}{l}{\multirow{21}[5]{*}{\makecell{Accuracy, \\EM, \\Recall, \\F1, \\BERTScore, \\GPT-4 Average Ranking}}} \\
          &       & WebQ~\cite{ref:webq}  &\cite{ref:rok},~\cite{ref:tog},~\cite{ref:oreolm},~\cite{ref:rra},~\cite{ref:rearev},~\cite{ref:unikqa},~\cite{ref:memnns},~\cite{ref:ewek-qa} &  \\
          &       & CWQ~\cite{ref:cwq}   & \multicolumn{1}{p{30.59em}}{\cite{ref:rog}, \cite{ref:tog}, \cite{ref:oreolm}, \cite{ref:sr}, \cite{ref:kd-cot}, \cite{ref:kg-agent}, \cite{ref:chatkbqa}, \cite{ref:rearev}, \cite{ref:noname8}, \cite{ref:etd}, \cite{ref:gnn-rag}, \cite{ref:pullnet}, \cite{ref:decaf}, 
 \cite{ref:gge}, \cite{ref:rasr}, \cite{ref:unikgqa}, \cite{ref:noname6}, \cite{ref:nsm}, \cite{ref:nutrea}, \cite{ref:ewek-qa}} &  \\
          &       & GrailQA~\cite{ref:grailqa} &\cite{ref:tog},~\cite{ref:kg-agent},~\cite{ref:tiara},~\cite{ref:ewek-qa} &  \\
          &       & QALD10-en~\cite{ref:qald} &\cite{ref:tog},~\cite{ref:tog2},~\cite{ref:noname6},~\cite{ref:oda} &  \\
          &       & SimpleQuestions~\cite{ref:simple} &\cite{ref:tog},~\cite{ref:difar},~\cite{ref:memnns},~\cite{ref:ewek-qa} &  \\
          &       & CMCQA\footnote{. https://github.com/WENGSYX/CMCQA} &\cite{ref:rok}   &  \\
          &       & MetaQA~\cite{ref:metaqa} & \multicolumn{1}{p{30.59em}}{\cite{ref:rog},~\cite{ref:rra},~\cite{ref:kn},~\cite{ref:kg-gpt},~\cite{ref:keqing},~\cite{ref:rearev},~\cite{ref:pullnet},~\cite{ref:embedkgqa},~\cite{ref:structgpt},~\cite{ref:etd},~\cite{ref:nsm},~\cite{ref:unikgqa},~\cite{ref:nutrea},~\cite{ref:kelp}} &  \\
          &       & Natural Question~\cite{ref:nq} &\cite{ref:oreolm},~\cite{ref:kalmv},~\cite{ref:ewek-qa} &  \\
          &       & TriviaQA~\cite{ref:triviaqa} &\cite{ref:oreolm},~\cite{ref:unikgqa} &  \\
          &       & HotpotQA~\cite{ref:hotpotqa} &\cite{ref:oreolm},~\cite{ref:hipporag},~\cite{ref:tog2},~\cite{ref:kalmv},~\cite{ref:ewek-qa} &  \\
          &       & Mintaka~\cite{ref:mintaka} &\cite{ref:difar},~\cite{ref:noname6},~\cite{ref:kaping},~\cite{ref:kalmv} &  \\
          &       & FreebaseQA~\cite{ref:freebaseqa} &\cite{ref:decaf},~\cite{ref:rasr} &  \\
\cmidrule{2-4}          & \multirow{6}[2]{*}{CSQA} & CSQA~\cite{ref:commonsenseqa}  &\cite{ref:grapeqa},~\cite{ref:qa-gnn},~\cite{ref:mvp-tuning},~\cite{ref:noname2},~\cite{ref:kagnet},~\cite{ref:ksl},~\cite{ref:hamqa} &  \\
          &       & OBQA~\cite{ref:openbookqa}  &\cite{ref:grapeqa},~\cite{ref:qa-gnn},~\cite{ref:mvp-tuning},~\cite{ref:noname2},~\cite{ref:ksl},~\cite{ref:opencsr},~\cite{ref:hamqa} &  \\
          &       & MedQA~\cite{ref:medqa} & \cite{ref:grapeqa},~\cite{ref:ksl},~\cite{ref:dalk} &  \\
          &       & SocialIQA~\cite{ref:socialIQA} &\cite{ref:mvp-tuning} &  \\
          &       & PIQA~\cite{ref:piqa}  &\cite{ref:mvp-tuning} &  \\
          &       & RiddleSenseQA~\cite{ref:riddle} &\cite{ref:mvp-tuning} &  \\

    \midrule
    \multirow{4}[1]{*}{IE} & \multirow{2}[0]{*}{Entity Linking} & ZESHEL~\cite{ref:zeshel} &  \cite{ref:ger}   & \multirow{2}[0]{*}{Recall@$K$} \\
          &       & CoNLL~\cite{ref:conll} & \cite{ref:ger}   &  \\
          & \multirow{2}[1]{*}{Relation Extraction} & T-Rex~\cite{ref:t-rex} & \cite{ref:oda},~\cite{ref:tog} & \multirow{2}[1]{*}{Hits@1} \\
          &       & ZsRE~\cite{ref:zsre}  & \cite{ref:noname6},~\cite{ref:oda},~\cite{ref:tog},~\cite{ref:tog2} &  \\
    \midrule
    \multirow{8}[2]{*}{Others} & \multirow{2}[0]{*}{Fact Verification} & Creak~\cite{ref:creak} & \cite{ref:noname6},~\cite{ref:oda},~\cite{ref:tog},~\cite{ref:tog2} & \multirow{2}[0]{*}{Accuracy, F1} \\
    &     & FACTKG~\cite{ref:factkg} &\cite{ref:kg-gpt},~\cite{ref:noname8},~\cite{ref:kelp} &  \\
          & \multirow{4}[0]{*}{Link Prediction} & FB15K-237~\cite{ref:fb15k-237} & \cite{ref:lark},~\cite{ref:kg-r3} & \multirow{4}[0]{*}{MRR, Hits@$K$} \\
          &       & FB15k~\cite{ref:fb15k} & \cite{ref:lark} &  \\
          &       & WN18RR~\cite{ref:wn} & \cite{ref:kg-r3} &  \\
          &       & NELL995~\cite{ref:nell} & \cite{ref:lark}  &  \\
          & Dialogue Systems & OpenDialKG~\cite{ref:opendialkg} & \cite{ref:difar} & MRR, Hits@$K$ \\
          & Recommendation & Yelp\footnote{https://www.yelp.com/dataset/} & \cite{ref:rete} & NDCG@$K$, Recall@$K$ \\
    \bottomrule
    \end{tabular}%
    }
  \label{tab:app}%
\end{table}%

\subsection{Downstream Tasks}
\label{sec:task}
GraphRAG is applied in various downstream tasks (especially NLP tasks), including Question Answering, Information Extraction, and others.

\subsubsection{Question Answering}
The QA tasks specifically include Knowledge Base Question Answering (KBQA) and CommonSense Question Answering (CSQA).

\paragraph{(1) KBQA}
KBQA serves as a cornerstone downstream task for GraphRAG. In KBQA, questions typically pertain to specific knowledge graphs, and answers often involve entities, relationships, or operations between sets of entities within the knowledge graph. The task tests the systems' ability to retrieve and reason over structured knowledge bases, which is crucial in facilitating complex query responses. 

\paragraph{(2) CSQA}
Distinguished from KBQA, CSQA primarily takes the form of multiple-choice questions. Commonsense reasoning typically presents a commonsense question along with several answer options, each potentially representing either the name of an entity or a statement. The objective is for machines to utilize external commonsense knowledge graphs, such as ConceptNet, to find relevant knowledge pertaining to the question and options, and to engage in appropriate reasoning and derive the correct answer. 

\subsubsection{Information Retrieval}

Information Retrieval tasks consist of two categories: Entity Linking (EL) and Relation Extraction (RE).

\paragraph{(1) Entity Linking} Entity Linking (EL) is a critical task in the field of natural language processing that involves identifying entities mentioned in text segments and linking them to their corresponding entities in a knowledge graph. By leveraging a system such as Graph RAG, it is possible to retrieve relevant information from the knowledge graph, which facilitates the accurate inference of the specific entities that match the mentions in the text~\cite{ref:ger}.

\paragraph{(2) Relation Extraction} Relation Extraction (RE) aims at identifying and classifying semantic relationships between entities within a text. GraphRAG can significantly enhance this task by using graph-based structures to encode and exploit the interdependencies among entities, thus facilitating more accurate and contextually nuanced extraction of relational data from diverse text sources~\cite{ref:tog,ref:oda,ref:noname6}.

\subsubsection{Others}
In addition to the aforementioned downstream tasks, GraphRAG can be applied to various other tasks in the realm of natural language processing such as fact verification, link prediction, dialogue systems, and recommendation.

\paragraph{(1) Fact Verification}
The fact verification task typically involves assessing the truthfulness of a factual statement using knowledge graphs. Models are tasked with determining the validity of a given factual assertion by leveraging structured knowledge repositories. GraphRAG techniques can be utilized to extract evidential connections between entities to enhance the system's efficiency and accuracy~\cite{ref:noname6,ref:tog,ref:oda,ref:foodgpt}.

\paragraph{(2) Link Prediction} Link prediction involves predicting missing relationships or potential connections between entities in a graph. GraphRAG is applied to this task~\cite{ref:lark,ref:kg-r3} by leveraging its ability to retrieve and analyze structured information from graphs, enhancing prediction accuracy by uncovering latent relationships and patterns within the graph data.

\paragraph{(3) Dialogue Systems} Dialogue Systems is designed to converse with humans using natural language, handling various tasks such as answering questions, providing information, or facilitating user interactions. By structuring conversation histories and contextual relationships in a graph-based framework, GraphRAG systems~\cite{ref:difar} can improve the model's ability to generate coherent and contextually relevant responses.

\paragraph{(4) Recommendation} 
In the context of E-commerce platforms, the purchase relationships between users and products naturally form a network graph. The primary objective of recommendation within these platforms is to predict the future purchasing intentions of users, effectively forecasting the potential connections within this graph~\cite{ref:rete}.

\subsection{Application Domains}
GraphRAG is widely applied in E-commerce and biomedical, academic, literature, legal, and other application scenarios for its outstanding ability to integrate structured knowledge graphs with natural language processing, which will be introduced below.

\subsubsection{E-Commerce}
The primary goal in the E-commerce area involves improving customer shopping experiences and increasing sales through personalized recommendations and intelligent customer services. In this area, historical interactions between users and products can naturally form a graph, which implicitly encapsulates users' behavioral patterns and preference information. However, due to the increasing number of E-commerce platforms and the growing volume of user interaction data, using GraphRAG technology to extract key subgraphs is crucial.~\citett{ref:rete} ensemble multiple retrievers under different types or with different parameters to extract relevant subgraphs, which are then encoded for temporal user action prediction. To improve the model performance of customer service question answering systems,~\citett{ref:noname1} construct a past-issue graph with intra-issue and inter-issue relations. For each given query, subgraphs of similar past issues are retrieved to enhance the system’s response quality.

\subsubsection{Biomedical}
Recently, GraphRAG techniques are increasingly applied in biomedical question answering systems, achieving advanced medical decision-making performance. In this area, each disease is associated with specific symptoms, and every medication contains certain active ingredients that target and treat particular diseases. Some researchers~\cite{ref:dalk,ref:noname5,ref:medgraphrag} construct KGs for specific task scenarios, while others~\cite{ref:hykge,ref:kg-rank,ref:mindmap} utilize open-source knowledge graphs such as CMeKG and CPubMed-KG as retrieval sources. Existing methods generally begin with non-parametric retrievers for initial search, followed by designing methods to filter retrieved content through re-ranking~\cite{ref:hykge,ref:kg-rank,ref:mindmap,ref:dalk,ref:noname5}. Additionally, some approaches propose rewriting model inputs using retrieved information to enhance generation effectiveness~\cite{ref:dalk}.

\subsubsection{Academic}
In the academic research domain, each paper is authored by one or more researchers and is associated with a field of study. Authors are affiliated with institutions, and there exist relationships among authors, such as collaboration or shared institutional affiliations. These elements can be structured into a graph format. Utilizing GraphRAG on this graph can facilitate academic exploration, including predicting potential collaborators for an author, identifying trends within a specific field, etc.

\subsubsection{Literature}
Similar to academic research, a knowledge graph can be constructed in the realm of literature, with nodes representing books, authors, publishers, and series, and edges labeled ``written-by'', ``published-in'', and ``book-series''. GraphRAG can be utilized to enhance realistic applications like smart libraries.

\subsubsection{Legal}
In legal contexts, extensive citation connections exist between cases and judicial opinions, as judges frequently reference previous opinions when making new decisions. This naturally creates a structured graph where nodes represent opinions, opinion clusters, dockets, and courts, and edges encompass relationships such as ``opinion-citation'', ``opinion-cluster'', ``cluster-docket'', and ``docket-court''. The application of GraphRAG in legal scenarios could aid lawyers and legal researchers in various tasks such as case analysis and legal consultation.

\subsubsection{Others}
In addition to the above applications, GraphRAG is also applied to other real-world scenarios such as intelligence report generation~\cite{ref:fabula}, patent phrase similarity detection~\cite{ref:ra-sim} and software understanding~\cite{ref:depsrag}.~\citett{ref:fabula} first construct an Event Plot Graph (EPG) and retrieve the critical aspects of the events to aid the generation of intelligence reports.~\citett{ref:ra-sim} create a patent-phrase graph and retrieve the ego network of the given patent phrase to assist the judgment of phrase similarity. ~\citett{ref:depsrag} propose a Chatbot to understand properties about dependencies in a given software package, which first automatically constructs the dependency graph and then the user can ask questions about the dependencies in the dependency graph.

\subsection{Benchmarks and Metrics}

\subsubsection{Benchmarks}
The benchmarks used to evaluate the performance of the GraphRAG system can be divided into two categories. The first category is the corresponding datasets of downstream tasks. We summarize the benchmarks and papers tested with them according to the classification in Section~\ref{sec:task}, details of which are shown in Table~\ref{tab:app}. The second category consists of benchmarks specifically designed for the GraphRAG systems. These benchmarks usually cover multiple task domains to provide a comprehensive test result. For example, STARK~\cite{ref:stark} benchmarks LLM Retrieval on semi-structured knowledge bases covering three domains, including product search, academic paper search, and queries in precision medicine to access the capacity of current GraphRAG systems.~\citett{ref:g-retriever} propose a flexible question-answering benchmark targeting real-world textual graphs, named GraphQA, which is applicable to multiple applications including scene graph understanding, commonsense reasoning, and knowledge graph reasoning. Graph Reasoning Benchmark (GRBENCH)~\cite{ref:graphcot} is constructed to facilitate the research of augmenting LLMs with graphs, which contains 1,740 questions that can be answered with the knowledge from 10 domain graphs. CRAG~\cite{ref:crag} provides a structured query dataset, with additional mock APIs to
access information from underlying mock KGs to achieve fair comparison.

\subsubsection{Metrics}
The evaluation metrics for GraphRAG can be broadly categorized into two main types: downstream task evaluation (generation quality) and retrieval quality.

\paragraph{(1) Downstream Task Evaluation (Generation Quality)}
In the majority of research studies, downstream task evaluation metrics serve as the primary method for assessing GraphRAG's performance. For example, in KBQA, Exact Match (EM) and F1 score are commonly used to measure the accuracy of answering entities. In addition, many researchers utilize BERT4Score and GPT4Score to mitigate instances where LLMs generate entities that are synonymous with the ground truth but not exact matches. In CSQA, Accuracy is the most commonly used evaluation metric. For generative tasks such as QA systems, metrics like BLEU, ROUGE-L, METEOR, and others are commonly employed to assess the quality of the text generated by the model.

\paragraph{(2) Retrieval Quality Evaluation}
While evaluating GraphRAG based on downstream task performance is feasible, directly measuring the accuracy of retrieved content poses challenges. Therefore, many studies employ specific metrics to gauge the precision of retrieved content. For instance, when ground truth entities are available, retrieval systems face a balance between the quantity of retrieved information and the coverage of answers. Hence, some studies utilize the ratio between answer coverage and the size of the retrieval subgraph to evaluate the performance of the retrieval system. In addition, several studies have explored metrics such as query relevance, diversity, and faithfulness score to respectively assess the similarity between retrieved content and queries, the diversity of retrieved content, and the faithfulness of the information retrieved.

\subsection{GraphRAG in Industry}
In this section, we mainly focus on industrial GraphRAG systems. These systems are characterized by their reliance on industrial graph database systems or their focus on large-scale graph data, details of which are as follows.

$\bullet$ GraphRAG (by Microsoft)\footnote{https://github.com/microsoft/graphrag}: The system uses LLMs to construct entity-based knowledge graphs and pre-generate community summaries of related entity groups, which enables the capture of both local and global relationships within a document collection, thereby enhancing Query-Focused Summarization (QFS) task~\cite{ref:graphrag}. The project can also utilize open-source RAG toolkits for rapid implementation, such as LlamaIndex\footnote{https://docs.llamaindex.ai/en/stable/ examples/index structs/knowledge graph/KnowledgeGraphDemo.html}, LangChain\footnote{https://python.langchain.com/docs/use\_cases/graph}, etc.

$\bullet$ GraphRAG (by NebulaGraph)\footnote{https://www.nebula-graph.io/posts/graph-RAG}: The project is the first industrial GraphRAG system, which is developed by NebulaGraph Corporation. The project integrates LLMs into the NebulaGraph database, which aims to deliver more intelligent and precise search results.

$\bullet$ GraphRAG (by Antgroup)\footnote{https://github.com/eosphoros-ai/DB-GPT}: The framework is developed on the foundation of several AI engineering frameworks such as DB-GPT, knowledge graph engine OpenSPG, and graph database TuGraph. Specifically, the system begins by extracting triples from documents using LLMs, which are then stored in the graph database. During the retrieval phase, it identifies keywords from the query, locates corresponding nodes in the graph database, and traverses the subgraph using BFS or DFS. In the generation phase, the retrieved subgraph data is formatted into text and submitted along with the context and query for processing by LLMs.

$\bullet$ NallM (by Neo4j)\footnote{https://github.com/neo4j/NaLLM}: The NaLLM (Neo4j and Large Language Models) framework integrates Neo4j graph database technology with LLMs. It aims to explore and demonstrate the synergy between Neo4j and LLMs, focusing on three primary use cases: Natural Language Interface to a Knowledge Graph, Creating a Knowledge Graph from Unstructured Data, and Generate Reports Using Both Static Data and LLM Data.

$\bullet$ LLM Graph Builder (by Neo4j)\footnote{https://github.com/neo4j-labs/llm-graph-builder}: It is a project developed by Neo4j for automatically constructing knowledge graphs, suitable for the GraphRAG's Graph Database Construction and Indexing phase. The project primarily utilizes LLMs to extract nodes, relationships, and their properties from unstructured data, and utilizes the LangChain framework to create structured knowledge graphs.

\section{Future Prospects}
\label{sec:future}
While GraphRAG technology has made substantial strides, it continues to face enduring challenges that demand comprehensive exploration. This section will delve into the prevalent obstacles and outline prospective avenues for future research in the field of GraphRAG.

\subsection{Dynamic and Adaptive Graphs}
Most GraphRAG methods~\cite{ref:graphrag,ref:kbqasurvey1,ref:kbqasurvey2,ref:kbqasurvey3,ref:kbqasurvey4,ref:chatkbqa} are built upon static databases; however, as time progresses, new entities and relationships inevitably emerge~\cite{ref:rag4dyg,ref:temple-mqa,ref:gentkgqa}. Rapidly updating these changes is both promising and challenging. Incorporating updated information is crucial for achieving better results and addressing emerging trends that require current data. Developing efficient methods for dynamic updates and real-time integration of new data will significantly enhance the effectiveness and relevance of GraphRAG systems.

\subsection{Multi-Modality Information Integration}
Most knowledge graphs primarily encompass textual information, thereby lacking the inclusion of other modalities such as images, audio, and videos, which hold the potential to significantly enhance the overall quality and richness of the database~\cite{ref:mmgcn}. The incorporation of these diverse modalities could provide a more comprehensive and nuanced understanding of the stored knowledge. However, the integration of such multi-modal data presents considerable challenges. As the volume of information increases, the graph's complexity and size grow exponentially, rendering it increasingly difficult to manage and maintain. This escalation in scale necessitates the development of advanced methodologies and sophisticated tools to efficiently handle and seamlessly integrate the diverse data types into the existing graph structure, ensuring both the accuracy and accessibility of the enriched knowledge graph.

\subsection{Scalable and Efficient Retrieval Mechanisms}
Knowledge graphs in the industrial setting may encompass millions or even billions of entities, representing a vast and intricate scale. However, most contemporary methods are tailored for small-scale knowledge graphs~\cite{ref:graphrag}, which may only comprise thousands of entities. Efficiently and effectively retrieving pertinent entities within large-scale knowledge graphs remains a practical and significant challenge. Developing advanced retrieval algorithms and scalable infrastructure is essential to address this issue, ensuring that the system can manage the extensive data volume while maintaining high performance and accuracy in entity retrieval.

\subsection{Combination with Graph Foundation Model}
Recently, graph foundation models~\cite{ref:ultra,ref:gfm}, which can effectively address a wide range of graph tasks, have achieved significant success. Deploying these models to enhance the current GraphRAG pipeline is an essential problem. The input data for graph foundation models is inherently graph-structured, enabling them to handle such data more efficiently than LLM models. Integrating these advanced models into the GraphRAG framework could greatly improve the system's ability to process and utilize graph-structured information, thereby enhancing overall performance and capability.

\subsection{Lossless Compression of Retrieved Context}
In GraphRAG, the retrieved information is organized into a graph structure containing entities and their interrelations. This information is then transformed into a sequence that can be understood by LLMs, resulting in a very long context. There are two issues with inputting such long contexts: LLMs cannot handle very long sequences, and extensive computation during inference can be a hindrance for individuals. To address these problems, lossless compression of long contexts is crucial. This approach removes redundant information and compresses lengthy sentences into shorter, yet meaningful ones. It helps LLMs capture the essential parts of the context and accelerates inference. However, designing a lossless compression technique is challenging. Current works~\cite{ref:kbqasurvey1,ref:kbqasurvey2} make a trade-off between compression and preserving information. Developing an effective lossless compression technique is crucial but challenging for GraphRAG.

\subsection{Standard Benchmarks}
GraphRAG is a relatively new field that lacks unified and standard benchmarks for evaluating different methods. Establishing a standard benchmark is crucial for this area as it can provide a consistent framework for comparison, facilitate objective assessments of various approaches, and drive progress by identifying strengths and weaknesses. This benchmark should encompass diverse and representative datasets, well-defined evaluation metrics, and comprehensive test scenarios to ensure robust and meaningful evaluations of GraphRAG methods.

\subsection{Broader Applications}
Current GraphRAG applications primarily focus on common tasks such as customer service systems~\cite{ref:noname1}, recommendation systems~\cite{ref:rrs}, and KBQA~\cite{ref:kbqasurvey1}. Extending GraphRAG to broader applications such as healthcare~\cite{ref:kashyap2024knowledge}, financial services~\cite{ref:financial}, legal and compliance~\cite{ref:legal}, smart cities and IoT~\cite{ref:iot}, and more, involves incorporating more complex techniques.
For instance, in healthcare, GraphRAG can support medical diagnosis, patient record analysis, and personalized treatment plans by integrating medical literature, patient histories, and real-time health data. In financial services, GraphRAG can be utilized for fraud detection, risk assessment, and personalized financial advice by analyzing transactional data, market trends, and customer profiles. Legal and compliance applications can benefit from GraphRAG by enabling comprehensive legal research, contract analysis, and regulatory compliance monitoring through the integration of legal documents, case law, and regulatory updates.
Expanding GraphRAG to these diverse and complex domains will enhance its utility and impact, providing more sophisticated and targeted solutions across various fields.

\section{Conclusion}
\label{sec:conclusion}
In summary, this survey offers a comprehensive retrospective of GraphRAG technology, systematically categorizing and organizing its fundamental techniques, training methodologies, and application scenarios. GraphRAG significantly enhances the relevance, accuracy, and comprehensiveness of information retrieval by leveraging pivotal relational knowledge derived from graph datasets, thereby addressing critical limitations associated with traditional Retrieval-Augmented Generation approaches. Furthermore, as GraphRAG represents a relatively nascent field of study, we delineate the benchmarks, analyze prevailing challenges, and illuminate prospective future research directions within this domain.

\begin{acks}
This work is supported by Ant Group through Ant Research Intern Program.
\end{acks}

\bibliographystyle{ACM-Reference-Format}
\bibliography{sample-base}


\begin{thebibliography}{204}


\ifx \showCODEN    \undefined \def \showCODEN     #1{\unskip}     \fi
\ifx \showDOI      \undefined \def \showDOI       #1{#1}\fi
\ifx \showISBNx    \undefined \def \showISBNx     #1{\unskip}     \fi
\ifx \showISBNxiii \undefined \def \showISBNxiii  #1{\unskip}     \fi
\ifx \showISSN     \undefined \def \showISSN      #1{\unskip}     \fi
\ifx \showLCCN     \undefined \def \showLCCN      #1{\unskip}     \fi
\ifx \shownote     \undefined \def \shownote      #1{#1}          \fi
\ifx \showarticletitle \undefined \def \showarticletitle #1{#1}   \fi
\ifx \showURL      \undefined \def \showURL       {\relax}        \fi
\providecommand\bibfield[2]{#2}
\providecommand\bibinfo[2]{#2}
\providecommand\natexlab[1]{#1}
\providecommand\showeprint[2][]{arXiv:#2}

\bibitem[Alhanahnah et~al\mbox{.}(2024)]%
        {ref:depsrag}
\bibfield{author}{\bibinfo{person}{Mohannad Alhanahnah}, \bibinfo{person}{Yazan Boshmaf}, {and} \bibinfo{person}{Benoit Baudry}.} \bibinfo{year}{2024}\natexlab{}.
\newblock \bibinfo{title}{DepsRAG: Towards Managing Software Dependencies using Large Language Models}.
\newblock
\newblock
\showeprint[arxiv]{2405.20455}~[cs.SE]
\urldef\tempurl%
\url{https://arxiv.org/abs/2405.20455}
\showURL{%
\tempurl}


\bibitem[An et~al\mbox{.}(2024)]%
        {ref:golden}
\bibfield{author}{\bibinfo{person}{Zhiyu An}, \bibinfo{person}{Xianzhong Ding}, \bibinfo{person}{Yen-Chun Fu}, \bibinfo{person}{Cheng-Chung Chu}, \bibinfo{person}{Yan Li}, {and} \bibinfo{person}{Wan Du}.} \bibinfo{year}{2024}\natexlab{}.
\newblock \bibinfo{title}{Golden-Retriever: High-Fidelity Agentic Retrieval Augmented Generation for Industrial Knowledge Base}.
\newblock
\newblock
\showeprint[arxiv]{2408.00798}~[cs.IR]
\urldef\tempurl%
\url{https://arxiv.org/abs/2408.00798}
\showURL{%
\tempurl}


\bibitem[Arslan and Cruz(2024)]%
        {ref:financial}
\bibfield{author}{\bibinfo{person}{Muhammad Arslan} {and} \bibinfo{person}{Christophe Cruz}.} \bibinfo{year}{2024}\natexlab{}.
\newblock \showarticletitle{Business-RAG: Information Extraction for Business Insights}.
\newblock \bibinfo{journal}{\emph{ICSBT 2024}} (\bibinfo{year}{2024}), \bibinfo{pages}{88}.
\newblock


\bibitem[Auer et~al\mbox{.}(2007)]%
        {ref:dbpedia}
\bibfield{author}{\bibinfo{person}{S{\"{o}}ren Auer}, \bibinfo{person}{Christian Bizer}, \bibinfo{person}{Georgi Kobilarov}, \bibinfo{person}{Jens Lehmann}, \bibinfo{person}{Richard Cyganiak}, {and} \bibinfo{person}{Zachary~G. Ives}.} \bibinfo{year}{2007}\natexlab{}.
\newblock \showarticletitle{DBpedia: {A} Nucleus for a Web of Open Data}. In \bibinfo{booktitle}{\emph{The Semantic Web, 6th International Semantic Web Conference, 2nd Asian Semantic Web Conference, {ISWC} 2007 + {ASWC} 2007, Busan, Korea, November 11-15, 2007}} \emph{(\bibinfo{series}{Lecture Notes in Computer Science}, Vol.~\bibinfo{volume}{4825})}. \bibinfo{pages}{722--735}.
\newblock


\bibitem[Baek et~al\mbox{.}(2023b)]%
        {ref:difar}
\bibfield{author}{\bibinfo{person}{Jinheon Baek}, \bibinfo{person}{Alham~Fikri Aji}, \bibinfo{person}{Jens Lehmann}, {and} \bibinfo{person}{Sung~Ju Hwang}.} \bibinfo{year}{2023}\natexlab{b}.
\newblock \showarticletitle{Direct Fact Retrieval from Knowledge Graphs without Entity Linking}. In \bibinfo{booktitle}{\emph{Proceedings of the 61st Annual Meeting of the Association for Computational Linguistics (Volume 1: Long Papers), {ACL} 2023, Toronto, Canada, July 9-14, 2023}}. \bibinfo{pages}{10038--10055}.
\newblock


\bibitem[Baek et~al\mbox{.}(2023a)]%
        {ref:kaping}
\bibfield{author}{\bibinfo{person}{Jinheon Baek}, \bibinfo{person}{Alham~Fikri Aji}, {and} \bibinfo{person}{Amir Saffari}.} \bibinfo{year}{2023}\natexlab{a}.
\newblock \bibinfo{title}{Knowledge-Augmented Language Model Prompting for Zero-Shot Knowledge Graph Question Answering}.
\newblock
\newblock
\showeprint[arxiv]{2306.04136}~[cs.CL]
\urldef\tempurl%
\url{https://arxiv.org/abs/2306.04136}
\showURL{%
\tempurl}


\bibitem[Baek et~al\mbox{.}(2023c)]%
        {ref:kalmv}
\bibfield{author}{\bibinfo{person}{Jinheon Baek}, \bibinfo{person}{Soyeong Jeong}, \bibinfo{person}{Minki Kang}, \bibinfo{person}{Jong~C. Park}, {and} \bibinfo{person}{Sung~Ju Hwang}.} \bibinfo{year}{2023}\natexlab{c}.
\newblock \showarticletitle{Knowledge-Augmented Language Model Verification}. In \bibinfo{booktitle}{\emph{Proceedings of the 2023 Conference on Empirical Methods in Natural Language Processing, {EMNLP} 2023, Singapore, December 6-10, 2023}}. \bibinfo{pages}{1720--1736}.
\newblock


\bibitem[Berant et~al\mbox{.}(2013)]%
        {ref:webq}
\bibfield{author}{\bibinfo{person}{Jonathan Berant}, \bibinfo{person}{Andrew Chou}, \bibinfo{person}{Roy Frostig}, {and} \bibinfo{person}{Percy Liang}.} \bibinfo{year}{2013}\natexlab{}.
\newblock \showarticletitle{Semantic Parsing on Freebase from Question-Answer Pairs}. In \bibinfo{booktitle}{\emph{Proceedings of the 2013 Conference on Empirical Methods in Natural Language Processing, {EMNLP} 2013, 18-21 October 2013, Grand Hyatt Seattle, Seattle, Washington, USA, {A} meeting of SIGDAT, a Special Interest Group of the {ACL}}}. \bibinfo{pages}{1533--1544}.
\newblock


\bibitem[Bisk et~al\mbox{.}(2020)]%
        {ref:piqa}
\bibfield{author}{\bibinfo{person}{Yonatan Bisk}, \bibinfo{person}{Rowan Zellers}, \bibinfo{person}{Ronan~Le Bras}, \bibinfo{person}{Jianfeng Gao}, {and} \bibinfo{person}{Yejin Choi}.} \bibinfo{year}{2020}\natexlab{}.
\newblock \showarticletitle{{PIQA:} Reasoning about Physical Commonsense in Natural Language}. In \bibinfo{booktitle}{\emph{The Thirty-Fourth {AAAI} Conference on Artificial Intelligence, {AAAI} 2020, The Thirty-Second Innovative Applications of Artificial Intelligence Conference, {IAAI} 2020, The Tenth {AAAI} Symposium on Educational Advances in Artificial Intelligence, {EAAI} 2020, New York, NY, USA, February 7-12, 2020}}. \bibinfo{pages}{7432--7439}.
\newblock


\bibitem[Bollacker et~al\mbox{.}(2008a)]%
        {ref:freebase}
\bibfield{author}{\bibinfo{person}{Kurt Bollacker}, \bibinfo{person}{Colin Evans}, \bibinfo{person}{Praveen Paritosh}, \bibinfo{person}{Tim Sturge}, {and} \bibinfo{person}{Jamie Taylor}.} \bibinfo{year}{2008}\natexlab{a}.
\newblock \showarticletitle{Freebase: a collaboratively created graph database for structuring human knowledge}. In \bibinfo{booktitle}{\emph{Proceedings of the 2008 ACM SIGMOD international conference on Management of data}}. \bibinfo{pages}{1247--1250}.
\newblock


\bibitem[Bollacker et~al\mbox{.}(2008b)]%
        {ref:fb15k}
\bibfield{author}{\bibinfo{person}{Kurt~D. Bollacker}, \bibinfo{person}{Colin Evans}, \bibinfo{person}{Praveen~K. Paritosh}, \bibinfo{person}{Tim Sturge}, {and} \bibinfo{person}{Jamie Taylor}.} \bibinfo{year}{2008}\natexlab{b}.
\newblock \showarticletitle{Freebase: a collaboratively created graph database for structuring human knowledge}. In \bibinfo{booktitle}{\emph{Proceedings of the {ACM} {SIGMOD} International Conference on Management of Data, {SIGMOD} 2008, Vancouver, BC, Canada, June 10-12, 2008}}. \bibinfo{pages}{1247--1250}.
\newblock


\bibitem[Bordes et~al\mbox{.}(2015a)]%
        {ref:memnns}
\bibfield{author}{\bibinfo{person}{Antoine Bordes}, \bibinfo{person}{Nicolas Usunier}, \bibinfo{person}{Sumit Chopra}, {and} \bibinfo{person}{Jason Weston}.} \bibinfo{year}{2015}\natexlab{a}.
\newblock \bibinfo{title}{Large-scale Simple Question Answering with Memory Networks}.
\newblock
\newblock
\showeprint[arxiv]{1506.02075}~[cs.LG]
\urldef\tempurl%
\url{https://arxiv.org/abs/1506.02075}
\showURL{%
\tempurl}


\bibitem[Bordes et~al\mbox{.}(2015b)]%
        {ref:simple}
\bibfield{author}{\bibinfo{person}{Antoine Bordes}, \bibinfo{person}{Nicolas Usunier}, \bibinfo{person}{Sumit Chopra}, {and} \bibinfo{person}{Jason Weston}.} \bibinfo{year}{2015}\natexlab{b}.
\newblock \bibinfo{title}{Large-scale Simple Question Answering with Memory Networks}.
\newblock
\newblock
\showeprint[arxiv]{1506.02075}~[cs.LG]
\urldef\tempurl%
\url{https://arxiv.org/abs/1506.02075}
\showURL{%
\tempurl}


\bibitem[Brown et~al\mbox{.}(2020)]%
        {ref:gpt3}
\bibfield{author}{\bibinfo{person}{Tom Brown}, \bibinfo{person}{Benjamin Mann}, \bibinfo{person}{Nick Ryder}, \bibinfo{person}{Melanie Subbiah}, \bibinfo{person}{Jared~D Kaplan}, \bibinfo{person}{Prafulla Dhariwal}, \bibinfo{person}{Arvind Neelakantan}, \bibinfo{person}{Pranav Shyam}, \bibinfo{person}{Girish Sastry}, \bibinfo{person}{Amanda Askell}, {et~al\mbox{.}}} \bibinfo{year}{2020}\natexlab{}.
\newblock \showarticletitle{Language models are few-shot learners}.
\newblock \bibinfo{journal}{\emph{Advances in neural information processing systems}}  \bibinfo{volume}{33} (\bibinfo{year}{2020}), \bibinfo{pages}{1877--1901}.
\newblock


\bibitem[Carlson et~al\mbox{.}(2010)]%
        {ref:nell}
\bibfield{author}{\bibinfo{person}{Andrew Carlson}, \bibinfo{person}{Justin Betteridge}, \bibinfo{person}{Bryan Kisiel}, \bibinfo{person}{Burr Settles}, \bibinfo{person}{Estevam R.~Hruschka Jr.}, {and} \bibinfo{person}{Tom~M. Mitchell}.} \bibinfo{year}{2010}\natexlab{}.
\newblock \showarticletitle{Toward an Architecture for Never-Ending Language Learning}. In \bibinfo{booktitle}{\emph{Proceedings of the Twenty-Fourth {AAAI} Conference on Artificial Intelligence, {AAAI} 2010, Atlanta, Georgia, USA, July 11-15, 2010}}. \bibinfo{pages}{1306--1313}.
\newblock


\bibitem[Chakraborty(2024)]%
        {ref:noname3}
\bibfield{author}{\bibinfo{person}{Abir Chakraborty}.} \bibinfo{year}{2024}\natexlab{}.
\newblock \bibinfo{title}{Multi-hop Question Answering over Knowledge Graphs using Large Language Models}.
\newblock
\newblock
\showeprint[arxiv]{2404.19234}~[cs.AI]
\urldef\tempurl%
\url{https://arxiv.org/abs/2404.19234}
\showURL{%
\tempurl}


\bibitem[Chen(2024)]%
        {ref:gfmsurvey8}
\bibfield{author}{\bibinfo{person}{Huajun Chen}.} \bibinfo{year}{2024}\natexlab{}.
\newblock \bibinfo{title}{Large Knowledge Model: Perspectives and Challenges}.
\newblock
\newblock
\showeprint[arxiv]{2312.02706}~[cs.AI]
\urldef\tempurl%
\url{https://arxiv.org/abs/2312.02706}
\showURL{%
\tempurl}


\bibitem[Chen et~al\mbox{.}(2024)]%
        {ref:llaga}
\bibfield{author}{\bibinfo{person}{Runjin Chen}, \bibinfo{person}{Tong Zhao}, \bibinfo{person}{Ajay Jaiswal}, \bibinfo{person}{Neil Shah}, {and} \bibinfo{person}{Zhangyang Wang}.} \bibinfo{year}{2024}\natexlab{}.
\newblock \bibinfo{title}{LLaGA: Large Language and Graph Assistant}.
\newblock
\newblock
\showeprint[arxiv]{2402.08170}~[cs.LG]
\urldef\tempurl%
\url{https://arxiv.org/abs/2402.08170}
\showURL{%
\tempurl}


\bibitem[Chen et~al\mbox{.}(2021)]%
        {ref:retrack}
\bibfield{author}{\bibinfo{person}{Shuang Chen}, \bibinfo{person}{Qian Liu}, \bibinfo{person}{Zhiwei Yu}, \bibinfo{person}{Chin-Yew Lin}, \bibinfo{person}{Jian-Guang Lou}, {and} \bibinfo{person}{Feng Jiang}.} \bibinfo{year}{2021}\natexlab{}.
\newblock \showarticletitle{ReTraCk: A flexible and efficient framework for knowledge base question answering}. In \bibinfo{booktitle}{\emph{Proceedings of the 59th annual meeting of the association for computational linguistics and the 11th international joint conference on natural language processing: system demonstrations}}. \bibinfo{pages}{325--336}.
\newblock


\bibitem[Cheng et~al\mbox{.}(2024)]%
        {ref:temple-mqa}
\bibfield{author}{\bibinfo{person}{Keyuan Cheng}, \bibinfo{person}{Gang Lin}, \bibinfo{person}{Haoyang Fei}, \bibinfo{person}{Yuxuan zhai}, \bibinfo{person}{Lu Yu}, \bibinfo{person}{Muhammad~Asif Ali}, \bibinfo{person}{Lijie Hu}, {and} \bibinfo{person}{Di Wang}.} \bibinfo{year}{2024}\natexlab{}.
\newblock \bibinfo{title}{Multi-hop Question Answering under Temporal Knowledge Editing}.
\newblock
\newblock
\showeprint[arxiv]{2404.00492}~[cs.CL]
\urldef\tempurl%
\url{https://arxiv.org/abs/2404.00492}
\showURL{%
\tempurl}


\bibitem[Choi et~al\mbox{.}(2023)]%
        {ref:nutrea}
\bibfield{author}{\bibinfo{person}{Hyeong~Kyu Choi}, \bibinfo{person}{Seunghun Lee}, \bibinfo{person}{Jaewon Chu}, {and} \bibinfo{person}{Hyunwoo~J. Kim}.} \bibinfo{year}{2023}\natexlab{}.
\newblock \showarticletitle{NuTrea: Neural Tree Search for Context-guided Multi-hop {KGQA}}. In \bibinfo{booktitle}{\emph{Advances in Neural Information Processing Systems 36: Annual Conference on Neural Information Processing Systems 2023, NeurIPS 2023, New Orleans, LA, USA, December 10 - 16, 2023}}.
\newblock


\bibitem[Choudhary and Reddy(2024)]%
        {ref:lark}
\bibfield{author}{\bibinfo{person}{Nurendra Choudhary} {and} \bibinfo{person}{Chandan~K. Reddy}.} \bibinfo{year}{2024}\natexlab{}.
\newblock \bibinfo{title}{Complex Logical Reasoning over Knowledge Graphs using Large Language Models}.
\newblock
\newblock
\showeprint[arxiv]{2305.01157}~[cs.LO]
\urldef\tempurl%
\url{https://arxiv.org/abs/2305.01157}
\showURL{%
\tempurl}


\bibitem[Das et~al\mbox{.}(2018)]%
        {ref:minerva}
\bibfield{author}{\bibinfo{person}{Rajarshi Das}, \bibinfo{person}{Shehzaad Dhuliawala}, \bibinfo{person}{Manzil Zaheer}, \bibinfo{person}{Luke Vilnis}, \bibinfo{person}{Ishan Durugkar}, \bibinfo{person}{Akshay Krishnamurthy}, \bibinfo{person}{Alex Smola}, {and} \bibinfo{person}{Andrew McCallum}.} \bibinfo{year}{2018}\natexlab{}.
\newblock \showarticletitle{Go for a Walk and Arrive at the Answer: Reasoning Over Paths in Knowledge Bases using Reinforcement Learning}. In \bibinfo{booktitle}{\emph{6th International Conference on Learning Representations, {ICLR} 2018, Vancouver, BC, Canada, April 30 - May 3, 2018, Conference Track Proceedings}}.
\newblock


\bibitem[Dehghan et~al\mbox{.}(2024)]%
        {ref:ewek-qa}
\bibfield{author}{\bibinfo{person}{Mohammad Dehghan}, \bibinfo{person}{Mohammad~Ali Alomrani}, \bibinfo{person}{Sunyam Bagga}, \bibinfo{person}{David Alfonso{-}Hermelo}, \bibinfo{person}{Khalil Bibi}, \bibinfo{person}{Abbas Ghaddar}, \bibinfo{person}{Yingxue Zhang}, \bibinfo{person}{Xiaoguang Li}, \bibinfo{person}{Jianye Hao}, \bibinfo{person}{Qun Liu}, \bibinfo{person}{Jimmy Lin}, \bibinfo{person}{Boxing Chen}, \bibinfo{person}{Prasanna Parthasarathi}, \bibinfo{person}{Mahdi Biparva}, {and} \bibinfo{person}{Mehdi Rezagholizadeh}.} \bibinfo{year}{2024}\natexlab{}.
\newblock \showarticletitle{{EWEK-QA} : Enhanced Web and Efficient Knowledge Graph Retrieval for Citation-based Question Answering Systems}. In \bibinfo{booktitle}{\emph{Proceedings of the 62nd Annual Meeting of the Association for Computational Linguistics (Volume 1: Long Papers), {ACL} 2024, Bangkok, Thailand, August 11-16, 2024}}. \bibinfo{pages}{14169--14187}.
\newblock


\bibitem[Deldjoo et~al\mbox{.}(2024)]%
        {ref:rrs}
\bibfield{author}{\bibinfo{person}{Yashar Deldjoo}, \bibinfo{person}{Zhankui He}, \bibinfo{person}{Julian McAuley}, \bibinfo{person}{Anton Korikov}, \bibinfo{person}{Scott Sanner}, \bibinfo{person}{Arnau Ramisa}, \bibinfo{person}{René Vidal}, \bibinfo{person}{Maheswaran Sathiamoorthy}, \bibinfo{person}{Atoosa Kasirzadeh}, {and} \bibinfo{person}{Silvia Milano}.} \bibinfo{year}{2024}\natexlab{}.
\newblock \bibinfo{title}{A Review of Modern Recommender Systems Using Generative Models (Gen-RecSys)}.
\newblock
\newblock
\showeprint[arxiv]{2404.00579}~[cs.IR]
\urldef\tempurl%
\url{https://arxiv.org/abs/2404.00579}
\showURL{%
\tempurl}


\bibitem[Delile et~al\mbox{.}(2024)]%
        {ref:noname5}
\bibfield{author}{\bibinfo{person}{Julien Delile}, \bibinfo{person}{Srayanta Mukherjee}, \bibinfo{person}{Anton~Van Pamel}, {and} \bibinfo{person}{Leonid Zhukov}.} \bibinfo{year}{2024}\natexlab{}.
\newblock \bibinfo{title}{Graph-Based Retriever Captures the Long Tail of Biomedical Knowledge}.
\newblock
\newblock
\showeprint[arxiv]{2402.12352}~[cs.CL]
\urldef\tempurl%
\url{https://arxiv.org/abs/2402.12352}
\showURL{%
\tempurl}


\bibitem[Dettmers et~al\mbox{.}(2018)]%
        {ref:wn}
\bibfield{author}{\bibinfo{person}{Tim Dettmers}, \bibinfo{person}{Pasquale Minervini}, \bibinfo{person}{Pontus Stenetorp}, {and} \bibinfo{person}{Sebastian Riedel}.} \bibinfo{year}{2018}\natexlab{}.
\newblock \showarticletitle{Convolutional 2D Knowledge Graph Embeddings}. In \bibinfo{booktitle}{\emph{Proceedings of the Thirty-Second {AAAI} Conference on Artificial Intelligence, (AAAI-18), the 30th innovative Applications of Artificial Intelligence (IAAI-18), and the 8th {AAAI} Symposium on Educational Advances in Artificial Intelligence (EAAI-18), New Orleans, Louisiana, USA, February 2-7, 2018}}. \bibinfo{pages}{1811--1818}.
\newblock


\bibitem[Devlin et~al\mbox{.}(2019)]%
        {ref:bert}
\bibfield{author}{\bibinfo{person}{Jacob Devlin}, \bibinfo{person}{Ming-Wei Chang}, \bibinfo{person}{Kenton Lee}, {and} \bibinfo{person}{Kristina Toutanova}.} \bibinfo{year}{2019}\natexlab{}.
\newblock \showarticletitle{BERT: Pre-training of Deep Bidirectional Transformers for Language Understanding}. In \bibinfo{booktitle}{\emph{Proceedings of the 2019 Conference of the North American Chapter of the Association for Computational Linguistics: Human Language Technologies, Volume 1 (Long and Short Papers)}}. \bibinfo{pages}{4171--4186}.
\newblock


\bibitem[Dong et~al\mbox{.}(2023a)]%
        {ref:skp}
\bibfield{author}{\bibinfo{person}{Guanting Dong}, \bibinfo{person}{Rumei Li}, \bibinfo{person}{Sirui Wang}, \bibinfo{person}{Yupeng Zhang}, \bibinfo{person}{Yunsen Xian}, {and} \bibinfo{person}{Weiran Xu}.} \bibinfo{year}{2023}\natexlab{a}.
\newblock \showarticletitle{Bridging the KB-Text Gap: Leveraging Structured Knowledge-aware Pre-training for {KBQA}}. In \bibinfo{booktitle}{\emph{Proceedings of the 32nd {ACM} International Conference on Information and Knowledge Management, {CIKM} 2023, Birmingham, United Kingdom, October 21-25, 2023}}. \bibinfo{pages}{3854--3859}.
\newblock


\bibitem[Dong et~al\mbox{.}(2023b)]%
        {ref:hamqa}
\bibfield{author}{\bibinfo{person}{Junnan Dong}, \bibinfo{person}{Qinggang Zhang}, \bibinfo{person}{Xiao Huang}, \bibinfo{person}{Keyu Duan}, \bibinfo{person}{Qiaoyu Tan}, {and} \bibinfo{person}{Zhimeng Jiang}.} \bibinfo{year}{2023}\natexlab{b}.
\newblock \showarticletitle{Hierarchy-Aware Multi-Hop Question Answering over Knowledge Graphs}. In \bibinfo{booktitle}{\emph{Proceedings of the {ACM} Web Conference 2023, {WWW} 2023, Austin, TX, USA, 30 April 2023 - 4 May 2023}}. \bibinfo{publisher}{{ACM}}, \bibinfo{pages}{2519--2527}.
\newblock


\bibitem[Dubey et~al\mbox{.}(2024)]%
        {ref:llama3}
\bibfield{author}{\bibinfo{person}{Abhimanyu Dubey}, \bibinfo{person}{Abhinav Jauhri}, {and} \bibinfo{person}{et al}.} \bibinfo{year}{2024}\natexlab{}.
\newblock \bibinfo{title}{The Llama 3 Herd of Models}.
\newblock
\newblock
\showeprint[arxiv]{2407.21783}~[cs.AI]
\urldef\tempurl%
\url{https://arxiv.org/abs/2407.21783}
\showURL{%
\tempurl}


\bibitem[Edge et~al\mbox{.}(2024)]%
        {ref:graphrag}
\bibfield{author}{\bibinfo{person}{Darren Edge}, \bibinfo{person}{Ha Trinh}, \bibinfo{person}{Newman Cheng}, \bibinfo{person}{Joshua Bradley}, \bibinfo{person}{Alex Chao}, \bibinfo{person}{Apurva Mody}, \bibinfo{person}{Steven Truitt}, {and} \bibinfo{person}{Jonathan Larson}.} \bibinfo{year}{2024}\natexlab{}.
\newblock \bibinfo{title}{From Local to Global: A Graph RAG Approach to Query-Focused Summarization}.
\newblock
\newblock
\showeprint[arxiv]{2404.16130}~[cs.CL]
\urldef\tempurl%
\url{https://arxiv.org/abs/2404.16130}
\showURL{%
\tempurl}


\bibitem[ElSahar et~al\mbox{.}(2018)]%
        {ref:t-rex}
\bibfield{author}{\bibinfo{person}{Hady ElSahar}, \bibinfo{person}{Pavlos Vougiouklis}, \bibinfo{person}{Arslen Remaci}, \bibinfo{person}{Christophe Gravier}, \bibinfo{person}{Jonathon~S. Hare}, \bibinfo{person}{Fr{\'{e}}d{\'{e}}rique Laforest}, {and} \bibinfo{person}{Elena Simperl}.} \bibinfo{year}{2018}\natexlab{}.
\newblock \showarticletitle{T-REx: {A} Large Scale Alignment of Natural Language with Knowledge Base Triples}. In \bibinfo{booktitle}{\emph{Proceedings of the Eleventh International Conference on Language Resources and Evaluation, {LREC} 2018, Miyazaki, Japan, May 7-12, 2018}}.
\newblock


\bibitem[Fan et~al\mbox{.}(2024a)]%
        {ref:ragsurvey1}
\bibfield{author}{\bibinfo{person}{Wenqi Fan}, \bibinfo{person}{Yujuan Ding}, \bibinfo{person}{Liangbo Ning}, \bibinfo{person}{Shijie Wang}, \bibinfo{person}{Hengyun Li}, \bibinfo{person}{Dawei Yin}, \bibinfo{person}{Tat-Seng Chua}, {and} \bibinfo{person}{Qing Li}.} \bibinfo{year}{2024}\natexlab{a}.
\newblock \bibinfo{title}{A Survey on RAG Meeting LLMs: Towards Retrieval-Augmented Large Language Models}.
\newblock
\newblock
\showeprint[arxiv]{2405.06211}~[cs.CL]
\urldef\tempurl%
\url{https://arxiv.org/abs/2405.06211}
\showURL{%
\tempurl}


\bibitem[Fan et~al\mbox{.}(2024b)]%
        {ref:gfmsurvey1}
\bibfield{author}{\bibinfo{person}{Wenqi Fan}, \bibinfo{person}{Shijie Wang}, \bibinfo{person}{Jiani Huang}, \bibinfo{person}{Zhikai Chen}, \bibinfo{person}{Yu Song}, \bibinfo{person}{Wenzhuo Tang}, \bibinfo{person}{Haitao Mao}, \bibinfo{person}{Hui Liu}, \bibinfo{person}{Xiaorui Liu}, \bibinfo{person}{Dawei Yin}, {and} \bibinfo{person}{Qing Li}.} \bibinfo{year}{2024}\natexlab{b}.
\newblock \bibinfo{title}{Graph Machine Learning in the Era of Large Language Models (LLMs)}.
\newblock
\newblock
\showeprint[arxiv]{2404.14928}~[cs.LG]
\urldef\tempurl%
\url{https://arxiv.org/abs/2404.14928}
\showURL{%
\tempurl}


\bibitem[Fang et~al\mbox{.}(2024b)]%
        {ref:dara}
\bibfield{author}{\bibinfo{person}{Haishuo Fang}, \bibinfo{person}{Xiaodan Zhu}, {and} \bibinfo{person}{Iryna Gurevych}.} \bibinfo{year}{2024}\natexlab{b}.
\newblock \bibinfo{title}{DARA: Decomposition-Alignment-Reasoning Autonomous Language Agent for Question Answering over Knowledge Graphs}.
\newblock
\newblock
\showeprint[arxiv]{2406.07080}~[cs.CL]
\urldef\tempurl%
\url{https://arxiv.org/abs/2406.07080}
\showURL{%
\tempurl}


\bibitem[Fang et~al\mbox{.}(2024a)]%
        {ref:reano}
\bibfield{author}{\bibinfo{person}{Jinyuan Fang}, \bibinfo{person}{Zaiqiao Meng}, {and} \bibinfo{person}{Craig MacDonald}.} \bibinfo{year}{2024}\natexlab{a}.
\newblock \showarticletitle{{REANO:} Optimising Retrieval-Augmented Reader Models through Knowledge Graph Generation}. In \bibinfo{booktitle}{\emph{Proceedings of the 62nd Annual Meeting of the Association for Computational Linguistics (Volume 1: Long Papers), {ACL} 2024, Bangkok, Thailand, August 11-16, 2024}}. \bibinfo{pages}{2094--2112}.
\newblock


\bibitem[Fatemi et~al\mbox{.}(2023)]%
        {ref:talk}
\bibfield{author}{\bibinfo{person}{Bahare Fatemi}, \bibinfo{person}{Jonathan Halcrow}, {and} \bibinfo{person}{Bryan Perozzi}.} \bibinfo{year}{2023}\natexlab{}.
\newblock \bibinfo{title}{Talk like a Graph: Encoding Graphs for Large Language Models}.
\newblock
\newblock
\showeprint[arxiv]{2310.04560}~[cs.LG]
\urldef\tempurl%
\url{https://arxiv.org/abs/2310.04560}
\showURL{%
\tempurl}


\bibitem[Feng et~al\mbox{.}(2023)]%
        {ref:ksl}
\bibfield{author}{\bibinfo{person}{Chao Feng}, \bibinfo{person}{Xinyu Zhang}, {and} \bibinfo{person}{Zichu Fei}.} \bibinfo{year}{2023}\natexlab{}.
\newblock \bibinfo{title}{Knowledge Solver: Teaching LLMs to Search for Domain Knowledge from Knowledge Graphs}.
\newblock
\newblock
\showeprint[arxiv]{2309.03118}~[cs.CL]
\urldef\tempurl%
\url{https://arxiv.org/abs/2309.03118}
\showURL{%
\tempurl}


\bibitem[Feng et~al\mbox{.}(2020)]%
        {ref:feng}
\bibfield{author}{\bibinfo{person}{Yanlin Feng}, \bibinfo{person}{Xinyue Chen}, \bibinfo{person}{Bill~Yuchen Lin}, \bibinfo{person}{Peifeng Wang}, \bibinfo{person}{Jun Yan}, {and} \bibinfo{person}{Xiang Ren}.} \bibinfo{year}{2020}\natexlab{}.
\newblock \showarticletitle{Scalable Multi-Hop Relational Reasoning for Knowledge-Aware Question Answering}. In \bibinfo{booktitle}{\emph{Proceedings of the 2020 Conference on Empirical Methods in Natural Language Processing, {EMNLP} 2020, Online, November 16-20, 2020}}. \bibinfo{pages}{1295--1309}.
\newblock


\bibitem[Fu et~al\mbox{.}(2020)]%
        {ref:kbqasurvey1}
\bibfield{author}{\bibinfo{person}{Bin Fu}, \bibinfo{person}{Yunqi Qiu}, \bibinfo{person}{Chengguang Tang}, \bibinfo{person}{Yang Li}, \bibinfo{person}{Haiyang Yu}, {and} \bibinfo{person}{Jian Sun}.} \bibinfo{year}{2020}\natexlab{}.
\newblock \bibinfo{title}{A Survey on Complex Question Answering over Knowledge Base: Recent Advances and Challenges}.
\newblock
\newblock
\showeprint[arxiv]{2007.13069}~[cs.CL]
\urldef\tempurl%
\url{https://arxiv.org/abs/2007.13069}
\showURL{%
\tempurl}


\bibitem[Galkin et~al\mbox{.}(2023)]%
        {ref:ultra}
\bibfield{author}{\bibinfo{person}{Mikhail Galkin}, \bibinfo{person}{Xinyu Yuan}, \bibinfo{person}{Hesham Mostafa}, \bibinfo{person}{Jian Tang}, {and} \bibinfo{person}{Zhaocheng Zhu}.} \bibinfo{year}{2023}\natexlab{}.
\newblock \showarticletitle{Towards Foundation Models for Knowledge Graph Reasoning}. In \bibinfo{booktitle}{\emph{The Twelfth International Conference on Learning Representations}}.
\newblock


\bibitem[Gao et~al\mbox{.}(2022)]%
        {ref:gge}
\bibfield{author}{\bibinfo{person}{Hanning Gao}, \bibinfo{person}{Lingfei Wu}, \bibinfo{person}{Po Hu}, \bibinfo{person}{Zhihua Wei}, \bibinfo{person}{Fangli Xu}, {and} \bibinfo{person}{Bo Long}.} \bibinfo{year}{2022}\natexlab{}.
\newblock \showarticletitle{Graph-augmented Learning to Rank for Querying Large-scale Knowledge Graph}. In \bibinfo{booktitle}{\emph{Proceedings of the 2nd Conference of the Asia-Pacific Chapter of the Association for Computational Linguistics and the 12th International Joint Conference on Natural Language Processing, {AACL/IJCNLP} 2022 - Volume 1: Long Papers, Online Only, November 20-23, 2022}}. \bibinfo{pages}{82--92}.
\newblock


\bibitem[Gao et~al\mbox{.}(2024a)]%
        {ref:gentkgqa}
\bibfield{author}{\bibinfo{person}{Yifu Gao}, \bibinfo{person}{Linbo Qiao}, \bibinfo{person}{Zhigang Kan}, \bibinfo{person}{Zhihua Wen}, \bibinfo{person}{Yongquan He}, {and} \bibinfo{person}{Dongsheng Li}.} \bibinfo{year}{2024}\natexlab{a}.
\newblock \bibinfo{title}{Two-stage Generative Question Answering on Temporal Knowledge Graph Using Large Language Models}.
\newblock
\newblock
\showeprint[arxiv]{2402.16568}~[cs.CL]
\urldef\tempurl%
\url{https://arxiv.org/abs/2402.16568}
\showURL{%
\tempurl}


\bibitem[Gao et~al\mbox{.}(2024b)]%
        {ref:ragsurvey5}
\bibfield{author}{\bibinfo{person}{Yunfan Gao}, \bibinfo{person}{Yun Xiong}, \bibinfo{person}{Xinyu Gao}, \bibinfo{person}{Kangxiang Jia}, \bibinfo{person}{Jinliu Pan}, \bibinfo{person}{Yuxi Bi}, \bibinfo{person}{Yi Dai}, \bibinfo{person}{Jiawei Sun}, \bibinfo{person}{Meng Wang}, {and} \bibinfo{person}{Haofen Wang}.} \bibinfo{year}{2024}\natexlab{b}.
\newblock \bibinfo{title}{Retrieval-Augmented Generation for Large Language Models: A Survey}.
\newblock
\newblock
\showeprint[arxiv]{2312.10997}~[cs.CL]
\urldef\tempurl%
\url{https://arxiv.org/abs/2312.10997}
\showURL{%
\tempurl}


\bibitem[Ghimire et~al\mbox{.}(2024)]%
        {ref:edu2}
\bibfield{author}{\bibinfo{person}{Aashish Ghimire}, \bibinfo{person}{James Prather}, {and} \bibinfo{person}{John Edwards}.} \bibinfo{year}{2024}\natexlab{}.
\newblock \bibinfo{title}{Generative AI in Education: A Study of Educators' Awareness, Sentiments, and Influencing Factors}.
\newblock
\newblock
\showeprint[arxiv]{2403.15586}~[cs.AI]
\urldef\tempurl%
\url{https://arxiv.org/abs/2403.15586}
\showURL{%
\tempurl}


\bibitem[Gu et~al\mbox{.}(2021)]%
        {ref:grailqa}
\bibfield{author}{\bibinfo{person}{Yu Gu}, \bibinfo{person}{Sue Kase}, \bibinfo{person}{Michelle Vanni}, \bibinfo{person}{Brian~M. Sadler}, \bibinfo{person}{Percy Liang}, \bibinfo{person}{Xifeng Yan}, {and} \bibinfo{person}{Yu Su}.} \bibinfo{year}{2021}\natexlab{}.
\newblock \showarticletitle{Beyond {I.I.D.:} Three Levels of Generalization for Question Answering on Knowledge Bases}. In \bibinfo{booktitle}{\emph{{WWW} '21: The Web Conference 2021, Virtual Event / Ljubljana, Slovenia, April 19-23, 2021}}. \bibinfo{pages}{3477--3488}.
\newblock


\bibitem[Gu and Su(2022)]%
        {ref:arcaneqa}
\bibfield{author}{\bibinfo{person}{Yu Gu} {and} \bibinfo{person}{Yu Su}.} \bibinfo{year}{2022}\natexlab{}.
\newblock \showarticletitle{ArcaneQA: Dynamic Program Induction and Contextualized Encoding for Knowledge Base Question Answering}. In \bibinfo{booktitle}{\emph{Proceedings of the 29th International Conference on Computational Linguistics}}. \bibinfo{pages}{1718--1731}.
\newblock


\bibitem[Guo et~al\mbox{.}(2023)]%
        {ref:gpt4graph}
\bibfield{author}{\bibinfo{person}{Jiayan Guo}, \bibinfo{person}{Lun Du}, \bibinfo{person}{Hengyu Liu}, \bibinfo{person}{Mengyu Zhou}, \bibinfo{person}{Xinyi He}, {and} \bibinfo{person}{Shi Han}.} \bibinfo{year}{2023}\natexlab{}.
\newblock \bibinfo{title}{GPT4Graph: Can Large Language Models Understand Graph Structured Data ? An Empirical Evaluation and Benchmarking}.
\newblock
\newblock
\showeprint[arxiv]{2305.15066}~[cs.AI]
\urldef\tempurl%
\url{https://arxiv.org/abs/2305.15066}
\showURL{%
\tempurl}


\bibitem[Guo et~al\mbox{.}(2024)]%
        {ref:kn}
\bibfield{author}{\bibinfo{person}{Tiezheng Guo}, \bibinfo{person}{Qingwen Yang}, \bibinfo{person}{Chen Wang}, \bibinfo{person}{Yanyi Liu}, \bibinfo{person}{Pan Li}, \bibinfo{person}{Jiawei Tang}, \bibinfo{person}{Dapeng Li}, {and} \bibinfo{person}{Yingyou Wen}.} \bibinfo{year}{2024}\natexlab{}.
\newblock \bibinfo{title}{KnowledgeNavigator: Leveraging Large Language Models for Enhanced Reasoning over Knowledge Graph}.
\newblock
\newblock
\showeprint[arxiv]{2312.15880}~[cs.CL]
\urldef\tempurl%
\url{https://arxiv.org/abs/2312.15880}
\showURL{%
\tempurl}


\bibitem[Gutiérrez et~al\mbox{.}(2024)]%
        {ref:hipporag}
\bibfield{author}{\bibinfo{person}{Bernal~Jiménez Gutiérrez}, \bibinfo{person}{Yiheng Shu}, \bibinfo{person}{Yu Gu}, \bibinfo{person}{Michihiro Yasunaga}, {and} \bibinfo{person}{Yu Su}.} \bibinfo{year}{2024}\natexlab{}.
\newblock \bibinfo{title}{HippoRAG: Neurobiologically Inspired Long-Term Memory for Large Language Models}.
\newblock
\newblock
\showeprint[arxiv]{2405.14831}~[cs.CL]
\urldef\tempurl%
\url{https://arxiv.org/abs/2405.14831}
\showURL{%
\tempurl}


\bibitem[Hamilton et~al\mbox{.}(2017)]%
        {ref:graphsage}
\bibfield{author}{\bibinfo{person}{William~L. Hamilton}, \bibinfo{person}{Zhitao Ying}, {and} \bibinfo{person}{Jure Leskovec}.} \bibinfo{year}{2017}\natexlab{}.
\newblock \showarticletitle{Inductive Representation Learning on Large Graphs}. In \bibinfo{booktitle}{\emph{Advances in Neural Information Processing Systems 30: Annual Conference on Neural Information Processing Systems 2017, December 4-9, 2017, Long Beach, CA, {USA}}}. \bibinfo{pages}{1024--1034}.
\newblock


\bibitem[Han et~al\mbox{.}(2023)]%
        {ref:opencsr}
\bibfield{author}{\bibinfo{person}{Zhen Han}, \bibinfo{person}{Yue Feng}, {and} \bibinfo{person}{Mingming Sun}.} \bibinfo{year}{2023}\natexlab{}.
\newblock \bibinfo{title}{A Graph-Guided Reasoning Approach for Open-ended Commonsense Question Answering}.
\newblock
\newblock
\showeprint[arxiv]{2303.10395}~[cs.CL]
\urldef\tempurl%
\url{https://arxiv.org/abs/2303.10395}
\showURL{%
\tempurl}


\bibitem[He et~al\mbox{.}(2021)]%
        {ref:nsm}
\bibfield{author}{\bibinfo{person}{Gaole He}, \bibinfo{person}{Yunshi Lan}, \bibinfo{person}{Jing Jiang}, \bibinfo{person}{Wayne~Xin Zhao}, {and} \bibinfo{person}{Ji{-}Rong Wen}.} \bibinfo{year}{2021}\natexlab{}.
\newblock \showarticletitle{Improving Multi-hop Knowledge Base Question Answering by Learning Intermediate Supervision Signals}. In \bibinfo{booktitle}{\emph{{WSDM} '21, The Fourteenth {ACM} International Conference on Web Search and Data Mining, Virtual Event, Israel, March 8-12, 2021}}. \bibinfo{pages}{553--561}.
\newblock


\bibitem[He et~al\mbox{.}(2024)]%
        {ref:g-retriever}
\bibfield{author}{\bibinfo{person}{Xiaoxin He}, \bibinfo{person}{Yijun Tian}, \bibinfo{person}{Yifei Sun}, \bibinfo{person}{Nitesh~V. Chawla}, \bibinfo{person}{Thomas Laurent}, \bibinfo{person}{Yann LeCun}, \bibinfo{person}{Xavier Bresson}, {and} \bibinfo{person}{Bryan Hooi}.} \bibinfo{year}{2024}\natexlab{}.
\newblock \bibinfo{title}{G-Retriever: Retrieval-Augmented Generation for Textual Graph Understanding and Question Answering}.
\newblock
\newblock
\showeprint[arxiv]{2402.07630}~[cs.LG]
\urldef\tempurl%
\url{https://arxiv.org/abs/2402.07630}
\showURL{%
\tempurl}


\bibitem[Himsolt(1996)]%
        {ref:gml}
\bibfield{author}{\bibinfo{person}{Michael Himsolt}.} \bibinfo{year}{1996}\natexlab{}.
\newblock \showarticletitle{GML: Graph Modelling Language}.
\newblock \bibinfo{journal}{\emph{University of Passau}} (\bibinfo{year}{1996}).
\newblock


\bibitem[Hoffart et~al\mbox{.}(2011)]%
        {ref:conll}
\bibfield{author}{\bibinfo{person}{Johannes Hoffart}, \bibinfo{person}{Mohamed~Amir Yosef}, \bibinfo{person}{Ilaria Bordino}, \bibinfo{person}{Hagen F{\"{u}}rstenau}, \bibinfo{person}{Manfred Pinkal}, \bibinfo{person}{Marc Spaniol}, \bibinfo{person}{Bilyana Taneva}, \bibinfo{person}{Stefan Thater}, {and} \bibinfo{person}{Gerhard Weikum}.} \bibinfo{year}{2011}\natexlab{}.
\newblock \showarticletitle{Robust Disambiguation of Named Entities in Text}. In \bibinfo{booktitle}{\emph{Proceedings of the 2011 Conference on Empirical Methods in Natural Language Processing, {EMNLP} 2011, 27-31 July 2011, John McIntyre Conference Centre, Edinburgh, UK, {A} meeting of SIGDAT, a Special Interest Group of the {ACL}}}. \bibinfo{pages}{782--792}.
\newblock


\bibitem[Hu et~al\mbox{.}(2024)]%
        {ref:grag}
\bibfield{author}{\bibinfo{person}{Yuntong Hu}, \bibinfo{person}{Zhihan Lei}, \bibinfo{person}{Zheng Zhang}, \bibinfo{person}{Bo Pan}, \bibinfo{person}{Chen Ling}, {and} \bibinfo{person}{Liang Zhao}.} \bibinfo{year}{2024}\natexlab{}.
\newblock \bibinfo{title}{GRAG: Graph Retrieval-Augmented Generation}.
\newblock
\newblock
\showeprint[arxiv]{2405.16506}~[cs.LG]
\urldef\tempurl%
\url{https://arxiv.org/abs/2405.16506}
\showURL{%
\tempurl}


\bibitem[Hu and Lu(2024)]%
        {ref:ragsurvey2}
\bibfield{author}{\bibinfo{person}{Yucheng Hu} {and} \bibinfo{person}{Yuxing Lu}.} \bibinfo{year}{2024}\natexlab{}.
\newblock \bibinfo{title}{RAG and RAU: A Survey on Retrieval-Augmented Language Model in Natural Language Processing}.
\newblock
\newblock
\showeprint[arxiv]{2404.19543}~[cs.CL]
\urldef\tempurl%
\url{https://arxiv.org/abs/2404.19543}
\showURL{%
\tempurl}


\bibitem[Hu et~al\mbox{.}(2022)]%
        {ref:oreolm}
\bibfield{author}{\bibinfo{person}{Ziniu Hu}, \bibinfo{person}{Yichong Xu}, \bibinfo{person}{Wenhao Yu}, \bibinfo{person}{Shuohang Wang}, \bibinfo{person}{Ziyi Yang}, \bibinfo{person}{Chenguang Zhu}, \bibinfo{person}{Kai{-}Wei Chang}, {and} \bibinfo{person}{Yizhou Sun}.} \bibinfo{year}{2022}\natexlab{}.
\newblock \showarticletitle{Empowering Language Models with Knowledge Graph Reasoning for Open-Domain Question Answering}. In \bibinfo{booktitle}{\emph{Proceedings of the 2022 Conference on Empirical Methods in Natural Language Processing, {EMNLP} 2022, Abu Dhabi, United Arab Emirates, December 7-11, 2022}}. \bibinfo{pages}{9562--9581}.
\newblock


\bibitem[Huang et~al\mbox{.}(2023b)]%
        {ref:hallucination}
\bibfield{author}{\bibinfo{person}{Lei Huang}, \bibinfo{person}{Weijiang Yu}, \bibinfo{person}{Weitao Ma}, \bibinfo{person}{Weihong Zhong}, \bibinfo{person}{Zhangyin Feng}, \bibinfo{person}{Haotian Wang}, \bibinfo{person}{Qianglong Chen}, \bibinfo{person}{Weihua Peng}, \bibinfo{person}{Xiaocheng Feng}, \bibinfo{person}{Bing Qin}, {and} \bibinfo{person}{Ting Liu}.} \bibinfo{year}{2023}\natexlab{b}.
\newblock \bibinfo{title}{A Survey on Hallucination in Large Language Models: Principles, Taxonomy, Challenges, and Open Questions}.
\newblock
\newblock
\showeprint[arxiv]{2311.05232}~[cs.CL]
\urldef\tempurl%
\url{https://arxiv.org/abs/2311.05232}
\showURL{%
\tempurl}


\bibitem[Huang and Huang(2024)]%
        {ref:ragsurvey3}
\bibfield{author}{\bibinfo{person}{Yizheng Huang} {and} \bibinfo{person}{Jimmy Huang}.} \bibinfo{year}{2024}\natexlab{}.
\newblock \bibinfo{title}{A Survey on Retrieval-Augmented Text Generation for Large Language Models}.
\newblock
\newblock
\showeprint[arxiv]{2404.10981}~[cs.IR]
\urldef\tempurl%
\url{https://arxiv.org/abs/2404.10981}
\showURL{%
\tempurl}


\bibitem[Huang et~al\mbox{.}(2023a)]%
        {ref:mvp-tuning}
\bibfield{author}{\bibinfo{person}{Yongfeng Huang}, \bibinfo{person}{Yanyang Li}, \bibinfo{person}{Yichong Xu}, \bibinfo{person}{Lin Zhang}, \bibinfo{person}{Ruyi Gan}, \bibinfo{person}{Jiaxing Zhang}, {and} \bibinfo{person}{Liwei Wang}.} \bibinfo{year}{2023}\natexlab{a}.
\newblock \showarticletitle{MVP-Tuning: Multi-View Knowledge Retrieval with Prompt Tuning for Commonsense Reasoning}. In \bibinfo{booktitle}{\emph{Proceedings of the 61st Annual Meeting of the Association for Computational Linguistics (Volume 1: Long Papers), {ACL} 2023, Toronto, Canada, July 9-14, 2023}}. \bibinfo{pages}{13417--13432}.
\newblock


\bibitem[Hwang et~al\mbox{.}(2021)]%
        {ref:atomic2}
\bibfield{author}{\bibinfo{person}{Jena~D. Hwang}, \bibinfo{person}{Chandra Bhagavatula}, \bibinfo{person}{Ronan~Le Bras}, \bibinfo{person}{Jeff Da}, \bibinfo{person}{Keisuke Sakaguchi}, \bibinfo{person}{Antoine Bosselut}, {and} \bibinfo{person}{Yejin Choi}.} \bibinfo{year}{2021}\natexlab{}.
\newblock \showarticletitle{(Comet-) Atomic 2020: On Symbolic and Neural Commonsense Knowledge Graphs}. In \bibinfo{booktitle}{\emph{Thirty-Fifth {AAAI} Conference on Artificial Intelligence, {AAAI} 2021, Thirty-Third Conference on Innovative Applications of Artificial Intelligence, {IAAI} 2021, The Eleventh Symposium on Educational Advances in Artificial Intelligence, {EAAI} 2021, Virtual Event, February 2-9, 2021}}. \bibinfo{pages}{6384--6392}.
\newblock


\bibitem[Izacard and Grave(2021)]%
        {ref:fid}
\bibfield{author}{\bibinfo{person}{Gautier Izacard} {and} \bibinfo{person}{Edouard Grave}.} \bibinfo{year}{2021}\natexlab{}.
\newblock \showarticletitle{Leveraging Passage Retrieval with Generative Models for Open Domain Question Answering}. In \bibinfo{booktitle}{\emph{Proceedings of the 16th Conference of the European Chapter of the Association for Computational Linguistics: Main Volume, {EACL} 2021, Online, April 19 - 23, 2021}}. \bibinfo{pages}{874--880}.
\newblock


\bibitem[Jafari et~al\mbox{.}(2021)]%
        {ref:lsh}
\bibfield{author}{\bibinfo{person}{Omid Jafari}, \bibinfo{person}{Preeti Maurya}, \bibinfo{person}{Parth Nagarkar}, \bibinfo{person}{Khandker~Mushfiqul Islam}, {and} \bibinfo{person}{Chidambaram Crushev}.} \bibinfo{year}{2021}\natexlab{}.
\newblock \bibinfo{title}{A Survey on Locality Sensitive Hashing Algorithms and their Applications}.
\newblock
\newblock
\showeprint[arxiv]{2102.08942}~[cs.DB]
\urldef\tempurl%
\url{https://arxiv.org/abs/2102.08942}
\showURL{%
\tempurl}


\bibitem[Jiang et~al\mbox{.}(2023a)]%
        {ref:structgpt}
\bibfield{author}{\bibinfo{person}{Jinhao Jiang}, \bibinfo{person}{Kun Zhou}, \bibinfo{person}{Zican Dong}, \bibinfo{person}{Keming Ye}, \bibinfo{person}{Xin Zhao}, {and} \bibinfo{person}{Ji{-}Rong Wen}.} \bibinfo{year}{2023}\natexlab{a}.
\newblock \showarticletitle{StructGPT: {A} General Framework for Large Language Model to Reason over Structured Data}. In \bibinfo{booktitle}{\emph{Proceedings of the 2023 Conference on Empirical Methods in Natural Language Processing, {EMNLP} 2023, Singapore, December 6-10, 2023}}. \bibinfo{pages}{9237--9251}.
\newblock


\bibitem[Jiang et~al\mbox{.}(2022)]%
        {ref:safe}
\bibfield{author}{\bibinfo{person}{Jinhao Jiang}, \bibinfo{person}{Kun Zhou}, \bibinfo{person}{Ji{-}Rong Wen}, {and} \bibinfo{person}{Wayne~Xin Zhao}.} \bibinfo{year}{2022}\natexlab{}.
\newblock \showarticletitle{{\textdollar}Great Truths are Always Simple: {\textdollar} {A} Rather Simple Knowledge Encoder for Enhancing the Commonsense Reasoning Capacity of Pre-Trained Models}. In \bibinfo{booktitle}{\emph{Findings of the Association for Computational Linguistics: {NAACL} 2022, Seattle, WA, United States, July 10-15, 2022}}. \bibinfo{pages}{1730--1741}.
\newblock


\bibitem[Jiang et~al\mbox{.}(2024b)]%
        {ref:kg-agent}
\bibfield{author}{\bibinfo{person}{Jinhao Jiang}, \bibinfo{person}{Kun Zhou}, \bibinfo{person}{Wayne~Xin Zhao}, \bibinfo{person}{Yang Song}, \bibinfo{person}{Chen Zhu}, \bibinfo{person}{Hengshu Zhu}, {and} \bibinfo{person}{Ji-Rong Wen}.} \bibinfo{year}{2024}\natexlab{b}.
\newblock \bibinfo{title}{KG-Agent: An Efficient Autonomous Agent Framework for Complex Reasoning over Knowledge Graph}.
\newblock
\newblock
\showeprint[arxiv]{2402.11163}~[cs.CL]
\urldef\tempurl%
\url{https://arxiv.org/abs/2402.11163}
\showURL{%
\tempurl}


\bibitem[Jiang et~al\mbox{.}(2023b)]%
        {ref:unikgqa}
\bibfield{author}{\bibinfo{person}{Jinhao Jiang}, \bibinfo{person}{Kun Zhou}, \bibinfo{person}{Xin Zhao}, {and} \bibinfo{person}{Ji{-}Rong Wen}.} \bibinfo{year}{2023}\natexlab{b}.
\newblock \showarticletitle{UniKGQA: Unified Retrieval and Reasoning for Solving Multi-hop Question Answering Over Knowledge Graph}. In \bibinfo{booktitle}{\emph{The Eleventh International Conference on Learning Representations, {ICLR} 2023, Kigali, Rwanda, May 1-5, 2023}}.
\newblock


\bibitem[Jiang et~al\mbox{.}(2023c)]%
        {ref:unikqa}
\bibfield{author}{\bibinfo{person}{Jinhao Jiang}, \bibinfo{person}{Kun Zhou}, \bibinfo{person}{Xin Zhao}, {and} \bibinfo{person}{Ji{-}Rong Wen}.} \bibinfo{year}{2023}\natexlab{c}.
\newblock \showarticletitle{UniKGQA: Unified Retrieval and Reasoning for Solving Multi-hop Question Answering Over Knowledge Graph}. In \bibinfo{booktitle}{\emph{The Eleventh International Conference on Learning Representations, {ICLR} 2023, Kigali, Rwanda, May 1-5, 2023}}.
\newblock


\bibitem[Jiang et~al\mbox{.}(2019)]%
        {ref:freebaseqa}
\bibfield{author}{\bibinfo{person}{Kelvin Jiang}, \bibinfo{person}{Dekun Wu}, {and} \bibinfo{person}{Hui Jiang}.} \bibinfo{year}{2019}\natexlab{}.
\newblock \showarticletitle{FreebaseQA: {A} New Factoid {QA} Data Set Matching Trivia-Style Question-Answer Pairs with Freebase}. In \bibinfo{booktitle}{\emph{Proceedings of the 2019 Conference of the North American Chapter of the Association for Computational Linguistics: Human Language Technologies, {NAACL-HLT} 2019, Minneapolis, MN, USA, June 2-7, 2019, Volume 1 (Long and Short Papers)}}. \bibinfo{pages}{318--323}.
\newblock


\bibitem[Jiang et~al\mbox{.}(2024a)]%
        {ref:hykge}
\bibfield{author}{\bibinfo{person}{Xinke Jiang}, \bibinfo{person}{Ruizhe Zhang}, \bibinfo{person}{Yongxin Xu}, \bibinfo{person}{Rihong Qiu}, \bibinfo{person}{Yue Fang}, \bibinfo{person}{Zhiyuan Wang}, \bibinfo{person}{Jinyi Tang}, \bibinfo{person}{Hongxin Ding}, \bibinfo{person}{Xu Chu}, \bibinfo{person}{Junfeng Zhao}, {and} \bibinfo{person}{Yasha Wang}.} \bibinfo{year}{2024}\natexlab{a}.
\newblock \bibinfo{title}{HyKGE: A Hypothesis Knowledge Graph Enhanced Framework for Accurate and Reliable Medical LLMs Responses}.
\newblock
\newblock
\showeprint[arxiv]{2312.15883}~[cs.CL]
\urldef\tempurl%
\url{https://arxiv.org/abs/2312.15883}
\showURL{%
\tempurl}


\bibitem[Jin et~al\mbox{.}(2024a)]%
        {ref:gfmsurvey7}
\bibfield{author}{\bibinfo{person}{Bowen Jin}, \bibinfo{person}{Gang Liu}, \bibinfo{person}{Chi Han}, \bibinfo{person}{Meng Jiang}, \bibinfo{person}{Heng Ji}, {and} \bibinfo{person}{Jiawei Han}.} \bibinfo{year}{2024}\natexlab{a}.
\newblock \bibinfo{title}{Large Language Models on Graphs: A Comprehensive Survey}.
\newblock
\newblock
\showeprint[arxiv]{2312.02783}~[cs.CL]
\urldef\tempurl%
\url{https://arxiv.org/abs/2312.02783}
\showURL{%
\tempurl}


\bibitem[Jin et~al\mbox{.}(2024b)]%
        {ref:graphcot}
\bibfield{author}{\bibinfo{person}{Bowen Jin}, \bibinfo{person}{Chulin Xie}, \bibinfo{person}{Jiawei Zhang}, \bibinfo{person}{Kashob~Kumar Roy}, \bibinfo{person}{Yu Zhang}, \bibinfo{person}{Zheng Li}, \bibinfo{person}{Ruirui Li}, \bibinfo{person}{Xianfeng Tang}, \bibinfo{person}{Suhang Wang}, \bibinfo{person}{Yu Meng}, {and} \bibinfo{person}{Jiawei Han}.} \bibinfo{year}{2024}\natexlab{b}.
\newblock \bibinfo{title}{Graph Chain-of-Thought: Augmenting Large Language Models by Reasoning on Graphs}.
\newblock
\newblock
\showeprint[arxiv]{2404.07103}~[cs.CL]
\urldef\tempurl%
\url{https://arxiv.org/abs/2404.07103}
\showURL{%
\tempurl}


\bibitem[Jin et~al\mbox{.}(2020)]%
        {ref:medqa}
\bibfield{author}{\bibinfo{person}{Di Jin}, \bibinfo{person}{Eileen Pan}, \bibinfo{person}{Nassim Oufattole}, \bibinfo{person}{Wei-Hung Weng}, \bibinfo{person}{Hanyi Fang}, {and} \bibinfo{person}{Peter Szolovits}.} \bibinfo{year}{2020}\natexlab{}.
\newblock \bibinfo{title}{What Disease does this Patient Have? A Large-scale Open Domain Question Answering Dataset from Medical Exams}.
\newblock
\newblock
\showeprint[arxiv]{2009.13081}~[cs.CL]
\urldef\tempurl%
\url{https://arxiv.org/abs/2009.13081}
\showURL{%
\tempurl}


\bibitem[Joshi et~al\mbox{.}(2017)]%
        {ref:triviaqa}
\bibfield{author}{\bibinfo{person}{Mandar Joshi}, \bibinfo{person}{Eunsol Choi}, \bibinfo{person}{Daniel~S. Weld}, {and} \bibinfo{person}{Luke Zettlemoyer}.} \bibinfo{year}{2017}\natexlab{}.
\newblock \showarticletitle{TriviaQA: {A} Large Scale Distantly Supervised Challenge Dataset for Reading Comprehension}. In \bibinfo{booktitle}{\emph{Proceedings of the 55th Annual Meeting of the Association for Computational Linguistics, {ACL} 2017, Vancouver, Canada, July 30 - August 4, Volume 1: Long Papers}}. \bibinfo{pages}{1601--1611}.
\newblock


\bibitem[Karpukhin et~al\mbox{.}(2020)]%
        {ref:dpr}
\bibfield{author}{\bibinfo{person}{Vladimir Karpukhin}, \bibinfo{person}{Barlas Oguz}, \bibinfo{person}{Sewon Min}, \bibinfo{person}{Patrick S.~H. Lewis}, \bibinfo{person}{Ledell Wu}, \bibinfo{person}{Sergey Edunov}, \bibinfo{person}{Danqi Chen}, {and} \bibinfo{person}{Wen{-}tau Yih}.} \bibinfo{year}{2020}\natexlab{}.
\newblock \showarticletitle{Dense Passage Retrieval for Open-Domain Question Answering}. In \bibinfo{booktitle}{\emph{Proceedings of the 2020 Conference on Empirical Methods in Natural Language Processing, {EMNLP} 2020, Online, November 16-20, 2020}}. \bibinfo{pages}{6769--6781}.
\newblock


\bibitem[Kashyap et~al\mbox{.}(2024)]%
        {ref:kashyap2024knowledge}
\bibfield{author}{\bibinfo{person}{Sohum Kashyap} {et~al\mbox{.}}} \bibinfo{year}{2024}\natexlab{}.
\newblock \showarticletitle{Knowledge Graph Assisted Large Language Models}.
\newblock  (\bibinfo{year}{2024}).
\newblock


\bibitem[Kim et~al\mbox{.}(2023a)]%
        {ref:kg-gpt}
\bibfield{author}{\bibinfo{person}{Jiho Kim}, \bibinfo{person}{Yeonsu Kwon}, \bibinfo{person}{Yohan Jo}, {and} \bibinfo{person}{Edward Choi}.} \bibinfo{year}{2023}\natexlab{a}.
\newblock \showarticletitle{{KG-GPT:} {A} General Framework for Reasoning on Knowledge Graphs Using Large Language Models}. In \bibinfo{booktitle}{\emph{Findings of the Association for Computational Linguistics: {EMNLP} 2023, Singapore, December 6-10, 2023}}. \bibinfo{pages}{9410--9421}.
\newblock


\bibitem[Kim and Min(2024)]%
        {ref:legal}
\bibfield{author}{\bibinfo{person}{Jaewoong Kim} {and} \bibinfo{person}{Moohong Min}.} \bibinfo{year}{2024}\natexlab{}.
\newblock \bibinfo{title}{From RAG to QA-RAG: Integrating Generative AI for Pharmaceutical Regulatory Compliance Process}.
\newblock
\newblock
\showeprint[arxiv]{2402.01717}~[cs.CL]
\urldef\tempurl%
\url{https://arxiv.org/abs/2402.01717}
\showURL{%
\tempurl}


\bibitem[Kim et~al\mbox{.}(2023b)]%
        {ref:factkg}
\bibfield{author}{\bibinfo{person}{Jiho Kim}, \bibinfo{person}{Sungjin Park}, \bibinfo{person}{Yeonsu Kwon}, \bibinfo{person}{Yohan Jo}, \bibinfo{person}{James Thorne}, {and} \bibinfo{person}{Edward Choi}.} \bibinfo{year}{2023}\natexlab{b}.
\newblock \showarticletitle{FactKG: Fact Verification via Reasoning on Knowledge Graphs}. In \bibinfo{booktitle}{\emph{Proceedings of the 61st Annual Meeting of the Association for Computational Linguistics (Volume 1: Long Papers), {ACL} 2023, Toronto, Canada, July 9-14, 2023}}. \bibinfo{pages}{16190--16206}.
\newblock


\bibitem[Kipf and Welling(2017)]%
        {ref:gcn}
\bibfield{author}{\bibinfo{person}{Thomas~N. Kipf} {and} \bibinfo{person}{Max Welling}.} \bibinfo{year}{2017}\natexlab{}.
\newblock \showarticletitle{Semi-Supervised Classification with Graph Convolutional Networks}. In \bibinfo{booktitle}{\emph{5th International Conference on Learning Representations, {ICLR} 2017, Toulon, France, April 24-26, 2017, Conference Track Proceedings}}.
\newblock


\bibitem[Kwiatkowski et~al\mbox{.}(2019)]%
        {ref:nq}
\bibfield{author}{\bibinfo{person}{Tom Kwiatkowski}, \bibinfo{person}{Jennimaria Palomaki}, \bibinfo{person}{Olivia Redfield}, \bibinfo{person}{Michael Collins}, \bibinfo{person}{Ankur~P. Parikh}, \bibinfo{person}{Chris Alberti}, \bibinfo{person}{Danielle Epstein}, \bibinfo{person}{Illia Polosukhin}, \bibinfo{person}{Jacob Devlin}, \bibinfo{person}{Kenton Lee}, \bibinfo{person}{Kristina Toutanova}, \bibinfo{person}{Llion Jones}, \bibinfo{person}{Matthew Kelcey}, \bibinfo{person}{Ming{-}Wei Chang}, \bibinfo{person}{Andrew~M. Dai}, \bibinfo{person}{Jakob Uszkoreit}, \bibinfo{person}{Quoc Le}, {and} \bibinfo{person}{Slav Petrov}.} \bibinfo{year}{2019}\natexlab{}.
\newblock \showarticletitle{Natural Questions: a Benchmark for Question Answering Research}.
\newblock \bibinfo{journal}{\emph{Trans. Assoc. Comput. Linguistics}}  \bibinfo{volume}{7} (\bibinfo{year}{2019}), \bibinfo{pages}{452--466}.
\newblock


\bibitem[Lan et~al\mbox{.}(2021)]%
        {ref:kbqasurvey3}
\bibfield{author}{\bibinfo{person}{Yunshi Lan}, \bibinfo{person}{Gaole He}, \bibinfo{person}{Jinhao Jiang}, \bibinfo{person}{Jing Jiang}, \bibinfo{person}{Wayne~Xin Zhao}, {and} \bibinfo{person}{Ji{-}Rong Wen}.} \bibinfo{year}{2021}\natexlab{}.
\newblock \showarticletitle{A Survey on Complex Knowledge Base Question Answering: Methods, Challenges and Solutions}. In \bibinfo{booktitle}{\emph{Proceedings of the Thirtieth International Joint Conference on Artificial Intelligence, {IJCAI} 2021, Virtual Event / Montreal, Canada, 19-27 August 2021}}. \bibinfo{pages}{4483--4491}.
\newblock


\bibitem[Lan et~al\mbox{.}(2023)]%
        {ref:kbqasurvey2}
\bibfield{author}{\bibinfo{person}{Yunshi Lan}, \bibinfo{person}{Gaole He}, \bibinfo{person}{Jinhao Jiang}, \bibinfo{person}{Jing Jiang}, \bibinfo{person}{Wayne~Xin Zhao}, {and} \bibinfo{person}{Ji{-}Rong Wen}.} \bibinfo{year}{2023}\natexlab{}.
\newblock \showarticletitle{Complex Knowledge Base Question Answering: {A} Survey}.
\newblock \bibinfo{journal}{\emph{{IEEE} Trans. Knowl. Data Eng.}} \bibinfo{volume}{35}, \bibinfo{number}{11} (\bibinfo{year}{2023}), \bibinfo{pages}{11196--11215}.
\newblock


\bibitem[Lan and Jiang(2020)]%
        {ref:noname8}
\bibfield{author}{\bibinfo{person}{Yunshi Lan} {and} \bibinfo{person}{Jing Jiang}.} \bibinfo{year}{2020}\natexlab{}.
\newblock \showarticletitle{Query Graph Generation for Answering Multi-hop Complex Questions from Knowledge Bases}. In \bibinfo{booktitle}{\emph{Proceedings of the 58th Annual Meeting of the Association for Computational Linguistics, {ACL} 2020, Online, July 5-10, 2020}}. \bibinfo{pages}{969--974}.
\newblock


\bibitem[Lester et~al\mbox{.}(2021)]%
        {ref:prompttuning}
\bibfield{author}{\bibinfo{person}{Brian Lester}, \bibinfo{person}{Rami Al{-}Rfou}, {and} \bibinfo{person}{Noah Constant}.} \bibinfo{year}{2021}\natexlab{}.
\newblock \showarticletitle{The Power of Scale for Parameter-Efficient Prompt Tuning}. In \bibinfo{booktitle}{\emph{Proceedings of the 2021 Conference on Empirical Methods in Natural Language Processing, {EMNLP} 2021, Virtual Event / Punta Cana, Dominican Republic, 7-11 November, 2021}}. \bibinfo{pages}{3045--3059}.
\newblock


\bibitem[Li et~al\mbox{.}(2024e)]%
        {ref:dalk}
\bibfield{author}{\bibinfo{person}{Dawei Li}, \bibinfo{person}{Shu Yang}, \bibinfo{person}{Zhen Tan}, \bibinfo{person}{Jae~Young Baik}, \bibinfo{person}{Sukwon Yun}, \bibinfo{person}{Joseph Lee}, \bibinfo{person}{Aaron Chacko}, \bibinfo{person}{Bojian Hou}, \bibinfo{person}{Duy Duong-Tran}, \bibinfo{person}{Ying Ding}, \bibinfo{person}{Huan Liu}, \bibinfo{person}{Li Shen}, {and} \bibinfo{person}{Tianlong Chen}.} \bibinfo{year}{2024}\natexlab{e}.
\newblock \bibinfo{title}{DALK: Dynamic Co-Augmentation of LLMs and KG to answer Alzheimer's Disease Questions with Scientific Literature}.
\newblock
\newblock
\showeprint[arxiv]{2405.04819}~[cs.CL]
\urldef\tempurl%
\url{https://arxiv.org/abs/2405.04819}
\showURL{%
\tempurl}


\bibitem[Li et~al\mbox{.}(2023)]%
        {ref:noname2}
\bibfield{author}{\bibinfo{person}{Shiyang Li}, \bibinfo{person}{Yifan Gao}, \bibinfo{person}{Haoming Jiang}, \bibinfo{person}{Qingyu Yin}, \bibinfo{person}{Zheng Li}, \bibinfo{person}{Xifeng Yan}, \bibinfo{person}{Chao Zhang}, {and} \bibinfo{person}{Bing Yin}.} \bibinfo{year}{2023}\natexlab{}.
\newblock \showarticletitle{Graph Reasoning for Question Answering with Triplet Retrieval}. In \bibinfo{booktitle}{\emph{Findings of the Association for Computational Linguistics: {ACL} 2023, Toronto, Canada, July 9-14, 2023}}. \bibinfo{pages}{3366--3375}.
\newblock


\bibitem[Li and Liang(2021)]%
        {ref:prefix-tuning}
\bibfield{author}{\bibinfo{person}{Xiang~Lisa Li} {and} \bibinfo{person}{Percy Liang}.} \bibinfo{year}{2021}\natexlab{}.
\newblock \showarticletitle{Prefix-Tuning: Optimizing Continuous Prompts for Generation}. In \bibinfo{booktitle}{\emph{Proceedings of the 59th Annual Meeting of the Association for Computational Linguistics and the 11th International Joint Conference on Natural Language Processing, {ACL/IJCNLP} 2021, (Volume 1: Long Papers), Virtual Event, August 1-6, 2021}}. \bibinfo{pages}{4582--4597}.
\newblock


\bibitem[Li et~al\mbox{.}(2024c)]%
        {ref:gfmsurvey4}
\bibfield{author}{\bibinfo{person}{Yuhan Li}, \bibinfo{person}{Zhixun Li}, \bibinfo{person}{Peisong Wang}, \bibinfo{person}{Jia Li}, \bibinfo{person}{Xiangguo Sun}, \bibinfo{person}{Hong Cheng}, {and} \bibinfo{person}{Jeffrey~Xu Yu}.} \bibinfo{year}{2024}\natexlab{c}.
\newblock \bibinfo{title}{A Survey of Graph Meets Large Language Model: Progress and Future Directions}.
\newblock
\newblock
\showeprint[arxiv]{2311.12399}~[cs.LG]
\urldef\tempurl%
\url{https://arxiv.org/abs/2311.12399}
\showURL{%
\tempurl}


\bibitem[Li et~al\mbox{.}(2024d)]%
        {ref:finalcialgpt2}
\bibfield{author}{\bibinfo{person}{Yinheng Li}, \bibinfo{person}{Shaofei Wang}, \bibinfo{person}{Han Ding}, {and} \bibinfo{person}{Hang Chen}.} \bibinfo{year}{2024}\natexlab{d}.
\newblock \bibinfo{title}{Large Language Models in Finance: A Survey}.
\newblock
\newblock
\showeprint[arxiv]{2311.10723}~[q-fin.GN]
\urldef\tempurl%
\url{https://arxiv.org/abs/2311.10723}
\showURL{%
\tempurl}


\bibitem[Li et~al\mbox{.}(2024f)]%
        {ref:noname6}
\bibfield{author}{\bibinfo{person}{Yihao Li}, \bibinfo{person}{Ru Zhang}, {and} \bibinfo{person}{Jianyi Liu}.} \bibinfo{year}{2024}\natexlab{f}.
\newblock \bibinfo{title}{An Enhanced Prompt-Based LLM Reasoning Scheme via Knowledge Graph-Integrated Collaboration}.
\newblock
\newblock
\showeprint[arxiv]{2402.04978}~[cs.CL]
\urldef\tempurl%
\url{https://arxiv.org/abs/2402.04978}
\showURL{%
\tempurl}


\bibitem[Li et~al\mbox{.}(2024a)]%
        {ref:unioqa}
\bibfield{author}{\bibinfo{person}{Zhuoyang Li}, \bibinfo{person}{Liran Deng}, \bibinfo{person}{Hui Liu}, \bibinfo{person}{Qiaoqiao Liu}, {and} \bibinfo{person}{Junzhao Du}.} \bibinfo{year}{2024}\natexlab{a}.
\newblock \bibinfo{title}{UniOQA: A Unified Framework for Knowledge Graph Question Answering with Large Language Models}.
\newblock
\newblock
\showeprint[arxiv]{2406.02110}~[cs.CL]
\urldef\tempurl%
\url{https://arxiv.org/abs/2406.02110}
\showURL{%
\tempurl}


\bibitem[Li et~al\mbox{.}(2024b)]%
        {ref:gnn-net}
\bibfield{author}{\bibinfo{person}{Zijian Li}, \bibinfo{person}{Qingyan Guo}, \bibinfo{person}{Jiawei Shao}, \bibinfo{person}{Lei Song}, \bibinfo{person}{Jiang Bian}, \bibinfo{person}{Jun Zhang}, {and} \bibinfo{person}{Rui Wang}.} \bibinfo{year}{2024}\natexlab{b}.
\newblock \bibinfo{title}{Graph Neural Network Enhanced Retrieval for Question Answering of LLMs}.
\newblock
\newblock
\showeprint[arxiv]{2406.06572}~[cs.CL]
\urldef\tempurl%
\url{https://arxiv.org/abs/2406.06572}
\showURL{%
\tempurl}


\bibitem[Lin et~al\mbox{.}(2019)]%
        {ref:kagnet}
\bibfield{author}{\bibinfo{person}{Bill~Yuchen Lin}, \bibinfo{person}{Xinyue Chen}, \bibinfo{person}{Jamin Chen}, {and} \bibinfo{person}{Xiang Ren}.} \bibinfo{year}{2019}\natexlab{}.
\newblock \showarticletitle{KagNet: Knowledge-Aware Graph Networks for Commonsense Reasoning}. In \bibinfo{booktitle}{\emph{Proceedings of the 2019 Conference on Empirical Methods in Natural Language Processing and the 9th International Joint Conference on Natural Language Processing, {EMNLP-IJCNLP} 2019, Hong Kong, China, November 3-7, 2019}}. \bibinfo{pages}{2829--2839}.
\newblock


\bibitem[Lin et~al\mbox{.}(2021)]%
        {ref:riddle}
\bibfield{author}{\bibinfo{person}{Bill~Yuchen Lin}, \bibinfo{person}{Ziyi Wu}, \bibinfo{person}{Yichi Yang}, \bibinfo{person}{Dong{-}Ho Lee}, {and} \bibinfo{person}{Xiang Ren}.} \bibinfo{year}{2021}\natexlab{}.
\newblock \showarticletitle{RiddleSense: Reasoning about Riddle Questions Featuring Linguistic Creativity and Commonsense Knowledge}. In \bibinfo{booktitle}{\emph{Findings of the Association for Computational Linguistics: {ACL/IJCNLP} 2021, Online Event, August 1-6, 2021}} \emph{(\bibinfo{series}{Findings of {ACL}}, Vol.~\bibinfo{volume}{{ACL/IJCNLP} 2021})}. \bibinfo{pages}{1504--1515}.
\newblock


\bibitem[Liu et~al\mbox{.}(2024e)]%
        {ref:etd}
\bibfield{author}{\bibinfo{person}{Guangyi Liu}, \bibinfo{person}{Yongqi Zhang}, \bibinfo{person}{Yong Li}, {and} \bibinfo{person}{Quanming Yao}.} \bibinfo{year}{2024}\natexlab{e}.
\newblock \bibinfo{title}{Explore then Determine: A GNN-LLM Synergy Framework for Reasoning over Knowledge Graph}.
\newblock
\newblock
\showeprint[arxiv]{2406.01145}~[cs.CL]
\urldef\tempurl%
\url{https://arxiv.org/abs/2406.01145}
\showURL{%
\tempurl}


\bibitem[Liu and Singh(2004)]%
        {ref:conceptnet}
\bibfield{author}{\bibinfo{person}{H Liu} {and} \bibinfo{person}{P Singh}.} \bibinfo{year}{2004}\natexlab{}.
\newblock \showarticletitle{ConceptNet—a practical commonsense reasoning tool-kit}.
\newblock \bibinfo{journal}{\emph{BT technology journal}} \bibinfo{volume}{22}, \bibinfo{number}{4} (\bibinfo{year}{2004}), \bibinfo{pages}{211--226}.
\newblock


\bibitem[Liu et~al\mbox{.}(2024b)]%
        {ref:kelp}
\bibfield{author}{\bibinfo{person}{Haochen Liu}, \bibinfo{person}{Song Wang}, \bibinfo{person}{Yaochen Zhu}, \bibinfo{person}{Yushun Dong}, {and} \bibinfo{person}{Jundong Li}.} \bibinfo{year}{2024}\natexlab{b}.
\newblock \showarticletitle{Knowledge Graph-Enhanced Large Language Models via Path Selection}. In \bibinfo{booktitle}{\emph{Findings of the Association for Computational Linguistics, {ACL} 2024, Bangkok, Thailand and virtual meeting, August 11-16, 2024}}. \bibinfo{pages}{6311--6321}.
\newblock


\bibitem[Liu et~al\mbox{.}(2024d)]%
        {ref:gfmsurvey3}
\bibfield{author}{\bibinfo{person}{Jiawei Liu}, \bibinfo{person}{Cheng Yang}, \bibinfo{person}{Zhiyuan Lu}, \bibinfo{person}{Junze Chen}, \bibinfo{person}{Yibo Li}, \bibinfo{person}{Mengmei Zhang}, \bibinfo{person}{Ting Bai}, \bibinfo{person}{Yuan Fang}, \bibinfo{person}{Lichao Sun}, \bibinfo{person}{Philip~S. Yu}, {and} \bibinfo{person}{Chuan Shi}.} \bibinfo{year}{2024}\natexlab{d}.
\newblock \bibinfo{title}{Towards Graph Foundation Models: A Survey and Beyond}.
\newblock
\newblock
\showeprint[arxiv]{2310.11829}~[cs.LG]
\urldef\tempurl%
\url{https://arxiv.org/abs/2310.11829}
\showURL{%
\tempurl}


\bibitem[Liu et~al\mbox{.}(2024c)]%
        {ref:med2}
\bibfield{author}{\bibinfo{person}{Lei Liu}, \bibinfo{person}{Xiaoyan Yang}, \bibinfo{person}{Junchi Lei}, \bibinfo{person}{Xiaoyang Liu}, \bibinfo{person}{Yue Shen}, \bibinfo{person}{Zhiqiang Zhang}, \bibinfo{person}{Peng Wei}, \bibinfo{person}{Jinjie Gu}, \bibinfo{person}{Zhixuan Chu}, \bibinfo{person}{Zhan Qin}, {and} \bibinfo{person}{Kui Ren}.} \bibinfo{year}{2024}\natexlab{c}.
\newblock \bibinfo{title}{A Survey on Medical Large Language Models: Technology, Application, Trustworthiness, and Future Directions}.
\newblock
\newblock
\showeprint[arxiv]{2406.03712}~[cs.CL]
\urldef\tempurl%
\url{https://arxiv.org/abs/2406.03712}
\showURL{%
\tempurl}


\bibitem[Liu et~al\mbox{.}(2024a)]%
        {ref:lostin}
\bibfield{author}{\bibinfo{person}{Nelson~F. Liu}, \bibinfo{person}{Kevin Lin}, \bibinfo{person}{John Hewitt}, \bibinfo{person}{Ashwin Paranjape}, \bibinfo{person}{Michele Bevilacqua}, \bibinfo{person}{Fabio Petroni}, {and} \bibinfo{person}{Percy Liang}.} \bibinfo{year}{2024}\natexlab{a}.
\newblock \showarticletitle{Lost in the Middle: How Language Models Use Long Contexts}.
\newblock \bibinfo{journal}{\emph{Trans. Assoc. Comput. Linguistics}}  \bibinfo{volume}{12} (\bibinfo{year}{2024}), \bibinfo{pages}{157--173}.
\newblock


\bibitem[Liu et~al\mbox{.}(2022)]%
        {ref:p-tuning2}
\bibfield{author}{\bibinfo{person}{Xiao Liu}, \bibinfo{person}{Kaixuan Ji}, \bibinfo{person}{Yicheng Fu}, \bibinfo{person}{Weng Tam}, \bibinfo{person}{Zhengxiao Du}, \bibinfo{person}{Zhilin Yang}, {and} \bibinfo{person}{Jie Tang}.} \bibinfo{year}{2022}\natexlab{}.
\newblock \showarticletitle{P-Tuning: Prompt Tuning Can Be Comparable to Fine-tuning Across Scales and Tasks}. In \bibinfo{booktitle}{\emph{Proceedings of the 60th Annual Meeting of the Association for Computational Linguistics (Volume 2: Short Papers)}}. \bibinfo{pages}{61--68}.
\newblock


\bibitem[Liu et~al\mbox{.}(2023)]%
        {ref:p-tuning}
\bibfield{author}{\bibinfo{person}{Xiao Liu}, \bibinfo{person}{Yanan Zheng}, \bibinfo{person}{Zhengxiao Du}, \bibinfo{person}{Ming Ding}, \bibinfo{person}{Yujie Qian}, \bibinfo{person}{Zhilin Yang}, {and} \bibinfo{person}{Jie Tang}.} \bibinfo{year}{2023}\natexlab{}.
\newblock \bibinfo{title}{GPT Understands, Too}.
\newblock
\newblock
\showeprint[arxiv]{2103.10385}~[cs.CL]
\urldef\tempurl%
\url{https://arxiv.org/abs/2103.10385}
\showURL{%
\tempurl}


\bibitem[Liu et~al\mbox{.}(2019)]%
        {ref:roberta}
\bibfield{author}{\bibinfo{person}{Yinhan Liu}, \bibinfo{person}{Myle Ott}, \bibinfo{person}{Naman Goyal}, \bibinfo{person}{Jingfei Du}, \bibinfo{person}{Mandar Joshi}, \bibinfo{person}{Danqi Chen}, \bibinfo{person}{Omer Levy}, \bibinfo{person}{Mike Lewis}, \bibinfo{person}{Luke Zettlemoyer}, {and} \bibinfo{person}{Veselin Stoyanov}.} \bibinfo{year}{2019}\natexlab{}.
\newblock \bibinfo{title}{RoBERTa: A Robustly Optimized BERT Pretraining Approach}.
\newblock
\newblock
\showeprint[arxiv]{1907.11692}~[cs.CL]
\urldef\tempurl%
\url{https://arxiv.org/abs/1907.11692}
\showURL{%
\tempurl}


\bibitem[Lo and Lim(2023)]%
        {ref:ecpr}
\bibfield{author}{\bibinfo{person}{Pei{-}Chi Lo} {and} \bibinfo{person}{Ee{-}Peng Lim}.} \bibinfo{year}{2023}\natexlab{}.
\newblock \showarticletitle{Contextual Path Retrieval: {A} Contextual Entity Relation Embedding-based Approach}.
\newblock \bibinfo{journal}{\emph{{ACM} Trans. Inf. Syst.}} \bibinfo{volume}{41}, \bibinfo{number}{1} (\bibinfo{year}{2023}), \bibinfo{pages}{1:1--1:38}.
\newblock


\bibitem[Logeswaran et~al\mbox{.}(2019)]%
        {ref:zeshel}
\bibfield{author}{\bibinfo{person}{Lajanugen Logeswaran}, \bibinfo{person}{Ming{-}Wei Chang}, \bibinfo{person}{Kenton Lee}, \bibinfo{person}{Kristina Toutanova}, \bibinfo{person}{Jacob Devlin}, {and} \bibinfo{person}{Honglak Lee}.} \bibinfo{year}{2019}\natexlab{}.
\newblock \showarticletitle{Zero-Shot Entity Linking by Reading Entity Descriptions}. In \bibinfo{booktitle}{\emph{Proceedings of the 57th Conference of the Association for Computational Linguistics, {ACL} 2019, Florence, Italy, July 28- August 2, 2019, Volume 1: Long Papers}}. \bibinfo{pages}{3449--3460}.
\newblock


\bibitem[Luo et~al\mbox{.}(2023)]%
        {ref:rasr}
\bibfield{author}{\bibinfo{person}{Dan Luo}, \bibinfo{person}{Jiawei Sheng}, \bibinfo{person}{Hongbo Xu}, \bibinfo{person}{Lihong Wang}, {and} \bibinfo{person}{Bin Wang}.} \bibinfo{year}{2023}\natexlab{}.
\newblock \showarticletitle{Improving Complex Knowledge Base Question Answering with Relation-Aware Subgraph Retrieval and Reasoning Network}. In \bibinfo{booktitle}{\emph{International Joint Conference on Neural Networks, {IJCNN} 2023, Gold Coast, Australia, June 18-23, 2023}}. \bibinfo{pages}{1--8}.
\newblock


\bibitem[Luo et~al\mbox{.}(2024a)]%
        {ref:chatkbqa}
\bibfield{author}{\bibinfo{person}{Haoran Luo}, \bibinfo{person}{Haihong E}, \bibinfo{person}{Zichen Tang}, \bibinfo{person}{Shiyao Peng}, \bibinfo{person}{Yikai Guo}, \bibinfo{person}{Wentai Zhang}, \bibinfo{person}{Chenghao Ma}, \bibinfo{person}{Guanting Dong}, \bibinfo{person}{Meina Song}, \bibinfo{person}{Wei Lin}, \bibinfo{person}{Yifan Zhu}, {and} \bibinfo{person}{Luu~Anh Tuan}.} \bibinfo{year}{2024}\natexlab{a}.
\newblock \bibinfo{title}{ChatKBQA: A Generate-then-Retrieve Framework for Knowledge Base Question Answering with Fine-tuned Large Language Models}.
\newblock
\newblock
\showeprint[arxiv]{2310.08975}~[cs.CL]
\urldef\tempurl%
\url{https://arxiv.org/abs/2310.08975}
\showURL{%
\tempurl}


\bibitem[Luo et~al\mbox{.}(2024b)]%
        {ref:rog}
\bibfield{author}{\bibinfo{person}{Linhao Luo}, \bibinfo{person}{Yuan-Fang Li}, \bibinfo{person}{Gholamreza Haffari}, {and} \bibinfo{person}{Shirui Pan}.} \bibinfo{year}{2024}\natexlab{b}.
\newblock \bibinfo{title}{Reasoning on Graphs: Faithful and Interpretable Large Language Model Reasoning}.
\newblock
\newblock
\showeprint[arxiv]{2310.01061}~[cs.CL]
\urldef\tempurl%
\url{https://arxiv.org/abs/2310.01061}
\showURL{%
\tempurl}


\bibitem[Ma et~al\mbox{.}(2024)]%
        {ref:tog2}
\bibfield{author}{\bibinfo{person}{Shengjie Ma}, \bibinfo{person}{Chengjin Xu}, \bibinfo{person}{Xuhui Jiang}, \bibinfo{person}{Muzhi Li}, \bibinfo{person}{Huaren Qu}, {and} \bibinfo{person}{Jian Guo}.} \bibinfo{year}{2024}\natexlab{}.
\newblock \bibinfo{title}{Think-on-Graph 2.0: Deep and Interpretable Large Language Model Reasoning with Knowledge Graph-guided Retrieval}.
\newblock
\newblock
\showeprint[arxiv]{2407.10805}~[cs.CL]
\urldef\tempurl%
\url{https://arxiv.org/abs/2407.10805}
\showURL{%
\tempurl}


\bibitem[Ma et~al\mbox{.}(2023)]%
        {ref:rewrite1}
\bibfield{author}{\bibinfo{person}{Xinbei Ma}, \bibinfo{person}{Yeyun Gong}, \bibinfo{person}{Pengcheng He}, \bibinfo{person}{Hai Zhao}, {and} \bibinfo{person}{Nan Duan}.} \bibinfo{year}{2023}\natexlab{}.
\newblock \bibinfo{title}{Query Rewriting for Retrieval-Augmented Large Language Models}.
\newblock
\newblock
\showeprint[arxiv]{2305.14283}~[cs.CL]
\urldef\tempurl%
\url{https://arxiv.org/abs/2305.14283}
\showURL{%
\tempurl}


\bibitem[Mao et~al\mbox{.}(2024a)]%
        {ref:gfm}
\bibfield{author}{\bibinfo{person}{Haitao Mao}, \bibinfo{person}{Zhikai Chen}, \bibinfo{person}{Wenzhuo Tang}, \bibinfo{person}{Jianan Zhao}, \bibinfo{person}{Yao Ma}, \bibinfo{person}{Tong Zhao}, \bibinfo{person}{Neil Shah}, \bibinfo{person}{Mikhail Galkin}, {and} \bibinfo{person}{Jiliang Tang}.} \bibinfo{year}{2024}\natexlab{a}.
\newblock \showarticletitle{Position: Graph Foundation Models Are Already Here}. In \bibinfo{booktitle}{\emph{Forty-first International Conference on Machine Learning}}.
\newblock


\bibitem[Mao et~al\mbox{.}(2024c)]%
        {ref:gfmsurvey2}
\bibfield{author}{\bibinfo{person}{Qiheng Mao}, \bibinfo{person}{Zemin Liu}, \bibinfo{person}{Chenghao Liu}, \bibinfo{person}{Zhuo Li}, {and} \bibinfo{person}{Jianling Sun}.} \bibinfo{year}{2024}\natexlab{c}.
\newblock \bibinfo{title}{Advancing Graph Representation Learning with Large Language Models: A Comprehensive Survey of Techniques}.
\newblock
\newblock
\showeprint[arxiv]{2402.05952}~[cs.LG]
\urldef\tempurl%
\url{https://arxiv.org/abs/2402.05952}
\showURL{%
\tempurl}


\bibitem[Mao et~al\mbox{.}(2024b)]%
        {ref:rewrite3}
\bibfield{author}{\bibinfo{person}{Shengyu Mao}, \bibinfo{person}{Yong Jiang}, \bibinfo{person}{Boli Chen}, \bibinfo{person}{Xiao Li}, \bibinfo{person}{Peng Wang}, \bibinfo{person}{Xinyu Wang}, \bibinfo{person}{Pengjun Xie}, \bibinfo{person}{Fei Huang}, \bibinfo{person}{Huajun Chen}, {and} \bibinfo{person}{Ningyu Zhang}.} \bibinfo{year}{2024}\natexlab{b}.
\newblock \bibinfo{title}{RaFe: Ranking Feedback Improves Query Rewriting for RAG}.
\newblock
\newblock
\showeprint[arxiv]{2405.14431}~[cs.CL]
\urldef\tempurl%
\url{https://arxiv.org/abs/2405.14431}
\showURL{%
\tempurl}


\bibitem[Mavromatis and Karypis(2022)]%
        {ref:rearev}
\bibfield{author}{\bibinfo{person}{Costas Mavromatis} {and} \bibinfo{person}{George Karypis}.} \bibinfo{year}{2022}\natexlab{}.
\newblock \showarticletitle{ReaRev: Adaptive Reasoning for Question Answering over Knowledge Graphs}. In \bibinfo{booktitle}{\emph{Findings of the Association for Computational Linguistics: {EMNLP} 2022, Abu Dhabi, United Arab Emirates, December 7-11, 2022}}. \bibinfo{pages}{2447--2458}.
\newblock


\bibitem[Mavromatis and Karypis(2024)]%
        {ref:gnn-rag}
\bibfield{author}{\bibinfo{person}{Costas Mavromatis} {and} \bibinfo{person}{George Karypis}.} \bibinfo{year}{2024}\natexlab{}.
\newblock \bibinfo{title}{GNN-RAG: Graph Neural Retrieval for Large Language Model Reasoning}.
\newblock
\newblock
\showeprint[arxiv]{2405.20139}~[cs.CL]
\urldef\tempurl%
\url{https://arxiv.org/abs/2405.20139}
\showURL{%
\tempurl}


\bibitem[Mihaylov et~al\mbox{.}(2018)]%
        {ref:openbookqa}
\bibfield{author}{\bibinfo{person}{Todor Mihaylov}, \bibinfo{person}{Peter Clark}, \bibinfo{person}{Tushar Khot}, {and} \bibinfo{person}{Ashish Sabharwal}.} \bibinfo{year}{2018}\natexlab{}.
\newblock \showarticletitle{Can a Suit of Armor Conduct Electricity? {A} New Dataset for Open Book Question Answering}. In \bibinfo{booktitle}{\emph{Proceedings of the 2018 Conference on Empirical Methods in Natural Language Processing, Brussels, Belgium, October 31 - November 4, 2018}}. \bibinfo{pages}{2381--2391}.
\newblock


\bibitem[Miller et~al\mbox{.}(2016)]%
        {ref:wikimovie}
\bibfield{author}{\bibinfo{person}{Alexander~H. Miller}, \bibinfo{person}{Adam Fisch}, \bibinfo{person}{Jesse Dodge}, \bibinfo{person}{Amir{-}Hossein Karimi}, \bibinfo{person}{Antoine Bordes}, {and} \bibinfo{person}{Jason Weston}.} \bibinfo{year}{2016}\natexlab{}.
\newblock \showarticletitle{Key-Value Memory Networks for Directly Reading Documents}. In \bibinfo{booktitle}{\emph{Proceedings of the 2016 Conference on Empirical Methods in Natural Language Processing, {EMNLP} 2016, Austin, Texas, USA, November 1-4, 2016}}. \bibinfo{pages}{1400--1409}.
\newblock


\bibitem[Moon et~al\mbox{.}(2019)]%
        {ref:opendialkg}
\bibfield{author}{\bibinfo{person}{Seungwhan Moon}, \bibinfo{person}{Pararth Shah}, \bibinfo{person}{Anuj Kumar}, {and} \bibinfo{person}{Rajen Subba}.} \bibinfo{year}{2019}\natexlab{}.
\newblock \showarticletitle{OpenDialKG: Explainable Conversational Reasoning with Attention-based Walks over Knowledge Graphs}. In \bibinfo{booktitle}{\emph{Proceedings of the 57th Conference of the Association for Computational Linguistics, {ACL} 2019, Florence, Italy, July 28- August 2, 2019, Volume 1: Long Papers}}. \bibinfo{pages}{845--854}.
\newblock


\bibitem[Morris et~al\mbox{.}(2020)]%
        {ref:tudataset}
\bibfield{author}{\bibinfo{person}{Christopher Morris}, \bibinfo{person}{Nils~M. Kriege}, \bibinfo{person}{Franka Bause}, \bibinfo{person}{Kristian Kersting}, \bibinfo{person}{Petra Mutzel}, {and} \bibinfo{person}{Marion Neumann}.} \bibinfo{year}{2020}\natexlab{}.
\newblock \showarticletitle{TUDataset: A collection of benchmark datasets for learning with graphs}. In \bibinfo{booktitle}{\emph{ICML 2020 Workshop on Graph Representation Learning and Beyond (GRL+ 2020)}}.
\newblock


\bibitem[Munikoti et~al\mbox{.}(2023)]%
        {ref:atlantic}
\bibfield{author}{\bibinfo{person}{Sai Munikoti}, \bibinfo{person}{Anurag Acharya}, \bibinfo{person}{Sridevi Wagle}, {and} \bibinfo{person}{Sameera Horawalavithana}.} \bibinfo{year}{2023}\natexlab{}.
\newblock \bibinfo{title}{ATLANTIC: Structure-Aware Retrieval-Augmented Language Model for Interdisciplinary Science}.
\newblock
\newblock
\showeprint[arxiv]{2311.12289}~[cs.CL]
\urldef\tempurl%
\url{https://arxiv.org/abs/2311.12289}
\showURL{%
\tempurl}


\bibitem[Nie et~al\mbox{.}(2024)]%
        {ref:financialgpt}
\bibfield{author}{\bibinfo{person}{Yuqi Nie}, \bibinfo{person}{Yaxuan Kong}, \bibinfo{person}{Xiaowen Dong}, \bibinfo{person}{John~M. Mulvey}, \bibinfo{person}{H.~Vincent Poor}, \bibinfo{person}{Qingsong Wen}, {and} \bibinfo{person}{Stefan Zohren}.} \bibinfo{year}{2024}\natexlab{}.
\newblock \bibinfo{title}{A Survey of Large Language Models for Financial Applications: Progress, Prospects and Challenges}.
\newblock
\newblock
\showeprint[arxiv]{2406.11903}~[q-fin.GN]
\urldef\tempurl%
\url{https://arxiv.org/abs/2406.11903}
\showURL{%
\tempurl}


\bibitem[Onoe et~al\mbox{.}(2021)]%
        {ref:creak}
\bibfield{author}{\bibinfo{person}{Yasumasa Onoe}, \bibinfo{person}{Michael J.~Q. Zhang}, \bibinfo{person}{Eunsol Choi}, {and} \bibinfo{person}{Greg Durrett}.} \bibinfo{year}{2021}\natexlab{}.
\newblock \showarticletitle{{CREAK:} {A} Dataset for Commonsense Reasoning over Entity Knowledge}. In \bibinfo{booktitle}{\emph{Proceedings of the Neural Information Processing Systems Track on Datasets and Benchmarks 1, NeurIPS Datasets and Benchmarks 2021, December 2021, virtual}}.
\newblock


\bibitem[OpenAI(2024)]%
        {ref:gpt4}
\bibfield{author}{\bibinfo{person}{OpenAI}.} \bibinfo{year}{2024}\natexlab{}.
\newblock \bibinfo{title}{GPT-4 Technical Report}.
\newblock
\newblock
\showeprint[arxiv]{2303.08774}~[cs.CL]
\urldef\tempurl%
\url{https://arxiv.org/abs/2303.08774}
\showURL{%
\tempurl}


\bibitem[Ouyang et~al\mbox{.}(2022)]%
        {ref:chatgpt}
\bibfield{author}{\bibinfo{person}{Long Ouyang}, \bibinfo{person}{Jeffrey Wu}, \bibinfo{person}{Xu Jiang}, \bibinfo{person}{Diogo Almeida}, \bibinfo{person}{Carroll Wainwright}, \bibinfo{person}{Pamela Mishkin}, \bibinfo{person}{Chong Zhang}, \bibinfo{person}{Sandhini Agarwal}, \bibinfo{person}{Katarina Slama}, \bibinfo{person}{Alex Ray}, {et~al\mbox{.}}} \bibinfo{year}{2022}\natexlab{}.
\newblock \showarticletitle{Training language models to follow instructions with human feedback}.
\newblock \bibinfo{journal}{\emph{Advances in neural information processing systems}}  \bibinfo{volume}{35} (\bibinfo{year}{2022}), \bibinfo{pages}{27730--27744}.
\newblock


\bibitem[Pahuja et~al\mbox{.}(2023)]%
        {ref:kg-r3}
\bibfield{author}{\bibinfo{person}{Vardaan Pahuja}, \bibinfo{person}{Boshi Wang}, \bibinfo{person}{Hugo Latapie}, \bibinfo{person}{Jayanth Srinivasa}, {and} \bibinfo{person}{Yu Su}.} \bibinfo{year}{2023}\natexlab{}.
\newblock \showarticletitle{A Retrieve-and-Read Framework for Knowledge Graph Link Prediction}. In \bibinfo{booktitle}{\emph{Proceedings of the 32nd {ACM} International Conference on Information and Knowledge Management, {CIKM} 2023, Birmingham, United Kingdom, October 21-25, 2023}}. \bibinfo{pages}{1992--2002}.
\newblock


\bibitem[Pan et~al\mbox{.}(2023)]%
        {ref:gfmsurvey5}
\bibfield{author}{\bibinfo{person}{Jeff~Z. Pan}, \bibinfo{person}{Simon Razniewski}, \bibinfo{person}{Jan{-}Christoph Kalo}, \bibinfo{person}{Sneha Singhania}, \bibinfo{person}{Jiaoyan Chen}, \bibinfo{person}{Stefan Dietze}, \bibinfo{person}{Hajira Jabeen}, \bibinfo{person}{Janna Omeliyanenko}, \bibinfo{person}{Wen Zhang}, \bibinfo{person}{Matteo Lissandrini}, \bibinfo{person}{Russa Biswas}, \bibinfo{person}{Gerard de Melo}, \bibinfo{person}{Angela Bonifati}, \bibinfo{person}{Edlira Vakaj}, \bibinfo{person}{Mauro Dragoni}, {and} \bibinfo{person}{Damien Graux}.} \bibinfo{year}{2023}\natexlab{}.
\newblock \showarticletitle{Large Language Models and Knowledge Graphs: Opportunities and Challenges}.
\newblock \bibinfo{journal}{\emph{{TGDK}}} \bibinfo{volume}{1}, \bibinfo{number}{1} (\bibinfo{year}{2023}), \bibinfo{pages}{2:1--2:38}.
\newblock


\bibitem[Pan et~al\mbox{.}(2024)]%
        {ref:gfmsurvey6}
\bibfield{author}{\bibinfo{person}{Shirui Pan}, \bibinfo{person}{Linhao Luo}, \bibinfo{person}{Yufei Wang}, \bibinfo{person}{Chen Chen}, \bibinfo{person}{Jiapu Wang}, {and} \bibinfo{person}{Xindong Wu}.} \bibinfo{year}{2024}\natexlab{}.
\newblock \showarticletitle{Unifying Large Language Models and Knowledge Graphs: {A} Roadmap}.
\newblock \bibinfo{journal}{\emph{{IEEE} Trans. Knowl. Data Eng.}} \bibinfo{volume}{36}, \bibinfo{number}{7} (\bibinfo{year}{2024}), \bibinfo{pages}{3580--3599}.
\newblock


\bibitem[Peng et~al\mbox{.}(2024)]%
        {ref:rewrite2}
\bibfield{author}{\bibinfo{person}{Wenjun Peng}, \bibinfo{person}{Guiyang Li}, \bibinfo{person}{Yue Jiang}, \bibinfo{person}{Zilong Wang}, \bibinfo{person}{Dan Ou}, \bibinfo{person}{Xiaoyi Zeng}, \bibinfo{person}{Derong Xu}, \bibinfo{person}{Tong Xu}, {and} \bibinfo{person}{Enhong Chen}.} \bibinfo{year}{2024}\natexlab{}.
\newblock \showarticletitle{Large Language Model based Long-tail Query Rewriting in Taobao Search}. In \bibinfo{booktitle}{\emph{Companion Proceedings of the {ACM} on Web Conference 2024, {WWW} 2024, Singapore, Singapore, May 13-17, 2024}}. \bibinfo{pages}{20--28}.
\newblock


\bibitem[Peng and Yang(2024)]%
        {ref:ra-sim}
\bibfield{author}{\bibinfo{person}{Zhuoyi Peng} {and} \bibinfo{person}{Yi Yang}.} \bibinfo{year}{2024}\natexlab{}.
\newblock \bibinfo{title}{Connecting the Dots: Inferring Patent Phrase Similarity with Retrieved Phrase Graphs}.
\newblock
\newblock
\showeprint[arxiv]{2403.16265}~[cs.CL]
\urldef\tempurl%
\url{https://arxiv.org/abs/2403.16265}
\showURL{%
\tempurl}


\bibitem[Perevalov et~al\mbox{.}(2022)]%
        {ref:qald}
\bibfield{author}{\bibinfo{person}{Aleksandr Perevalov}, \bibinfo{person}{Dennis Diefenbach}, \bibinfo{person}{Ricardo Usbeck}, {and} \bibinfo{person}{Andreas Both}.} \bibinfo{year}{2022}\natexlab{}.
\newblock \showarticletitle{QALD-9-plus: {A} Multilingual Dataset for Question Answering over DBpedia and Wikidata Translated by Native Speakers}. In \bibinfo{booktitle}{\emph{16th {IEEE} International Conference on Semantic Computing, {ICSC} 2022, Laguna Hills, CA, USA, January 26-28, 2022}}. \bibinfo{pages}{229--234}.
\newblock


\bibitem[Petroni et~al\mbox{.}(2021)]%
        {ref:zsre}
\bibfield{author}{\bibinfo{person}{Fabio Petroni}, \bibinfo{person}{Aleksandra Piktus}, \bibinfo{person}{Angela Fan}, \bibinfo{person}{Patrick S.~H. Lewis}, \bibinfo{person}{Majid Yazdani}, \bibinfo{person}{Nicola~De Cao}, \bibinfo{person}{James Thorne}, \bibinfo{person}{Yacine Jernite}, \bibinfo{person}{Vladimir Karpukhin}, \bibinfo{person}{Jean Maillard}, \bibinfo{person}{Vassilis Plachouras}, \bibinfo{person}{Tim Rockt{\"{a}}schel}, {and} \bibinfo{person}{Sebastian Riedel}.} \bibinfo{year}{2021}\natexlab{}.
\newblock \showarticletitle{{KILT:} a Benchmark for Knowledge Intensive Language Tasks}. In \bibinfo{booktitle}{\emph{Proceedings of the 2021 Conference of the North American Chapter of the Association for Computational Linguistics: Human Language Technologies, {NAACL-HLT} 2021, Online, June 6-11, 2021}}. \bibinfo{pages}{2523--2544}.
\newblock


\bibitem[Qi et~al\mbox{.}(2023)]%
        {ref:foodgpt}
\bibfield{author}{\bibinfo{person}{Zhixiao Qi}, \bibinfo{person}{Yijiong Yu}, \bibinfo{person}{Meiqi Tu}, \bibinfo{person}{Junyi Tan}, {and} \bibinfo{person}{Yongfeng Huang}.} \bibinfo{year}{2023}\natexlab{}.
\newblock \bibinfo{title}{FoodGPT: A Large Language Model in Food Testing Domain with Incremental Pre-training and Knowledge Graph Prompt}.
\newblock
\newblock
\showeprint[arxiv]{2308.10173}~[cs.CL]
\urldef\tempurl%
\url{https://arxiv.org/abs/2308.10173}
\showURL{%
\tempurl}


\bibitem[Qiao et~al\mbox{.}(2024)]%
        {ref:rewrite4}
\bibfield{author}{\bibinfo{person}{Zile Qiao}, \bibinfo{person}{Wei Ye}, \bibinfo{person}{Yong Jiang}, \bibinfo{person}{Tong Mo}, \bibinfo{person}{Pengjun Xie}, \bibinfo{person}{Weiping Li}, \bibinfo{person}{Fei Huang}, {and} \bibinfo{person}{Shikun Zhang}.} \bibinfo{year}{2024}\natexlab{}.
\newblock \bibinfo{title}{Supportiveness-based Knowledge Rewriting for Retrieval-augmented Language Modeling}.
\newblock
\newblock
\showeprint[arxiv]{2406.08116}~[cs.CL]
\urldef\tempurl%
\url{https://arxiv.org/abs/2406.08116}
\showURL{%
\tempurl}


\bibitem[Raffel et~al\mbox{.}(2020)]%
        {ref:t5}
\bibfield{author}{\bibinfo{person}{Colin Raffel}, \bibinfo{person}{Noam Shazeer}, \bibinfo{person}{Adam Roberts}, \bibinfo{person}{Katherine Lee}, \bibinfo{person}{Sharan Narang}, \bibinfo{person}{Michael Matena}, \bibinfo{person}{Yanqi Zhou}, \bibinfo{person}{Wei Li}, {and} \bibinfo{person}{Peter~J. Liu}.} \bibinfo{year}{2020}\natexlab{}.
\newblock \showarticletitle{Exploring the Limits of Transfer Learning with a Unified Text-to-Text Transformer}.
\newblock \bibinfo{journal}{\emph{J. Mach. Learn. Res.}}  \bibinfo{volume}{21} (\bibinfo{year}{2020}), \bibinfo{pages}{140:1--140:67}.
\newblock


\bibitem[Ranade and Joshi(2023)]%
        {ref:fabula}
\bibfield{author}{\bibinfo{person}{Priyanka Ranade} {and} \bibinfo{person}{Anupam Joshi}.} \bibinfo{year}{2023}\natexlab{}.
\newblock \showarticletitle{{FABULA:} Intelligence Report Generation Using Retrieval-Augmented Narrative Construction}. In \bibinfo{booktitle}{\emph{Proceedings of the International Conference on Advances in Social Networks Analysis and Mining, {ASONAM} 2023, Kusadasi, Turkey, November 6-9, 2023}}. \bibinfo{pages}{603--610}.
\newblock


\bibitem[Reimers and Gurevych(2019)]%
        {ref:sentence-bert}
\bibfield{author}{\bibinfo{person}{Nils Reimers} {and} \bibinfo{person}{Iryna Gurevych}.} \bibinfo{year}{2019}\natexlab{}.
\newblock \showarticletitle{Sentence-BERT: Sentence Embeddings using Siamese BERT-Networks}. In \bibinfo{booktitle}{\emph{Proceedings of the 2019 Conference on Empirical Methods in Natural Language Processing and the 9th International Joint Conference on Natural Language Processing, {EMNLP-IJCNLP} 2019, Hong Kong, China, November 3-7, 2019}}. \bibinfo{pages}{3980--3990}.
\newblock


\bibitem[Rong et~al\mbox{.}(2020)]%
        {ref:graphml}
\bibfield{author}{\bibinfo{person}{Yu Rong}, \bibinfo{person}{Wenbing Huang}, \bibinfo{person}{Tingyang Xu}, {and} \bibinfo{person}{Junzhou Huang}.} \bibinfo{year}{2020}\natexlab{}.
\newblock \showarticletitle{DropEdge: Towards Deep Graph Convolutional Networks on Node Classification}. In \bibinfo{booktitle}{\emph{8th International Conference on Learning Representations, {ICLR} 2020, Addis Ababa, Ethiopia, April 26-30, 2020}}.
\newblock


\bibitem[Sap et~al\mbox{.}(2019a)]%
        {ref:atomic}
\bibfield{author}{\bibinfo{person}{Maarten Sap}, \bibinfo{person}{Ronan~Le Bras}, \bibinfo{person}{Emily Allaway}, \bibinfo{person}{Chandra Bhagavatula}, \bibinfo{person}{Nicholas Lourie}, \bibinfo{person}{Hannah Rashkin}, \bibinfo{person}{Brendan Roof}, \bibinfo{person}{Noah~A. Smith}, {and} \bibinfo{person}{Yejin Choi}.} \bibinfo{year}{2019}\natexlab{a}.
\newblock \showarticletitle{{ATOMIC:} An Atlas of Machine Commonsense for If-Then Reasoning}. In \bibinfo{booktitle}{\emph{The Thirty-Third {AAAI} Conference on Artificial Intelligence, {AAAI} 2019, The Thirty-First Innovative Applications of Artificial Intelligence Conference, {IAAI} 2019, The Ninth {AAAI} Symposium on Educational Advances in Artificial Intelligence, {EAAI} 2019, Honolulu, Hawaii, USA, January 27 - February 1, 2019}}. \bibinfo{pages}{3027--3035}.
\newblock


\bibitem[Sap et~al\mbox{.}(2019b)]%
        {ref:socialIQA}
\bibfield{author}{\bibinfo{person}{Maarten Sap}, \bibinfo{person}{Hannah Rashkin}, \bibinfo{person}{Derek Chen}, \bibinfo{person}{Ronan LeBras}, {and} \bibinfo{person}{Yejin Choi}.} \bibinfo{year}{2019}\natexlab{b}.
\newblock \bibinfo{title}{SocialIQA: Commonsense Reasoning about Social Interactions}.
\newblock
\newblock
\showeprint[arxiv]{1904.09728}~[cs.CL]
\urldef\tempurl%
\url{https://arxiv.org/abs/1904.09728}
\showURL{%
\tempurl}


\bibitem[Sarmah et~al\mbox{.}(2024)]%
        {ref:hybridrag}
\bibfield{author}{\bibinfo{person}{Bhaskarjit Sarmah}, \bibinfo{person}{Benika Hall}, \bibinfo{person}{Rohan Rao}, \bibinfo{person}{Sunil Patel}, \bibinfo{person}{Stefano Pasquali}, {and} \bibinfo{person}{Dhagash Mehta}.} \bibinfo{year}{2024}\natexlab{}.
\newblock \bibinfo{title}{HybridRAG: Integrating Knowledge Graphs and Vector Retrieval Augmented Generation for Efficient Information Extraction}.
\newblock
\newblock
\showeprint[arxiv]{2408.04948}~[cs.CL]
\urldef\tempurl%
\url{https://arxiv.org/abs/2408.04948}
\showURL{%
\tempurl}


\bibitem[Saxena et~al\mbox{.}(2020)]%
        {ref:embedkgqa}
\bibfield{author}{\bibinfo{person}{Apoorv Saxena}, \bibinfo{person}{Aditay Tripathi}, {and} \bibinfo{person}{Partha~P. Talukdar}.} \bibinfo{year}{2020}\natexlab{}.
\newblock \showarticletitle{Improving Multi-hop Question Answering over Knowledge Graphs using Knowledge Base Embeddings}. In \bibinfo{booktitle}{\emph{Proceedings of the 58th Annual Meeting of the Association for Computational Linguistics, {ACL} 2020, Online, July 5-10, 2020}}. \bibinfo{pages}{4498--4507}.
\newblock


\bibitem[Sen et~al\mbox{.}(2022)]%
        {ref:mintaka}
\bibfield{author}{\bibinfo{person}{Priyanka Sen}, \bibinfo{person}{Alham~Fikri Aji}, {and} \bibinfo{person}{Amir Saffari}.} \bibinfo{year}{2022}\natexlab{}.
\newblock \showarticletitle{Mintaka: {A} Complex, Natural, and Multilingual Dataset for End-to-End Question Answering}. In \bibinfo{booktitle}{\emph{Proceedings of the 29th International Conference on Computational Linguistics, {COLING} 2022, Gyeongju, Republic of Korea, October 12-17, 2022}}. \bibinfo{pages}{1604--1619}.
\newblock


\bibitem[Shehzad et~al\mbox{.}(2024)]%
        {ref:graphtransformer}
\bibfield{author}{\bibinfo{person}{Ahsan Shehzad}, \bibinfo{person}{Feng Xia}, \bibinfo{person}{Shagufta Abid}, \bibinfo{person}{Ciyuan Peng}, \bibinfo{person}{Shuo Yu}, \bibinfo{person}{Dongyu Zhang}, {and} \bibinfo{person}{Karin Verspoor}.} \bibinfo{year}{2024}\natexlab{}.
\newblock \bibinfo{title}{Graph Transformers: A Survey}.
\newblock
\newblock
\showeprint[arxiv]{2407.09777}~[cs.LG]
\urldef\tempurl%
\url{https://arxiv.org/abs/2407.09777}
\showURL{%
\tempurl}


\bibitem[Shu et~al\mbox{.}(2022)]%
        {ref:tiara}
\bibfield{author}{\bibinfo{person}{Yiheng Shu}, \bibinfo{person}{Zhiwei Yu}, \bibinfo{person}{Yuhan Li}, \bibinfo{person}{Börje~F. Karlsson}, \bibinfo{person}{Tingting Ma}, \bibinfo{person}{Yuzhong Qu}, {and} \bibinfo{person}{Chin-Yew Lin}.} \bibinfo{year}{2022}\natexlab{}.
\newblock \bibinfo{title}{TIARA: Multi-grained Retrieval for Robust Question Answering over Large Knowledge Bases}.
\newblock
\newblock
\showeprint[arxiv]{2210.12925}~[cs.CL]
\urldef\tempurl%
\url{https://arxiv.org/abs/2210.12925}
\showURL{%
\tempurl}


\bibitem[Srivastava et~al\mbox{.}(2020)]%
        {ref:iot}
\bibfield{author}{\bibinfo{person}{Saurabh Srivastava}, \bibinfo{person}{Milind~D Jain}, \bibinfo{person}{Harshita Jain}, \bibinfo{person}{Kritik Jaroli}, \bibinfo{person}{VJ~Mayank Patel}, {and} \bibinfo{person}{L Khan}.} \bibinfo{year}{2020}\natexlab{}.
\newblock \showarticletitle{IOT monitoring bin for smart cities}. In \bibinfo{booktitle}{\emph{3rd Smart Cities Symposium (SCS 2020)}}, Vol.~\bibinfo{volume}{2020}. IET, \bibinfo{pages}{533--536}.
\newblock


\bibitem[Suchanek et~al\mbox{.}(2007)]%
        {ref:yago}
\bibfield{author}{\bibinfo{person}{Fabian~M Suchanek}, \bibinfo{person}{Gjergji Kasneci}, {and} \bibinfo{person}{Gerhard Weikum}.} \bibinfo{year}{2007}\natexlab{}.
\newblock \showarticletitle{Yago: a core of semantic knowledge}. In \bibinfo{booktitle}{\emph{Proceedings of the 16th international conference on World Wide Web}}. \bibinfo{pages}{697--706}.
\newblock


\bibitem[Sun et~al\mbox{.}(2019)]%
        {ref:pullnet}
\bibfield{author}{\bibinfo{person}{Haitian Sun}, \bibinfo{person}{Tania Bedrax{-}Weiss}, {and} \bibinfo{person}{William~W. Cohen}.} \bibinfo{year}{2019}\natexlab{}.
\newblock \showarticletitle{PullNet: Open Domain Question Answering with Iterative Retrieval on Knowledge Bases and Text}. In \bibinfo{booktitle}{\emph{Proceedings of the 2019 Conference on Empirical Methods in Natural Language Processing and the 9th International Joint Conference on Natural Language Processing, {EMNLP-IJCNLP} 2019, Hong Kong, China, November 3-7, 2019}}. \bibinfo{pages}{2380--2390}.
\newblock


\bibitem[Sun et~al\mbox{.}(2018)]%
        {ref:graft}
\bibfield{author}{\bibinfo{person}{Haitian Sun}, \bibinfo{person}{Bhuwan Dhingra}, \bibinfo{person}{Manzil Zaheer}, \bibinfo{person}{Kathryn Mazaitis}, \bibinfo{person}{Ruslan Salakhutdinov}, {and} \bibinfo{person}{William~W. Cohen}.} \bibinfo{year}{2018}\natexlab{}.
\newblock \showarticletitle{Open Domain Question Answering Using Early Fusion of Knowledge Bases and Text}. In \bibinfo{booktitle}{\emph{Proceedings of the 2018 Conference on Empirical Methods in Natural Language Processing, Brussels, Belgium, October 31 - November 4, 2018}}. \bibinfo{pages}{4231--4242}.
\newblock


\bibitem[Sun et~al\mbox{.}(2023)]%
        {ref:hsge}
\bibfield{author}{\bibinfo{person}{Hao Sun}, \bibinfo{person}{Yang Li}, \bibinfo{person}{Liwei Deng}, \bibinfo{person}{Bowen Li}, \bibinfo{person}{Binyuan Hui}, \bibinfo{person}{Binhua Li}, \bibinfo{person}{Yunshi Lan}, \bibinfo{person}{Yan Zhang}, {and} \bibinfo{person}{Yongbin Li}.} \bibinfo{year}{2023}\natexlab{}.
\newblock \showarticletitle{History Semantic Graph Enhanced Conversational {KBQA} with Temporal Information Modeling}. In \bibinfo{booktitle}{\emph{Proceedings of the 61st Annual Meeting of the Association for Computational Linguistics (Volume 1: Long Papers), {ACL} 2023, Toronto, Canada, July 9-14, 2023}}. \bibinfo{pages}{3521--3533}.
\newblock


\bibitem[Sun et~al\mbox{.}(2024b)]%
        {ref:tog}
\bibfield{author}{\bibinfo{person}{Jiashuo Sun}, \bibinfo{person}{Chengjin Xu}, \bibinfo{person}{Lumingyuan Tang}, \bibinfo{person}{Saizhuo Wang}, \bibinfo{person}{Chen Lin}, \bibinfo{person}{Yeyun Gong}, \bibinfo{person}{Lionel~M. Ni}, \bibinfo{person}{Heung-Yeung Shum}, {and} \bibinfo{person}{Jian Guo}.} \bibinfo{year}{2024}\natexlab{b}.
\newblock \bibinfo{title}{Think-on-Graph: Deep and Responsible Reasoning of Large Language Model on Knowledge Graph}.
\newblock
\newblock
\showeprint[arxiv]{2307.07697}~[cs.CL]
\urldef\tempurl%
\url{https://arxiv.org/abs/2307.07697}
\showURL{%
\tempurl}


\bibitem[Sun et~al\mbox{.}(2024a)]%
        {ref:oda}
\bibfield{author}{\bibinfo{person}{Lei Sun}, \bibinfo{person}{Zhengwei Tao}, \bibinfo{person}{Youdi Li}, {and} \bibinfo{person}{Hiroshi Arakawa}.} \bibinfo{year}{2024}\natexlab{a}.
\newblock \bibinfo{title}{ODA: Observation-Driven Agent for integrating LLMs and Knowledge Graphs}.
\newblock
\newblock
\showeprint[arxiv]{2404.07677}~[cs.CL]
\urldef\tempurl%
\url{https://arxiv.org/abs/2404.07677}
\showURL{%
\tempurl}


\bibitem[Talmor and Berant(2018)]%
        {ref:cwq}
\bibfield{author}{\bibinfo{person}{Alon Talmor} {and} \bibinfo{person}{Jonathan Berant}.} \bibinfo{year}{2018}\natexlab{}.
\newblock \showarticletitle{The Web as a Knowledge-Base for Answering Complex Questions}. In \bibinfo{booktitle}{\emph{Proceedings of the 2018 Conference of the North American Chapter of the Association for Computational Linguistics: Human Language Technologies, {NAACL-HLT} 2018, New Orleans, Louisiana, USA, June 1-6, 2018, Volume 1 (Long Papers)}}. \bibinfo{pages}{641--651}.
\newblock


\bibitem[Talmor et~al\mbox{.}(2019)]%
        {ref:commonsenseqa}
\bibfield{author}{\bibinfo{person}{Alon Talmor}, \bibinfo{person}{Jonathan Herzig}, \bibinfo{person}{Nicholas Lourie}, {and} \bibinfo{person}{Jonathan Berant}.} \bibinfo{year}{2019}\natexlab{}.
\newblock \showarticletitle{CommonsenseQA: {A} Question Answering Challenge Targeting Commonsense Knowledge}. In \bibinfo{booktitle}{\emph{Proceedings of the 2019 Conference of the North American Chapter of the Association for Computational Linguistics: Human Language Technologies, {NAACL-HLT} 2019, Minneapolis, MN, USA, June 2-7, 2019, Volume 1 (Long and Short Papers)}}. \bibinfo{pages}{4149--4158}.
\newblock


\bibitem[Taunk et~al\mbox{.}(2023)]%
        {ref:grapeqa}
\bibfield{author}{\bibinfo{person}{Dhaval Taunk}, \bibinfo{person}{Lakshya Khanna}, \bibinfo{person}{Siri Venkata Pavan~Kumar Kandru}, \bibinfo{person}{Vasudeva Varma}, \bibinfo{person}{Charu Sharma}, {and} \bibinfo{person}{Makarand Tapaswi}.} \bibinfo{year}{2023}\natexlab{}.
\newblock \showarticletitle{GrapeQA: GRaph Augmentation and Pruning to Enhance Question-Answering}. In \bibinfo{booktitle}{\emph{Companion Proceedings of the {ACM} Web Conference 2023, {WWW} 2023, Austin, TX, USA, 30 April 2023 - 4 May 2023}}. \bibinfo{pages}{1138--1144}.
\newblock


\bibitem[Toutanova et~al\mbox{.}(2015)]%
        {ref:fb15k-237}
\bibfield{author}{\bibinfo{person}{Kristina Toutanova}, \bibinfo{person}{Danqi Chen}, \bibinfo{person}{Patrick Pantel}, \bibinfo{person}{Hoifung Poon}, \bibinfo{person}{Pallavi Choudhury}, {and} \bibinfo{person}{Michael Gamon}.} \bibinfo{year}{2015}\natexlab{}.
\newblock \showarticletitle{Representing Text for Joint Embedding of Text and Knowledge Bases}. In \bibinfo{booktitle}{\emph{Proceedings of the 2015 Conference on Empirical Methods in Natural Language Processing, {EMNLP} 2015, Lisbon, Portugal, September 17-21, 2015}}. \bibinfo{pages}{1499--1509}.
\newblock


\bibitem[Touvron et~al\mbox{.}(2023)]%
        {ref:llama}
\bibfield{author}{\bibinfo{person}{Hugo Touvron}, \bibinfo{person}{Louis Martin}, {and} \bibinfo{person}{et al}.} \bibinfo{year}{2023}\natexlab{}.
\newblock \bibinfo{title}{Llama 2: Open Foundation and Fine-Tuned Chat Models}.
\newblock
\newblock
\showeprint[arxiv]{2307.09288}~[cs.CL]
\urldef\tempurl%
\url{https://arxiv.org/abs/2307.09288}
\showURL{%
\tempurl}


\bibitem[Vaswani et~al\mbox{.}(2017)]%
        {ref:transformer}
\bibfield{author}{\bibinfo{person}{Ashish Vaswani}, \bibinfo{person}{Noam Shazeer}, \bibinfo{person}{Niki Parmar}, \bibinfo{person}{Jakob Uszkoreit}, \bibinfo{person}{Llion Jones}, \bibinfo{person}{Aidan~N. Gomez}, \bibinfo{person}{Lukasz Kaiser}, {and} \bibinfo{person}{Illia Polosukhin}.} \bibinfo{year}{2017}\natexlab{}.
\newblock \showarticletitle{Attention is All you Need}. In \bibinfo{booktitle}{\emph{Advances in Neural Information Processing Systems 30: Annual Conference on Neural Information Processing Systems 2017, December 4-9, 2017, Long Beach, CA, {USA}}}. \bibinfo{pages}{5998--6008}.
\newblock


\bibitem[Veličković et~al\mbox{.}(2018)]%
        {ref:gat}
\bibfield{author}{\bibinfo{person}{Petar Veličković}, \bibinfo{person}{Guillem Cucurull}, \bibinfo{person}{Arantxa Casanova}, \bibinfo{person}{Adriana Romero}, \bibinfo{person}{Pietro Liò}, {and} \bibinfo{person}{Yoshua Bengio}.} \bibinfo{year}{2018}\natexlab{}.
\newblock \bibinfo{title}{Graph Attention Networks}.
\newblock
\newblock
\showeprint[arxiv]{1710.10903}~[stat.ML]
\urldef\tempurl%
\url{https://arxiv.org/abs/1710.10903}
\showURL{%
\tempurl}


\bibitem[Vrande{\v{c}}i{\'c} and Kr{\"o}tzsch(2014)]%
        {ref:wikidata}
\bibfield{author}{\bibinfo{person}{Denny Vrande{\v{c}}i{\'c}} {and} \bibinfo{person}{Markus Kr{\"o}tzsch}.} \bibinfo{year}{2014}\natexlab{}.
\newblock \showarticletitle{Wikidata: a free collaborative knowledgebase}.
\newblock \bibinfo{journal}{\emph{Commun. ACM}} \bibinfo{volume}{57}, \bibinfo{number}{10} (\bibinfo{year}{2014}), \bibinfo{pages}{78--85}.
\newblock


\bibitem[Wang et~al\mbox{.}(2023c)]%
        {ref:keqing}
\bibfield{author}{\bibinfo{person}{Chaojie Wang}, \bibinfo{person}{Yishi Xu}, \bibinfo{person}{Zhong Peng}, \bibinfo{person}{Chenxi Zhang}, \bibinfo{person}{Bo Chen}, \bibinfo{person}{Xinrun Wang}, \bibinfo{person}{Lei Feng}, {and} \bibinfo{person}{Bo An}.} \bibinfo{year}{2023}\natexlab{c}.
\newblock \bibinfo{title}{keqing: knowledge-based question answering is a nature chain-of-thought mentor of LLM}.
\newblock
\newblock
\showeprint[arxiv]{2401.00426}~[cs.CL]
\urldef\tempurl%
\url{https://arxiv.org/abs/2401.00426}
\showURL{%
\tempurl}


\bibitem[Wang et~al\mbox{.}(2023b)]%
        {ref:cansolve}
\bibfield{author}{\bibinfo{person}{Heng Wang}, \bibinfo{person}{Shangbin Feng}, \bibinfo{person}{Tianxing He}, \bibinfo{person}{Zhaoxuan Tan}, \bibinfo{person}{Xiaochuang Han}, {and} \bibinfo{person}{Yulia Tsvetkov}.} \bibinfo{year}{2023}\natexlab{b}.
\newblock \showarticletitle{Can Language Models Solve Graph Problems in Natural Language?}. In \bibinfo{booktitle}{\emph{Advances in Neural Information Processing Systems 36: Annual Conference on Neural Information Processing Systems 2023, NeurIPS 2023, New Orleans, LA, USA, December 10 - 16, 2023}}.
\newblock


\bibitem[Wang et~al\mbox{.}(2024c)]%
        {ref:med1}
\bibfield{author}{\bibinfo{person}{Jinqiang Wang}, \bibinfo{person}{Huansheng Ning}, \bibinfo{person}{Yi Peng}, \bibinfo{person}{Qikai Wei}, \bibinfo{person}{Daniel Tesfai}, \bibinfo{person}{Wenwei Mao}, \bibinfo{person}{Tao Zhu}, {and} \bibinfo{person}{Runhe Huang}.} \bibinfo{year}{2024}\natexlab{c}.
\newblock \bibinfo{title}{A Survey on Large Language Models from General Purpose to Medical Applications: Datasets, Methodologies, and Evaluations}.
\newblock
\newblock
\showeprint[arxiv]{2406.10303}~[cs.CL]
\urldef\tempurl%
\url{https://arxiv.org/abs/2406.10303}
\showURL{%
\tempurl}


\bibitem[Wang et~al\mbox{.}(2023a)]%
        {ref:kd-cot}
\bibfield{author}{\bibinfo{person}{Keheng Wang}, \bibinfo{person}{Feiyu Duan}, \bibinfo{person}{Sirui Wang}, \bibinfo{person}{Peiguang Li}, \bibinfo{person}{Yunsen Xian}, \bibinfo{person}{Chuantao Yin}, \bibinfo{person}{Wenge Rong}, {and} \bibinfo{person}{Zhang Xiong}.} \bibinfo{year}{2023}\natexlab{a}.
\newblock \bibinfo{title}{Knowledge-Driven CoT: Exploring Faithful Reasoning in LLMs for Knowledge-intensive Question Answering}.
\newblock
\newblock
\showeprint[arxiv]{2308.13259}~[cs.CL]
\urldef\tempurl%
\url{https://arxiv.org/abs/2308.13259}
\showURL{%
\tempurl}


\bibitem[Wang et~al\mbox{.}(2022)]%
        {ref:rete}
\bibfield{author}{\bibinfo{person}{Ruijie Wang}, \bibinfo{person}{Zheng Li}, \bibinfo{person}{Danqing Zhang}, \bibinfo{person}{Qingyu Yin}, \bibinfo{person}{Tong Zhao}, \bibinfo{person}{Bing Yin}, {and} \bibinfo{person}{Tarek~F. Abdelzaher}.} \bibinfo{year}{2022}\natexlab{}.
\newblock \showarticletitle{{RETE:} Retrieval-Enhanced Temporal Event Forecasting on Unified Query Product Evolutionary Graph}. In \bibinfo{booktitle}{\emph{{WWW} '22: The {ACM} Web Conference 2022, Virtual Event, Lyon, France, April 25 - 29, 2022}}. \bibinfo{pages}{462--472}.
\newblock


\bibitem[Wang et~al\mbox{.}(2024d)]%
        {ref:edu1}
\bibfield{author}{\bibinfo{person}{Shen Wang}, \bibinfo{person}{Tianlong Xu}, \bibinfo{person}{Hang Li}, \bibinfo{person}{Chaoli Zhang}, \bibinfo{person}{Joleen Liang}, \bibinfo{person}{Jiliang Tang}, \bibinfo{person}{Philip~S. Yu}, {and} \bibinfo{person}{Qingsong Wen}.} \bibinfo{year}{2024}\natexlab{d}.
\newblock \bibinfo{title}{Large Language Models for Education: A Survey and Outlook}.
\newblock
\newblock
\showeprint[arxiv]{2403.18105}~[cs.CL]
\urldef\tempurl%
\url{https://arxiv.org/abs/2403.18105}
\showURL{%
\tempurl}


\bibitem[Wang et~al\mbox{.}(2023d)]%
        {ref:knowledgpt}
\bibfield{author}{\bibinfo{person}{Xintao Wang}, \bibinfo{person}{Qianwen Yang}, \bibinfo{person}{Yongting Qiu}, \bibinfo{person}{Jiaqing Liang}, \bibinfo{person}{Qianyu He}, \bibinfo{person}{Zhouhong Gu}, \bibinfo{person}{Yanghua Xiao}, {and} \bibinfo{person}{Wei Wang}.} \bibinfo{year}{2023}\natexlab{d}.
\newblock \bibinfo{title}{KnowledGPT: Enhancing Large Language Models with Retrieval and Storage Access on Knowledge Bases}.
\newblock
\newblock
\showeprint[arxiv]{2308.11761}~[cs.CL]
\urldef\tempurl%
\url{https://arxiv.org/abs/2308.11761}
\showURL{%
\tempurl}


\bibitem[Wang et~al\mbox{.}(2024a)]%
        {ref:rok}
\bibfield{author}{\bibinfo{person}{Yuqi Wang}, \bibinfo{person}{Boran Jiang}, \bibinfo{person}{Yi Luo}, \bibinfo{person}{Dawei He}, \bibinfo{person}{Peng Cheng}, {and} \bibinfo{person}{Liangcai Gao}.} \bibinfo{year}{2024}\natexlab{a}.
\newblock \bibinfo{title}{Reasoning on Efficient Knowledge Paths:Knowledge Graph Guides Large Language Model for Domain Question Answering}.
\newblock
\newblock
\showeprint[arxiv]{2404.10384}~[cs.CL]
\urldef\tempurl%
\url{https://arxiv.org/abs/2404.10384}
\showURL{%
\tempurl}


\bibitem[Wang et~al\mbox{.}(2024b)]%
        {ref:kgp}
\bibfield{author}{\bibinfo{person}{Yu Wang}, \bibinfo{person}{Nedim Lipka}, \bibinfo{person}{Ryan~A. Rossi}, \bibinfo{person}{Alexa~F. Siu}, \bibinfo{person}{Ruiyi Zhang}, {and} \bibinfo{person}{Tyler Derr}.} \bibinfo{year}{2024}\natexlab{b}.
\newblock \showarticletitle{Knowledge Graph Prompting for Multi-Document Question Answering}. In \bibinfo{booktitle}{\emph{Thirty-Eighth {AAAI} Conference on Artificial Intelligence, {AAAI} 2024, Thirty-Sixth Conference on Innovative Applications of Artificial Intelligence, {IAAI} 2024, Fourteenth Symposium on Educational Advances in Artificial Intelligence, {EAAI} 2014, February 20-27, 2024, Vancouver, Canada}}. \bibinfo{pages}{19206--19214}.
\newblock


\bibitem[Wang et~al\mbox{.}(2024e)]%
        {ref:graphbridge}
\bibfield{author}{\bibinfo{person}{Yaoke Wang}, \bibinfo{person}{Yun Zhu}, \bibinfo{person}{Wenqiao Zhang}, \bibinfo{person}{Yueting Zhuang}, \bibinfo{person}{Yunfei Li}, {and} \bibinfo{person}{Siliang Tang}.} \bibinfo{year}{2024}\natexlab{e}.
\newblock \bibinfo{title}{Bridging Local Details and Global Context in Text-Attributed Graphs}.
\newblock
\newblock
\showeprint[arxiv]{2406.12608}~[cs.CL]
\urldef\tempurl%
\url{https://arxiv.org/abs/2406.12608}
\showURL{%
\tempurl}


\bibitem[Wei et~al\mbox{.}(2019)]%
        {ref:mmgcn}
\bibfield{author}{\bibinfo{person}{Yinwei Wei}, \bibinfo{person}{Xiang Wang}, \bibinfo{person}{Liqiang Nie}, \bibinfo{person}{Xiangnan He}, \bibinfo{person}{Richang Hong}, {and} \bibinfo{person}{Tat-Seng Chua}.} \bibinfo{year}{2019}\natexlab{}.
\newblock \showarticletitle{MMGCN: Multi-modal graph convolution network for personalized recommendation of micro-video}. In \bibinfo{booktitle}{\emph{Proceedings of the 27th ACM international conference on multimedia}}. \bibinfo{pages}{1437--1445}.
\newblock


\bibitem[Wen et~al\mbox{.}(2024)]%
        {ref:mindmap}
\bibfield{author}{\bibinfo{person}{Yilin Wen}, \bibinfo{person}{Zifeng Wang}, {and} \bibinfo{person}{Jimeng Sun}.} \bibinfo{year}{2024}\natexlab{}.
\newblock \bibinfo{title}{MindMap: Knowledge Graph Prompting Sparks Graph of Thoughts in Large Language Models}.
\newblock
\newblock
\showeprint[arxiv]{2308.09729}~[cs.AI]
\urldef\tempurl%
\url{https://arxiv.org/abs/2308.09729}
\showURL{%
\tempurl}


\bibitem[Wold et~al\mbox{.}(2023)]%
        {ref:noname7}
\bibfield{author}{\bibinfo{person}{Sondre Wold}, \bibinfo{person}{Lilja Øvrelid}, {and} \bibinfo{person}{Erik Velldal}.} \bibinfo{year}{2023}\natexlab{}.
\newblock \bibinfo{title}{Text-To-KG Alignment: Comparing Current Methods on Classification Tasks}.
\newblock
\newblock
\showeprint[arxiv]{2306.02871}~[cs.CL]
\urldef\tempurl%
\url{https://arxiv.org/abs/2306.02871}
\showURL{%
\tempurl}


\bibitem[Wu et~al\mbox{.}(2024d)]%
        {ref:medgraphrag}
\bibfield{author}{\bibinfo{person}{Junde Wu}, \bibinfo{person}{Jiayuan Zhu}, {and} \bibinfo{person}{Yunli Qi}.} \bibinfo{year}{2024}\natexlab{d}.
\newblock \bibinfo{title}{Medical Graph RAG: Towards Safe Medical Large Language Model via Graph Retrieval-Augmented Generation}.
\newblock
\newblock
\showeprint[arxiv]{2408.04187}~[cs.CV]
\urldef\tempurl%
\url{https://arxiv.org/abs/2408.04187}
\showURL{%
\tempurl}


\bibitem[Wu et~al\mbox{.}(2024b)]%
        {ref:ragsurvey7}
\bibfield{author}{\bibinfo{person}{Shangyu Wu}, \bibinfo{person}{Ying Xiong}, \bibinfo{person}{Yufei Cui}, \bibinfo{person}{Haolun Wu}, \bibinfo{person}{Can Chen}, \bibinfo{person}{Ye Yuan}, \bibinfo{person}{Lianming Huang}, \bibinfo{person}{Xue Liu}, \bibinfo{person}{Tei-Wei Kuo}, \bibinfo{person}{Nan Guan}, {and} \bibinfo{person}{Chun~Jason Xue}.} \bibinfo{year}{2024}\natexlab{b}.
\newblock \bibinfo{title}{Retrieval-Augmented Generation for Natural Language Processing: A Survey}.
\newblock
\newblock
\showeprint[arxiv]{2407.13193}~[cs.CL]
\urldef\tempurl%
\url{https://arxiv.org/abs/2407.13193}
\showURL{%
\tempurl}


\bibitem[Wu et~al\mbox{.}(2024c)]%
        {ref:stark}
\bibfield{author}{\bibinfo{person}{Shirley Wu}, \bibinfo{person}{Shiyu Zhao}, \bibinfo{person}{Michihiro Yasunaga}, \bibinfo{person}{Kexin Huang}, \bibinfo{person}{Kaidi Cao}, \bibinfo{person}{Qian Huang}, \bibinfo{person}{Vassilis~N. Ioannidis}, \bibinfo{person}{Karthik Subbian}, \bibinfo{person}{James Zou}, {and} \bibinfo{person}{Jure Leskovec}.} \bibinfo{year}{2024}\natexlab{c}.
\newblock \bibinfo{title}{STaRK: Benchmarking LLM Retrieval on Textual and Relational Knowledge Bases}.
\newblock
\newblock
\showeprint[arxiv]{2404.13207}~[cs.IR]
\urldef\tempurl%
\url{https://arxiv.org/abs/2404.13207}
\showURL{%
\tempurl}


\bibitem[Wu et~al\mbox{.}(2023a)]%
        {ref:ger}
\bibfield{author}{\bibinfo{person}{Taiqiang Wu}, \bibinfo{person}{Xingyu Bai}, \bibinfo{person}{Weigang Guo}, \bibinfo{person}{Weijie Liu}, \bibinfo{person}{Siheng Li}, {and} \bibinfo{person}{Yujiu Yang}.} \bibinfo{year}{2023}\natexlab{a}.
\newblock \showarticletitle{Modeling Fine-grained Information via Knowledge-aware Hierarchical Graph for Zero-shot Entity Retrieval}. In \bibinfo{booktitle}{\emph{Proceedings of the Sixteenth {ACM} International Conference on Web Search and Data Mining, {WSDM} 2023, Singapore, 27 February 2023 - 3 March 2023}}. \bibinfo{pages}{1021--1029}.
\newblock


\bibitem[Wu et~al\mbox{.}(2024a)]%
        {ref:rag4dyg}
\bibfield{author}{\bibinfo{person}{Yuxia Wu}, \bibinfo{person}{Yuan Fang}, {and} \bibinfo{person}{Lizi Liao}.} \bibinfo{year}{2024}\natexlab{a}.
\newblock \bibinfo{title}{Retrieval Augmented Generation for Dynamic Graph Modeling}.
\newblock
\newblock
\showeprint[arxiv]{2408.14523}~[cs.LG]
\urldef\tempurl%
\url{https://arxiv.org/abs/2408.14523}
\showURL{%
\tempurl}


\bibitem[Wu et~al\mbox{.}(2023b)]%
        {ref:rra}
\bibfield{author}{\bibinfo{person}{Yike Wu}, \bibinfo{person}{Nan Hu}, \bibinfo{person}{Sheng Bi}, \bibinfo{person}{Guilin Qi}, \bibinfo{person}{Jie Ren}, \bibinfo{person}{Anhuan Xie}, {and} \bibinfo{person}{Wei Song}.} \bibinfo{year}{2023}\natexlab{b}.
\newblock \bibinfo{title}{Retrieve-Rewrite-Answer: A KG-to-Text Enhanced LLMs Framework for Knowledge Graph Question Answering}.
\newblock
\newblock
\showeprint[arxiv]{2309.11206}~[cs.CL]
\urldef\tempurl%
\url{https://arxiv.org/abs/2309.11206}
\showURL{%
\tempurl}


\bibitem[Xu et~al\mbox{.}(2024)]%
        {ref:noname1}
\bibfield{author}{\bibinfo{person}{Zhentao Xu}, \bibinfo{person}{Mark~Jerome Cruz}, \bibinfo{person}{Matthew Guevara}, \bibinfo{person}{Tie Wang}, \bibinfo{person}{Manasi Deshpande}, \bibinfo{person}{Xiaofeng Wang}, {and} \bibinfo{person}{Zheng Li}.} \bibinfo{year}{2024}\natexlab{}.
\newblock \showarticletitle{Retrieval-Augmented Generation with Knowledge Graphs for Customer Service Question Answering}. In \bibinfo{booktitle}{\emph{Proceedings of the 47th International {ACM} {SIGIR} Conference on Research and Development in Information Retrieval, {SIGIR} 2024, Washington DC, USA, July 14-18, 2024}}. \bibinfo{pages}{2905--2909}.
\newblock


\bibitem[Yang et~al\mbox{.}(2024c)]%
        {ref:qwen2}
\bibfield{author}{\bibinfo{person}{An Yang}, \bibinfo{person}{Baosong Yang}, {and} \bibinfo{person}{et al}.} \bibinfo{year}{2024}\natexlab{c}.
\newblock \bibinfo{title}{Qwen2 Technical Report}.
\newblock
\newblock
\showeprint[arxiv]{2407.10671}~[cs.CL]
\urldef\tempurl%
\url{https://arxiv.org/abs/2407.10671}
\showURL{%
\tempurl}


\bibitem[Yang et~al\mbox{.}(2024a)]%
        {ref:kg-rank}
\bibfield{author}{\bibinfo{person}{Rui Yang}, \bibinfo{person}{Haoran Liu}, \bibinfo{person}{Edison Marrese-Taylor}, \bibinfo{person}{Qingcheng Zeng}, \bibinfo{person}{Yu~He Ke}, \bibinfo{person}{Wanxin Li}, \bibinfo{person}{Lechao Cheng}, \bibinfo{person}{Qingyu Chen}, \bibinfo{person}{James Caverlee}, \bibinfo{person}{Yutaka Matsuo}, {and} \bibinfo{person}{Irene Li}.} \bibinfo{year}{2024}\natexlab{a}.
\newblock \bibinfo{title}{KG-Rank: Enhancing Large Language Models for Medical QA with Knowledge Graphs and Ranking Techniques}.
\newblock
\newblock
\showeprint[arxiv]{2403.05881}~[cs.CL]
\urldef\tempurl%
\url{https://arxiv.org/abs/2403.05881}
\showURL{%
\tempurl}


\bibitem[Yang et~al\mbox{.}(2024b)]%
        {ref:crag}
\bibfield{author}{\bibinfo{person}{Xiao Yang}, \bibinfo{person}{Kai Sun}, \bibinfo{person}{Hao Xin}, \bibinfo{person}{Yushi Sun}, \bibinfo{person}{Nikita Bhalla}, \bibinfo{person}{Xiangsen Chen}, \bibinfo{person}{Sajal Choudhary}, \bibinfo{person}{Rongze~Daniel Gui}, \bibinfo{person}{Ziran~Will Jiang}, \bibinfo{person}{Ziyu Jiang}, \bibinfo{person}{Lingkun Kong}, \bibinfo{person}{Brian Moran}, \bibinfo{person}{Jiaqi Wang}, \bibinfo{person}{Yifan~Ethan Xu}, \bibinfo{person}{An Yan}, \bibinfo{person}{Chenyu Yang}, \bibinfo{person}{Eting Yuan}, \bibinfo{person}{Hanwen Zha}, \bibinfo{person}{Nan Tang}, \bibinfo{person}{Lei Chen}, \bibinfo{person}{Nicolas Scheffer}, \bibinfo{person}{Yue Liu}, \bibinfo{person}{Nirav Shah}, \bibinfo{person}{Rakesh Wanga}, \bibinfo{person}{Anuj Kumar}, \bibinfo{person}{Wen tau Yih}, {and} \bibinfo{person}{Xin~Luna Dong}.} \bibinfo{year}{2024}\natexlab{b}.
\newblock \bibinfo{title}{CRAG -- Comprehensive RAG Benchmark}.
\newblock
\newblock
\showeprint[arxiv]{2406.04744}~[cs.CL]
\urldef\tempurl%
\url{https://arxiv.org/abs/2406.04744}
\showURL{%
\tempurl}


\bibitem[Yang et~al\mbox{.}(2018)]%
        {ref:hotpotqa}
\bibfield{author}{\bibinfo{person}{Zhilin Yang}, \bibinfo{person}{Peng Qi}, \bibinfo{person}{Saizheng Zhang}, \bibinfo{person}{Yoshua Bengio}, \bibinfo{person}{William~W. Cohen}, \bibinfo{person}{Ruslan Salakhutdinov}, {and} \bibinfo{person}{Christopher~D. Manning}.} \bibinfo{year}{2018}\natexlab{}.
\newblock \showarticletitle{HotpotQA: {A} Dataset for Diverse, Explainable Multi-hop Question Answering}. In \bibinfo{booktitle}{\emph{Proceedings of the 2018 Conference on Empirical Methods in Natural Language Processing, Brussels, Belgium, October 31 - November 4, 2018}}. \bibinfo{pages}{2369--2380}.
\newblock


\bibitem[Yani and Krisnadhi(2021)]%
        {ref:kbqasurvey4}
\bibfield{author}{\bibinfo{person}{Mohammad Yani} {and} \bibinfo{person}{Adila~Alfa Krisnadhi}.} \bibinfo{year}{2021}\natexlab{}.
\newblock \showarticletitle{Challenges, Techniques, and Trends of Simple Knowledge Graph Question Answering: {A} Survey}.
\newblock \bibinfo{journal}{\emph{Inf.}} \bibinfo{volume}{12}, \bibinfo{number}{7} (\bibinfo{year}{2021}), \bibinfo{pages}{271}.
\newblock


\bibitem[Yasunaga et~al\mbox{.}(2021)]%
        {ref:qa-gnn}
\bibfield{author}{\bibinfo{person}{Michihiro Yasunaga}, \bibinfo{person}{Hongyu Ren}, \bibinfo{person}{Antoine Bosselut}, \bibinfo{person}{Percy Liang}, {and} \bibinfo{person}{Jure Leskovec}.} \bibinfo{year}{2021}\natexlab{}.
\newblock \showarticletitle{{QA-GNN:} Reasoning with Language Models and Knowledge Graphs for Question Answering}. In \bibinfo{booktitle}{\emph{Proceedings of the 2021 Conference of the North American Chapter of the Association for Computational Linguistics: Human Language Technologies, {NAACL-HLT} 2021, Online, June 6-11, 2021}}. \bibinfo{pages}{535--546}.
\newblock


\bibitem[Ye et~al\mbox{.}(2024)]%
        {ref:instructglm}
\bibfield{author}{\bibinfo{person}{Ruosong Ye}, \bibinfo{person}{Caiqi Zhang}, \bibinfo{person}{Runhui Wang}, \bibinfo{person}{Shuyuan Xu}, {and} \bibinfo{person}{Yongfeng Zhang}.} \bibinfo{year}{2024}\natexlab{}.
\newblock \bibinfo{title}{Language is All a Graph Needs}.
\newblock
\newblock
\showeprint[arxiv]{2308.07134}~[cs.CL]
\urldef\tempurl%
\url{https://arxiv.org/abs/2308.07134}
\showURL{%
\tempurl}


\bibitem[Ye et~al\mbox{.}(2021)]%
        {ref:rng}
\bibfield{author}{\bibinfo{person}{Xi Ye}, \bibinfo{person}{Semih Yavuz}, \bibinfo{person}{Kazuma Hashimoto}, \bibinfo{person}{Yingbo Zhou}, {and} \bibinfo{person}{Caiming Xiong}.} \bibinfo{year}{2021}\natexlab{}.
\newblock \showarticletitle{Rng-kbqa: Generation augmented iterative ranking for knowledge base question answering}.
\newblock \bibinfo{journal}{\emph{arXiv preprint arXiv:2109.08678}} (\bibinfo{year}{2021}).
\newblock


\bibitem[Yih et~al\mbox{.}(2016)]%
        {ref:webqsp}
\bibfield{author}{\bibinfo{person}{Wen{-}tau Yih}, \bibinfo{person}{Matthew Richardson}, \bibinfo{person}{Christopher Meek}, \bibinfo{person}{Ming{-}Wei Chang}, {and} \bibinfo{person}{Jina Suh}.} \bibinfo{year}{2016}\natexlab{}.
\newblock \showarticletitle{The Value of Semantic Parse Labeling for Knowledge Base Question Answering}. In \bibinfo{booktitle}{\emph{Proceedings of the 54th Annual Meeting of the Association for Computational Linguistics, {ACL} 2016, August 7-12, 2016, Berlin, Germany, Volume 2: Short Papers}}.
\newblock


\bibitem[Yu et~al\mbox{.}(2023)]%
        {ref:decaf}
\bibfield{author}{\bibinfo{person}{Donghan Yu}, \bibinfo{person}{Sheng Zhang}, \bibinfo{person}{Patrick Ng}, \bibinfo{person}{Henghui Zhu}, \bibinfo{person}{Alexander~Hanbo Li}, \bibinfo{person}{Jun Wang}, \bibinfo{person}{Yiqun Hu}, \bibinfo{person}{William~Yang Wang}, \bibinfo{person}{Zhiguo Wang}, {and} \bibinfo{person}{Bing Xiang}.} \bibinfo{year}{2023}\natexlab{}.
\newblock \showarticletitle{DecAF: Joint Decoding of Answers and Logical Forms for Question Answering over Knowledge Bases}. In \bibinfo{booktitle}{\emph{The Eleventh International Conference on Learning Representations, {ICLR} 2023, Kigali, Rwanda, May 1-5, 2023}}.
\newblock


\bibitem[Yu et~al\mbox{.}(2022)]%
        {ref:kgfid}
\bibfield{author}{\bibinfo{person}{Donghan Yu}, \bibinfo{person}{Chenguang Zhu}, \bibinfo{person}{Yuwei Fang}, \bibinfo{person}{Wenhao Yu}, \bibinfo{person}{Shuohang Wang}, \bibinfo{person}{Yichong Xu}, \bibinfo{person}{Xiang Ren}, \bibinfo{person}{Yiming Yang}, {and} \bibinfo{person}{Michael Zeng}.} \bibinfo{year}{2022}\natexlab{}.
\newblock \showarticletitle{KG-FiD: Infusing Knowledge Graph in Fusion-in-Decoder for Open-Domain Question Answering}. In \bibinfo{booktitle}{\emph{Proceedings of the 60th Annual Meeting of the Association for Computational Linguistics (Volume 1: Long Papers), {ACL} 2022, Dublin, Ireland, May 22-27, 2022}}. \bibinfo{pages}{4961--4974}.
\newblock


\bibitem[Yu et~al\mbox{.}(2024)]%
        {ref:ragsurvey6}
\bibfield{author}{\bibinfo{person}{Hao Yu}, \bibinfo{person}{Aoran Gan}, \bibinfo{person}{Kai Zhang}, \bibinfo{person}{Shiwei Tong}, \bibinfo{person}{Qi Liu}, {and} \bibinfo{person}{Zhaofeng Liu}.} \bibinfo{year}{2024}\natexlab{}.
\newblock \bibinfo{title}{Evaluation of Retrieval-Augmented Generation: A Survey}.
\newblock
\newblock
\showeprint[arxiv]{2405.07437}~[cs.CL]
\urldef\tempurl%
\url{https://arxiv.org/abs/2405.07437}
\showURL{%
\tempurl}


\bibitem[Zhang et~al\mbox{.}(2022b)]%
        {ref:sr}
\bibfield{author}{\bibinfo{person}{Jing Zhang}, \bibinfo{person}{Xiaokang Zhang}, \bibinfo{person}{Jifan Yu}, \bibinfo{person}{Jian Tang}, \bibinfo{person}{Jie Tang}, \bibinfo{person}{Cuiping Li}, {and} \bibinfo{person}{Hong Chen}.} \bibinfo{year}{2022}\natexlab{b}.
\newblock \showarticletitle{Subgraph Retrieval Enhanced Model for Multi-hop Knowledge Base Question Answering}. In \bibinfo{booktitle}{\emph{Proceedings of the 60th Annual Meeting of the Association for Computational Linguistics (Volume 1: Long Papers), {ACL} 2022, Dublin, Ireland, May 22-27, 2022}}. \bibinfo{pages}{5773--5784}.
\newblock


\bibitem[Zhang et~al\mbox{.}(2024b)]%
        {ref:graphtranslator}
\bibfield{author}{\bibinfo{person}{Mengmei Zhang}, \bibinfo{person}{Mingwei Sun}, \bibinfo{person}{Peng Wang}, \bibinfo{person}{Shen Fan}, \bibinfo{person}{Yanhu Mo}, \bibinfo{person}{Xiaoxiao Xu}, \bibinfo{person}{Hong Liu}, \bibinfo{person}{Cheng Yang}, {and} \bibinfo{person}{Chuan Shi}.} \bibinfo{year}{2024}\natexlab{b}.
\newblock \showarticletitle{GraphTranslator: Aligning Graph Model to Large Language Model for Open-ended Tasks}. In \bibinfo{booktitle}{\emph{Proceedings of the {ACM} on Web Conference 2024, {WWW} 2024, Singapore, May 13-17, 2024}}. \bibinfo{pages}{1003--1014}.
\newblock


\bibitem[Zhang et~al\mbox{.}(2024a)]%
        {ref:knowgpt}
\bibfield{author}{\bibinfo{person}{Qinggang Zhang}, \bibinfo{person}{Junnan Dong}, \bibinfo{person}{Hao Chen}, \bibinfo{person}{Daochen Zha}, \bibinfo{person}{Zailiang Yu}, {and} \bibinfo{person}{Xiao Huang}.} \bibinfo{year}{2024}\natexlab{a}.
\newblock \bibinfo{title}{KnowGPT: Knowledge Graph based Prompting for Large Language Models}.
\newblock
\newblock
\showeprint[arxiv]{2312.06185}~[cs.CL]
\urldef\tempurl%
\url{https://arxiv.org/abs/2312.06185}
\showURL{%
\tempurl}


\bibitem[Zhang et~al\mbox{.}(2022a)]%
        {ref:greaselm}
\bibfield{author}{\bibinfo{person}{Xikun Zhang}, \bibinfo{person}{Antoine Bosselut}, \bibinfo{person}{Michihiro Yasunaga}, \bibinfo{person}{Hongyu Ren}, \bibinfo{person}{Percy Liang}, \bibinfo{person}{Christopher~D. Manning}, {and} \bibinfo{person}{Jure Leskovec}.} \bibinfo{year}{2022}\natexlab{a}.
\newblock \showarticletitle{GreaseLM: Graph REASoning Enhanced Language Models}. In \bibinfo{booktitle}{\emph{The Tenth International Conference on Learning Representations, {ICLR} 2022, Virtual Event, April 25-29, 2022}}.
\newblock


\bibitem[Zhang et~al\mbox{.}(2018)]%
        {ref:metaqa}
\bibfield{author}{\bibinfo{person}{Yuyu Zhang}, \bibinfo{person}{Hanjun Dai}, \bibinfo{person}{Zornitsa Kozareva}, \bibinfo{person}{Alexander~J. Smola}, {and} \bibinfo{person}{Le Song}.} \bibinfo{year}{2018}\natexlab{}.
\newblock \showarticletitle{Variational Reasoning for Question Answering With Knowledge Graph}. In \bibinfo{booktitle}{\emph{Proceedings of the Thirty-Second {AAAI} Conference on Artificial Intelligence, (AAAI-18), the 30th innovative Applications of Artificial Intelligence (IAAI-18), and the 8th {AAAI} Symposium on Educational Advances in Artificial Intelligence (EAAI-18), New Orleans, Louisiana, USA, February 2-7, 2018}}. \bibinfo{pages}{6069--6076}.
\newblock


\bibitem[Zhao et~al\mbox{.}(2023)]%
        {ref:graphtext}
\bibfield{author}{\bibinfo{person}{Jianan Zhao}, \bibinfo{person}{Le Zhuo}, \bibinfo{person}{Yikang Shen}, \bibinfo{person}{Meng Qu}, \bibinfo{person}{Kai Liu}, \bibinfo{person}{Michael Bronstein}, \bibinfo{person}{Zhaocheng Zhu}, {and} \bibinfo{person}{Jian Tang}.} \bibinfo{year}{2023}\natexlab{}.
\newblock \bibinfo{title}{GraphText: Graph Reasoning in Text Space}.
\newblock
\newblock
\showeprint[arxiv]{2310.01089}~[cs.CL]
\urldef\tempurl%
\url{https://arxiv.org/abs/2310.01089}
\showURL{%
\tempurl}


\bibitem[Zhao et~al\mbox{.}(2024)]%
        {ref:ragsurvey4}
\bibfield{author}{\bibinfo{person}{Penghao Zhao}, \bibinfo{person}{Hailin Zhang}, \bibinfo{person}{Qinhan Yu}, \bibinfo{person}{Zhengren Wang}, \bibinfo{person}{Yunteng Geng}, \bibinfo{person}{Fangcheng Fu}, \bibinfo{person}{Ling Yang}, \bibinfo{person}{Wentao Zhang}, \bibinfo{person}{Jie Jiang}, {and} \bibinfo{person}{Bin Cui}.} \bibinfo{year}{2024}\natexlab{}.
\newblock \bibinfo{title}{Retrieval-Augmented Generation for AI-Generated Content: A Survey}.
\newblock
\newblock
\showeprint[arxiv]{2402.19473}~[cs.CV]
\urldef\tempurl%
\url{https://arxiv.org/abs/2402.19473}
\showURL{%
\tempurl}


\bibitem[Zheng et~al\mbox{.}(2024)]%
        {ref:med3}
\bibfield{author}{\bibinfo{person}{Yanxin Zheng}, \bibinfo{person}{Wensheng Gan}, \bibinfo{person}{Zefeng Chen}, \bibinfo{person}{Zhenlian Qi}, \bibinfo{person}{Qian Liang}, {and} \bibinfo{person}{Philip~S. Yu}.} \bibinfo{year}{2024}\natexlab{}.
\newblock \bibinfo{title}{Large Language Models for Medicine: A Survey}.
\newblock
\newblock
\showeprint[arxiv]{2405.13055}~[cs.CL]
\urldef\tempurl%
\url{https://arxiv.org/abs/2405.13055}
\showURL{%
\tempurl}


\bibitem[Zhu et~al\mbox{.}(2024)]%
        {ref:engine}
\bibfield{author}{\bibinfo{person}{Yun Zhu}, \bibinfo{person}{Yaoke Wang}, \bibinfo{person}{Haizhou Shi}, {and} \bibinfo{person}{Siliang Tang}.} \bibinfo{year}{2024}\natexlab{}.
\newblock \bibinfo{title}{Efficient Tuning and Inference for Large Language Models on Textual Graphs}.
\newblock
\newblock
\showeprint[arxiv]{2401.15569}~[cs.CL]
\urldef\tempurl%
\url{https://arxiv.org/abs/2401.15569}
\showURL{%
\tempurl}


\end{thebibliography}

\end{document}